\documentclass[journal]{IEEEtran}
\ifCLASSINFOpdf
\else
\fi
\usepackage{graphicx}
\usepackage{subfigure}  
\usepackage{underscore}
\usepackage{amsmath}
\usepackage{color}
\usepackage{stfloats}
\usepackage{lipsum}
\usepackage{amssymb}
\usepackage{float}
\usepackage{booktabs}
\usepackage{bbding} 
\usepackage{amssymb}
\usepackage{pifont} 
\graphicspath{{figures/}}
\begin{document}
%
\title{Multi-attention Associate Prediction Network\\ for Visual Tracking}
%
%
%

\author{Xinglong Sun, Haijiang Sun$*$, Shan Jiang, Jiacheng Wang, Xilai Wei, Zhonghe Hu
\thanks{$*$ Corresponding author}
\thanks{X. Sun, H. Sun, S. Jiang, J. Wang and X. Wei are with the Changchun Institute of Optics, Fine Mechanics and Physics, Chinese Academy of Science, Changchun 130033, China (e-mail: sunxinglong@ciomp.ac.cn; sunhj@ciomp.ac.cn; jiangshan$\_$ciomp@qq.com; wangjiacheng@ciomp.ac.cn; ln$\_$weixilai@163.com)}
\thanks{Z. Hu are with the Northwest institute of nuclear technology, Xian 710600, China (e-mail:unusualHz@163.com)}

\thanks{This work was supported by the National Natural Science Foundation of China under Grant No. 61401425, 61602432.}
\thanks{Manuscript received April 19, 2005; revised August 26, 2015.}}

%
%

\markboth{IEEE TRANSACTIONS ON MULTIMEDIA,~Vol.~14, No.~8, August~2015}%
{Shell \MakeLowercase{\textit{et al.}}: Bare Demo of IEEEtran.cls for IEEE Journals}
%



\maketitle

\begin{abstract}
Classification-regression prediction networks have realized impressive success in several modern deep trackers. However, there is an inherent difference between classification and regression tasks, so they have diverse even opposite demands for feature matching. Existed models always ignore the key issue and only employ a unified matching block in two task branches, decaying the decision quality. Besides, these models also struggle with decision misalignment situation. In this paper, we propose a multi-attention associate prediction network (MAPNet) to tackle the above problems. Concretely, two novel matchers, i.e., category-aware matcher and spatial-aware matcher, are first designed for feature comparison by integrating self, cross, channel or spatial attentions organically. They are capable of fully capturing the category-related semantics for classification and the local spatial contexts for regression, respectively. Then, we present a dual alignment module to enhance the correspondences between two branches, which is useful to find the optimal tracking solution. Finally, we describe a Siamese tracker built upon the proposed prediction network, which achieves the leading performance on five tracking benchmarks, consisting of LaSOT, TrackingNet, GOT-10k, TNL2k and UAV123, and surpasses other state-of-the-art approaches.
\end{abstract}


%
\IEEEpeerreviewmaketitle

\section{Introduction}
%
%
%
%
\IEEEPARstart{V}{isual} object tracking is a fundamental and important topic in computer vision, aiming to estimate the location state of a given arbitrary target in the whole video sequence. In recent decades, the technology attracts massive attentions due to its wide applications ranging from visual surveillance \cite{application1}, robotics \cite{application2}, augmented reality \cite{application3} to human computer interaction \cite{application4}. However, it remains challenging to achieve high-quality tracking due to occlusion, illumination variation, background clutter and other distractors. 

With the development of deep learning, some more efficient and intelligent algorithms are exploited to address the above interference factors, which pay massive efforts to improve the tracking performance from different perspectives. Specifically, several methods \cite{SiamRPN++}, \cite{sbt} aim to enhance feature representation by introducing more abstract backbones, like ResNet \cite{ResNet} and transformer \cite{trnsformer}, etc. In addition, other works \cite{hcf}, \cite{mamltracker} expect to promote the efficiency and quality of offline optimization and online learning by exploring transferring learning or meta learning \cite{oslopt2}. Nowadays, numerous studies uncover that state prediction is extremely critical for object tracking, which usually directly determines the overall performance of trackers. In this case, various state-of-the-art prediction paradigms are discussed to better estimate the object state \cite{mdnet}, \cite{stark}, \cite{siamrpn}.

Classification-regression model is the most excellent and representative among all kinds of prediction architectures. It generally decomposes visual tracking into two subtasks, and adopts two parallel decision branches to distinguish the object from background and locate its bounding box simultaneously. Whereas, despite realizing satisfactory state prediction, existed models still suffer from several fatal drawbacks. Firstly, there are diverse even contradictory demands for feature matching between classification and regression. Regression expects that the matcher focuses more on low-level spatial details to lift the location precision, while classification hopes to abandon these details and prefers high-level semantic attributes to effectively identify the object. Previous works \cite{siamrpn}, \cite{transt}, \cite{ostrack} always ignore the above key issues, and employ only one unified matching block for both branches, limiting the robustness and precision of tracking. Moreover, classification and regression are often performed in a separate manner, which never communicate each other in decision phase. It may cause the correspondence between two prediction branches is poor, i.e., the sample with a high classification score may have an inferior regression accuracy, producing imperfect tracking outputs. 

\begin{figure}[t]
	\begin{center}
		\includegraphics[width=0.9\linewidth]{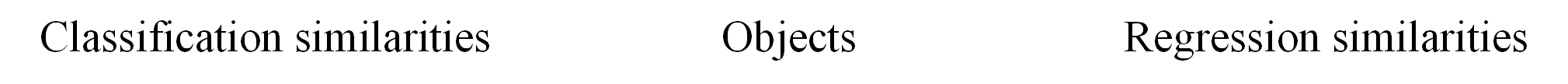}
		\includegraphics[width=0.9\linewidth]{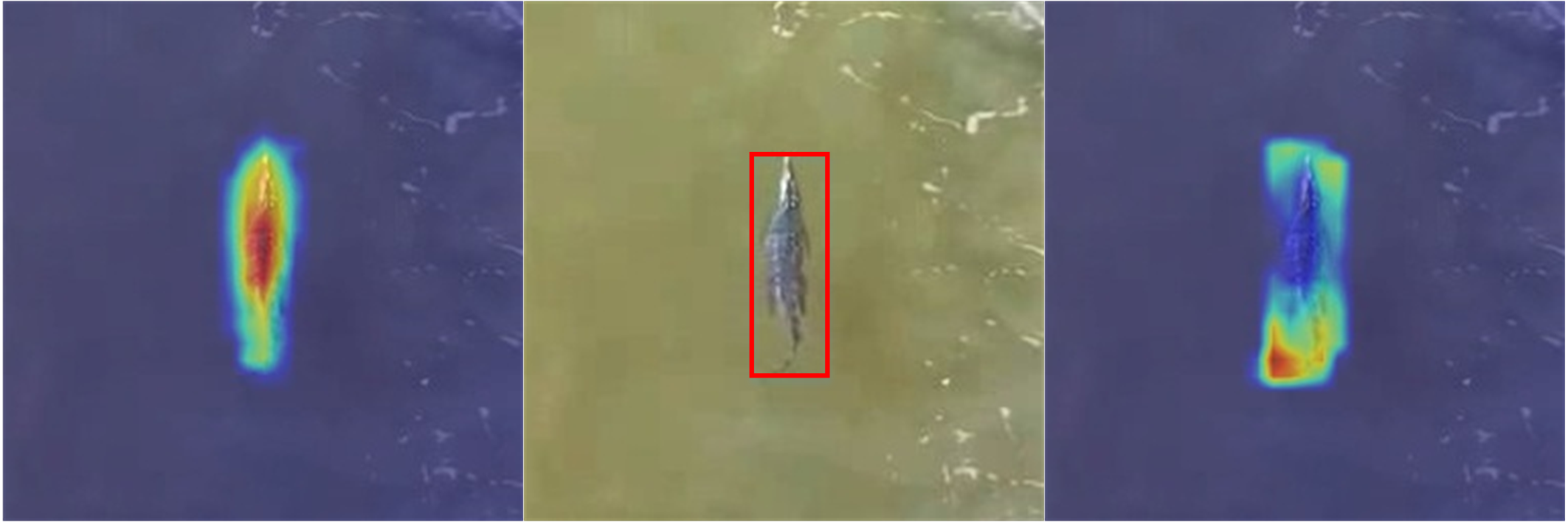}
		\includegraphics[width=0.9\linewidth]{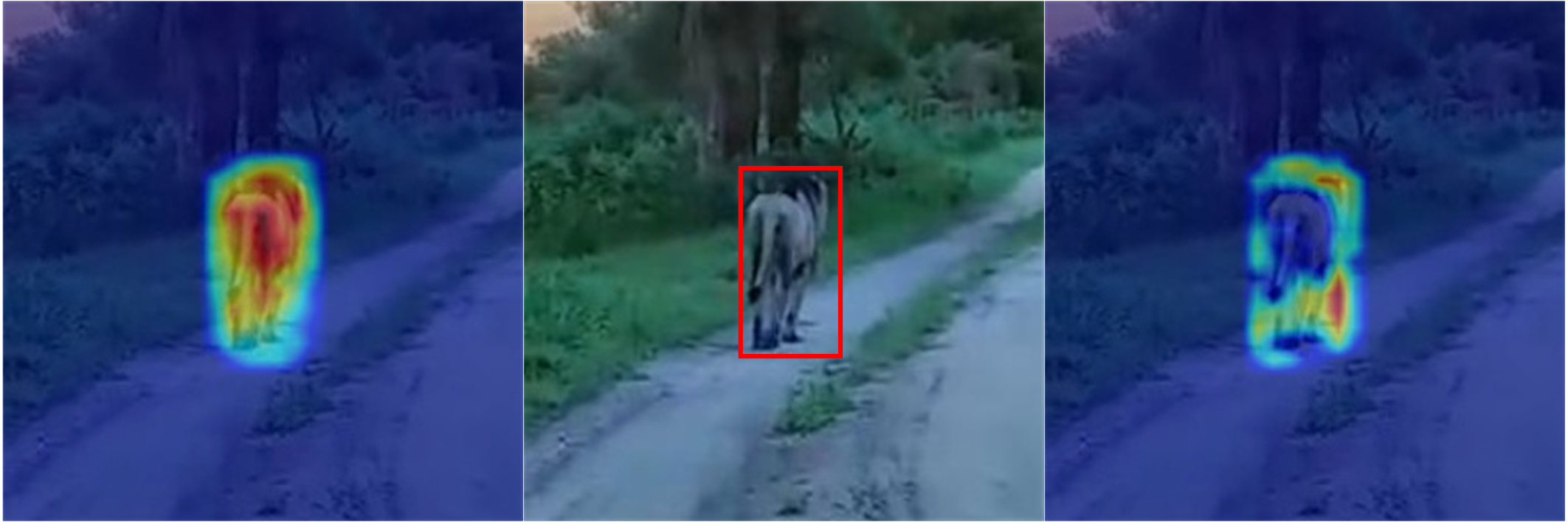}
	\end{center}
	\caption{Classification and regression similarity maps produced by MAPNet. The prediction network can extract more category-related responses for classification and local texture information for location.}
	\label{first-fig}
\end{figure}

To address these problems, this paper proposes a multi-attention associate prediction network by exploiting different attention mechanisms. Concretely, we first design two specific matchers for feature interaction, i.e., category-aware matcher and spatial-aware matcher. The former carefully combines the channel, self and cross attentions to compare the features of template and search region, which is able to fully model their dependence relationships as well as encode the channel-based category patterns. While the latter takes advantage of spatial attentions rather than channel attentions to perceive the spatial detail distribution of object. The proposed network embeds the above two matchers into classification and regression branches respectively, obtaining more abundant and suitable matching responses for state prediction. Then, a novel dual alignment module is presented to promote the decision correspondence of two branches. For classification and regression similarity features, the module exploits two cascaded cross-attentions to progressively aggregate them, which may modulate each other to decrease the misalignment probabilities. Fig. \ref{first-fig} provides a few representative similarity response maps, illustrating that our prediction network is helpful to achieve both robust instance classification and precise coordinate location. 

For visual tracking, we describe a Siamese tracker built upon the proposed prediction network, named as MAPNet-R. The tracker first extracts the features of template and search region with ResNet-50 \cite{ResNet}, and then compares them using the prediction network. Finally, a classification and a regression heads are employed to complete object tracking in a per-pixel manner. To ensure the availability, we also design two cross-guided loss functions for model optimization. The presented tracker is evaluated on five public benchmarks, including LaSOT \cite{lasot}, TrackingNet \cite{trackingnet}, GOT-10k \cite{GOT-10k}, TNL2k \cite{tnl2k} and UAV123 \cite{uav}. Experimental results manifest the superiorities of our method, proving that the proposed network is more effective than other prediction models. 

In summary, the main contributions of our work are listed as follows:

1. We propose two powerful feature matchers by exploring multiple types of attentions, which are useful to fully capture the category semantic patterns for classification and the spatial detailed cues for location, respectively.

2. An associate prediction network is designed based on the proposed matchers. It allows for simultaneously obtaining more accurate similarity maps for classification and regression, and enhancing their correspondence for high-quality object state estimation.
 
3. Numerous experiments are executed on several popular benchmarks to evaluate the capability of the presented method, demonstrating that it surpasses other state-of-the-art trackers with the leading performance.

The rest of this paper is organized as follows. We first review the related works in Section II. Then, the proposed prediction network is carefully introduced in Section III, and the Siamese tracker based on this network is presented in Section IV. After analyzing the experimental results on some latest benchmarks in Section V, we conclude the paper and discuss the future works in Section VI.  

\section{Related Works}
In this section, we carefully review the related works about state prediction approaches and attention mechanisms, as well as briefly introduce the recent literatures about Siamese trackers. 

\subsection{State Prediction Approaches} \label{sect2-1}
 Recently, a large number of powerful prediction paradigms are developed based on neural networks, such as classification models, regression models and classification-regression models. Classification models \cite{mdnet}, \cite{Conference29} compute the confidence scores of all candidates and take the sample with the highest score as tracking result, while regression models \cite{stark}, \cite{seqtrack} directly refine the object coordinates on deep feature maps. Different from them, classification-regression models \cite{siamrpn}, \cite{transt} estimate the confidence scores and coordinate offsets simultaneously. Due to obvious performance advantage, the frameworks are widely discussed in a lot of literatures. Concretely, SiamRPN \cite{siamrpn} first combined Region Proposal Network (RPN) \cite{faster-rcnn} into Siamese pipeline for object-background classification and bounding-box regression, following by C-RPN \cite{c-rpn} to furtherly release its potentials. Then, to avoid the massive hyperparameters of RPN, several anchor-free prediction models continued to be presented, like SiamFC++ \cite{siamfc++}, Ocean \cite{ocean} and SiamBAN \cite{siamban}, which could infer the object state without presetting any prior points or boxes. Nowadays, scholars found that the key of classification and regression is the similarity comparison. As a result, TrSiam \cite{trsiam} studied an improved transformer to capture the temporal-spatial contexts among multi-time samples, while TransT \cite{transt} designed an attention-based fusion block for dependence modeling. Besides, a few transformer-based backbone are explored, i.e., SBT \cite{sbt} and SwinTrack \cite{swintrack}, which directly compare the object features during extracting them. Furtherly, OSTrack \cite{ostrack} employed a one-stream framework to unify feature extraction and relation learning, and VideoTrack \cite{videotrack} designed a feedforward video model to encode temporal contexts into spatial features.
 
 For the above methods, all of them ignore the requirement differences for feature comparison between classification and regression, and only adopt single matcher for diverse decision issues. Moreover, these works pay little attention to lifting the prediction correspondence of two branches. Unfortunately, these drawbacks may prevent them to find the optimal tracking solution.  
\subsection{Attention Mechanisms}
As universal visual operators, attention mechanisms have been widely applied in various aspects, such as segmentation, detection, tracking, etc. According to the principle differences, Existed attentions can be divided into two categories. The first types try to highlight the discriminative feature components, consisting of the channel and the spatial attentions. SENet \cite{senet} presented a squeeze-and-excitation module to channel-wisely adjust the original features. CBAM \cite{cbam} studied a convolution-based attentional block, which adaptively refines the features on both channel and spatial dimensions. For tracking, Sa-Siam \cite{sa-siam} used a channel attention to adjust the channel distribution of object features. Besides, channel and spatial attentions are adopted to analyze features simultaneously in several recent trackers, like RasNet \cite{rasnet} and Ta-ASiam \cite{ta-asiam}. The other type of attentions aim to learn the dependence relationships inside or between feature sequences, including self-attention and cross-attention, both of which are originated from the multi-head transformer attention \cite{trnsformer}. The attention scans each element in the whole input sequence when updating the current element, and thus learning the global dependence attributes. At present, these attentions have been adopted to complete different issues while tracking an object, i.e., feature representation \cite{transclass}, feature comparison \cite{transt}, temporal modeling \cite{trsiam}, etc. In this work, we will combine the above two kinds of attentions to execute more sufficient and reliable feature interaction for state prediction.

\begin{figure*}
	\begin{center}
		\includegraphics[width=0.9\linewidth]{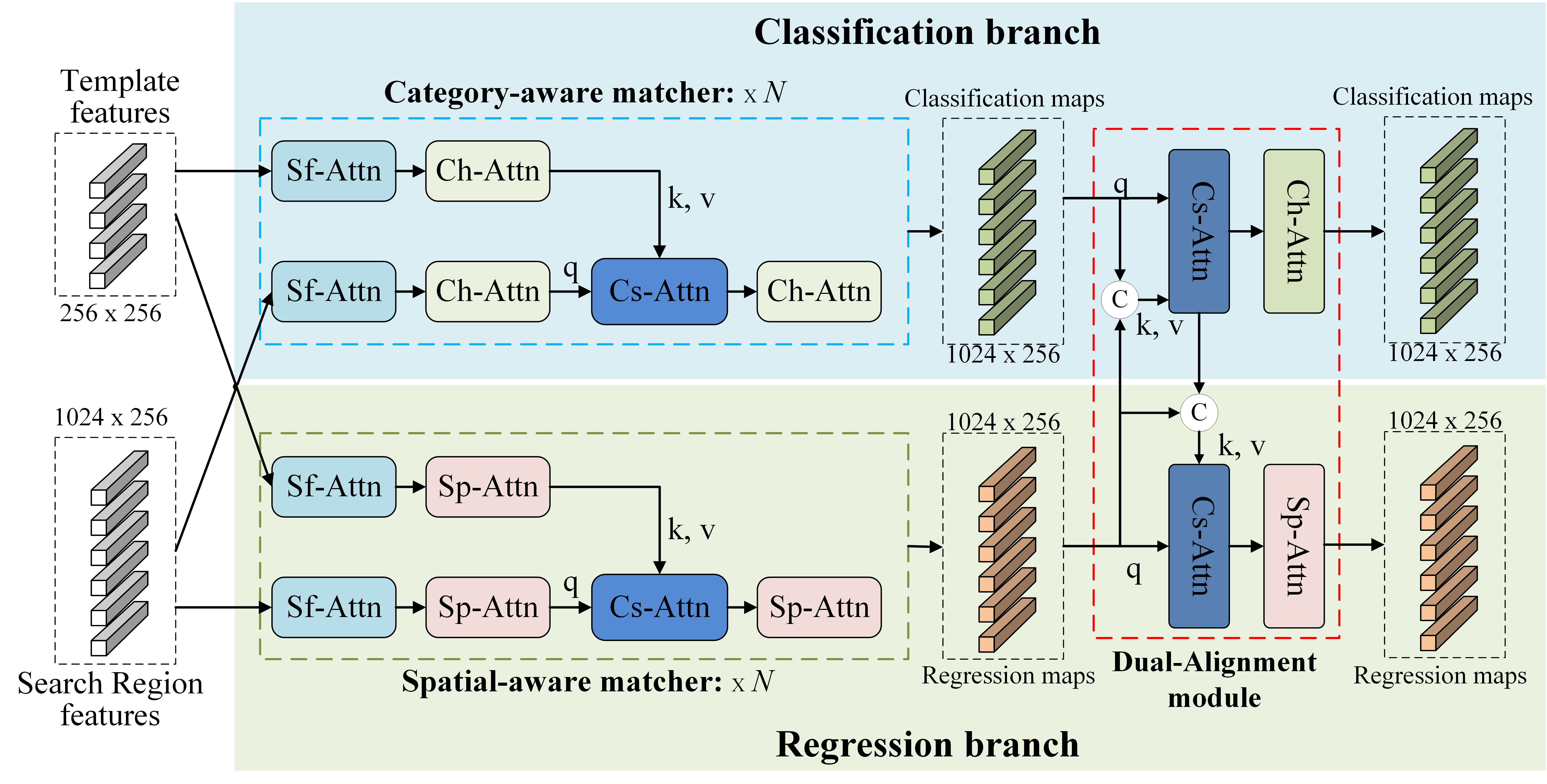}
	\end{center}
	\caption{Overview of the proposed prediction network, consisting of category-aware matchers, spatial-aware matchers and dual alignment module. \emph{Ch-Attn}, \emph{Sp-Attn}, \emph{Sf-Attn} and \emph{Cs-Attn} represent the channel, spatial, self and cross attentions, respectively. The features of template and search region are first compared by diverse matchers, and then two kinds of similarity maps are aligned by the dual alignment module.}
	\label{second-fig}
\end{figure*}

\subsection{Siamese Trackers}
Siamese network serves as a popular and strong tracking architecture, which formulates object tracking as learning a metric function in high-dimensional feature space. Following the seminal work i.e., SiamFC \cite{siamfc}, which exploited a cross-correlation layer to match the features of template and search region, massive efforts are paid to fully promote the tracking capabilities. Among them, an important direction is to improve the state prediction level, so various state-of-the-art prediction models \cite{stark}, \cite{siamrpn}, \cite{transt} have been introduced into Siamese trackers. Another representative development is the evolutions of feature representation. To obtain more abstract features, SiamRPN++ \cite{SiamRPN++} collected spatial-aware samples to avoid the location bias induced by padding operation, while SiamDW \cite{SiamDW} directly designed a novel residual block without padding. SBT \cite{sbt} and MixFormer \cite{mixformer} recently employed transformer networks as backbones, which are able to extract and compare multi-stage object features simultaneously. In addition to the above issues, how to improve the quality of offline training \cite{dasiamrpn} and online learning \cite{stark} are also carefully discussed.

\section{Multi-attention Associate Prediction Network}
In this section, we describe the proposed prediction network carefully. After giving the overall network architecture, the basic theories of multiple kinds of attentions are presented. Then, we introduce the key components of our network, i.e., category-aware matcher, spatial-aware matcher and dual alignment module.

\subsection{Network Architecture}
The architecture of our presented network is depicted in Fig. \ref{second-fig}. In contrast to existed prediction models with only one kind of matchers, our work exploits both category-aware matchers and spatial-aware matchers to match the features of template and search region, which is critical to simultaneously satisfy the opposite matching requirements of classification and regression tasks. Moreover, a feature alignment module is designed to solve the misalignment problem faced by previous cases [14], enhancing the prediction consistency of two branches. 

Specifically, for the feature vectors of template $v_z \in \mathbb{R}^{n_{z} \times d}$ and search region $v_x \in \mathbb{R}^{n_{x} \times d}$, the category-aware and the spatial-aware matchers are first used to compare them, which combine multiple types of attentions for effective correlation learning. In our algorithm, each prediction branch consists of $N$ corresponding matchers ($N = 3$), and the search region vectors provided by the last matcher are viewed as the initial matching responses, i.e., classification similarity vectors $s_c \in \mathbb{R}^{n_{x} \times d}$ and regression similarity vectors $s_p \in \mathbb{R}^{n_{x} \times d}$. Then, these raw similarity vectors are furtherly adjusted by the dual alignment module, which introduces two cross-attentions to iteratively aggregate them. Last of all, the adjusted similarity vectors $s_c^{\prime} \in \mathbb{R}^{n_x \times d}$ and $s_p^{\prime} \in \mathbb{R}^{n_x \times d}$ are outputted to estimate the object state.

\subsection{Attentions}
Attention is the key and fundamental unit of the proposed prediction network, so we expound the adopted attentions as follows.
\subsubsection{Channel attention}
Channel attention is explored to channel-wisely highlight the category-related feature components. In a classical channel attention \cite{cbam}, both average and maxing global pooling layers are first utilized to compress the spatial size of features, following by a multi-layer perception ($MLP$) to encode pooling features. Then, two kinds of encoded features are accumulated, and the sum is normalized by a sigmoid function. The channel attention can be formulated as:
\begin{equation}
C(x)=x \cdot g\left(M L P\left(f^m(x)\right)+M L P\left(f^a(x)\right)\right)
\label{eq-1}
\end{equation}
in which, $g$ denotes the sigmoid layer, while $f^m$ and $f^a$ depict the max and the average global pooling layers, respectively. The dot denotes channel-wise product operation.

\begin{figure*}
	\begin{center}
		\includegraphics[width=0.8\linewidth]{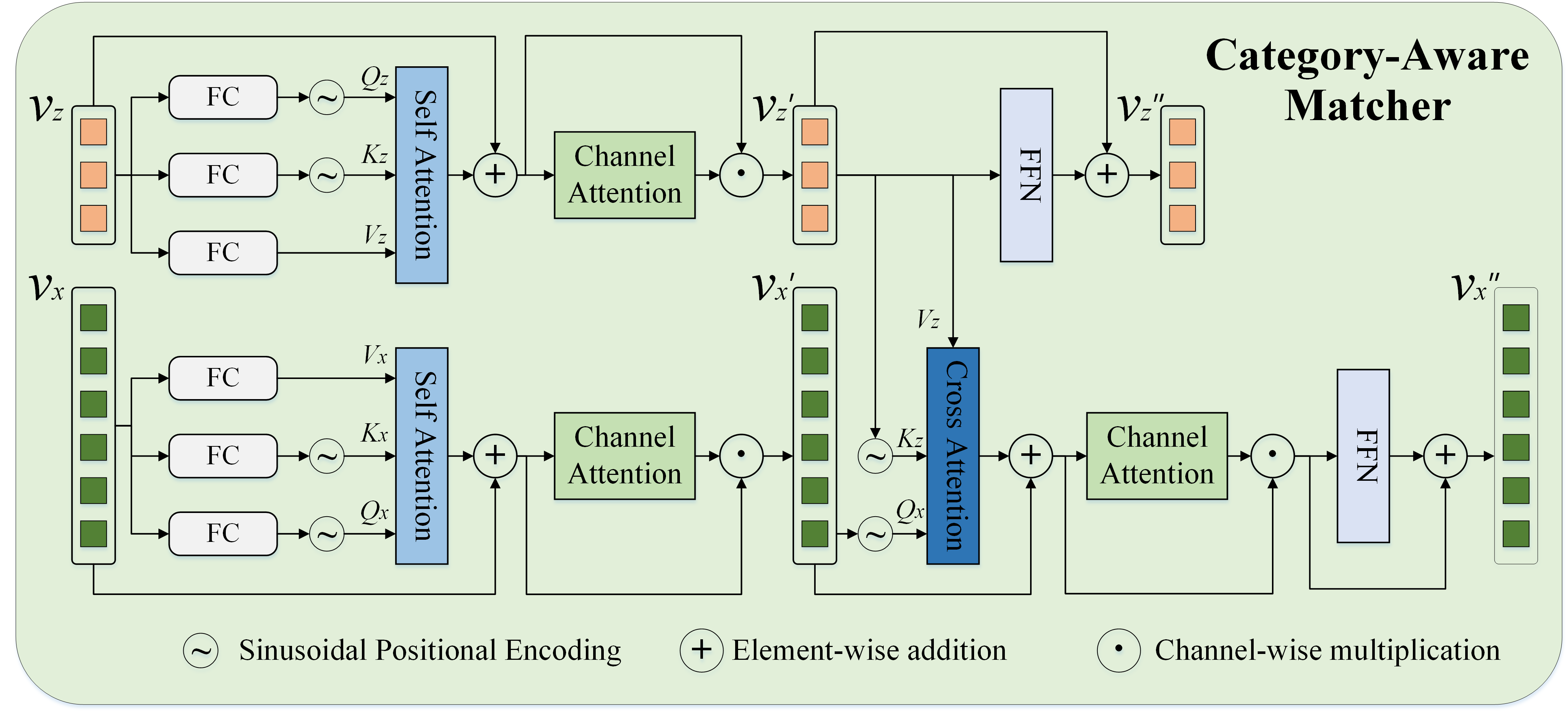}
	\end{center}
	\caption{Architecture of our designed category-aware matcher, which is composed of combining self, cross and channel attentions.}
	\label{third-fig}
\end{figure*}

\subsubsection{Spatial attention}
Spatial attention is able to find the critical local contexts of features, lifting the precision of object location. As described in \cite{cbam}, a typical spatial attention reduces the channel quantity of features with both average and max channel pooling layers, and then learns the local spatial patterns with a convolutional layer. Finally, a sigmoid layer is imposed on the sum of the pooling features. The spatial attention can be described as:
\begin{equation}
S(x)=x \times g\left({Conv}\left(f^{c m}(x)\right)+{Conv}\left(f^{c a}(x)\right)\right)
\label{eq-2}
\end{equation}
where, $Conv$ denotes the convolution layer, while $f^{cm}$ and $f^{ca}$ depict the max and the average channel pooling layers, respectively. $\times$ indicates the pixel-wise product operation.
  
\subsubsection{Self-attention and Cross-attention}
Both self-attention and cross-attention are sourced from the multi-head attention, i.e., the core block of transformer \cite{trnsformer}. Giving the inputs of query $Q \in \mathbb{R}^{N_q \times C}$, key $K\in \mathbb{R}^{N_k \times C}$ and value $V \in \mathbb{R}^{N_v \times C}$, the attention first computes the dot-products of query and key sequences, and adopts a Softmax function to get the weight matrix. Then, the value sequence is weighted by the matrix to output the final responses:
\begin{equation}
\operatorname{Attention}(Q, K, V)=\operatorname{softmax}\left(\frac{Q K^T}{\sqrt{d_k}}\right) V
\end{equation} 
in which, $d_k$ is the dimensionality of key sequence.

The multi-head attention contains $M$ single attention heads which are simply concatenated in channel axis, which is very helpful to lift the diversity of representation.
\begin{equation}
\text {MultiHead}(Q, K, V)=\operatorname{Concat}\left(H_1, \ldots, H_n\right) W^o
\label{eq-4}
\end{equation} 
\begin{equation}
H_i=\operatorname{Attention}\left(Q W_i^Q, K W_i^K, V W_i^V\right)
\end{equation} 
in which, $W_i^{Q} \in \mathbb{R}^{d_{m} \times d_{k}}$, $W_i^{K}  \in \mathbb{R}^{d_{m} \times d_{k}}$, and $W_i^{V}  \in \mathbb{R}^{d_{m} \times d_{v}}$ denote the projection matrices. In practice, we adopt $n=8$ single attention heads, and set $d_k=d_v=d_m⁄n=64$.

\subsection{Category-aware and Spatial-aware Matchers}
Feature matcher plays a vital role in our prediction network to recognize the current object according to its historic attributes. However, previous matchers \cite{siamrpn}, \cite{transt} have no ability to fully model the feature dependences and filter out the real valuable similarity cues simultaneously. Moreover, there is a significant difference between classification and regression, whose matching requirements cannot be satisfied by a single class of matchers. Considering these issues, this work designs two new feature matchers, i.e., category-aware matcher and spatial-aware matcher, which integrate multiple types of attentions to compare and analyze features.

\subsubsection{Category-aware matcher}
For classification, the key of feature matching is to measure the correlations between template and search region features, as well as enhance the category semantic expressions of object. Based on this opinion, we present an efficient category-aware feature matcher, whose overall architecture is shown in Fig. \ref{third-fig}. In contrast to previous matching models \cite{transt}, the matcher performs more abundant attention operations on two sequences of template and search region. Specifically, it first adopts two self-attentions to process every sequence separately to encode the object-specific information. After that, channel attentions are used to channel-wisely adjust two sequences, enhancing the category-related feature components. Then, this matcher models the global dependences between two sequences with a cross-attention, which modulates the features of search region with object template contexts, following by a channel attention to furtherly adjust the channel distributions. Finally, each vector is transmitted into the corresponding feed-forward network to obtain the comparison results. By performing relationship modeling and channel selection alternately, our matcher can capture more discriminative similarity features for classifying object from background. 

Formally, given the features of object template $v_z \in \mathbb{R}^{n_{z} \times d}$ and search region $v_x \in \mathbb{R}^{n_{x} \times d}$, we first introduce several no-shared fully-connect layers to transform them into the tokens of query, key and value, i.e., $Q_z$, $K_z$, $V_z$ and $Q_x$, $K_x$, $V_x$. Next, considering that multi-head attention is permutation-invariant which is not sensitive to the spatial distributions of sequences, sinusoidal positional encoding is added to the query $Q$ and the key $K$. Another notable point is the flatten and unflatten operations during employing channel attentions. Before inputting into channel attentions, feature tokens should be unflatten to 2D dimensions to recover the local structural contexts, i.e., $f_z \in \mathbb{R}^{\sqrt{n_{z}} \times \sqrt{n_{z}} \times d}$ 
and $f_x \in \mathbb{R}^{\sqrt{n_{x}} \times \sqrt{n_{x}} \times d}$. For the features outputted from channel attentions, which need to be flatted to match with the inputted dimensions of self or cross attentions. The core function of category-aware matcher can be formulated as:
\begin{equation}
v_i^{\prime}=C\left(v_i+\text {MultiHead}\left(Q_i, K_i, V_i\right)\right) \quad i \in\{z, x\}
\end{equation}
\begin{equation}
v_x^{\prime \prime}=C\left(v_x^{\prime}+\text {MultiHead}\left(Q_x^{\prime}, K_z^{\prime}, V_z^{\prime}\right)\right)
\end{equation}
in which, $C$ represents the channel attention described in Eq. \ref{eq-1}, and $MultiHead$ is the multi-head attention in Eq. \ref{eq-4}. Generally, the feature vectors of $v_x^{\prime \prime}$ provided by the last category-aware matcher is regarded as the classification similarity map $s_c$. 
\subsubsection{Spatial-aware matcher}
The spatial-aware matcher is comprised by self, cross and spatial attentions, where its basic structure is extremely similar with category-aware matcher. In practice, the only difference is that the channel attention is replaced by the spatial attention, which is important to better capture local detailed information, lifting the precision of object location. The function of spatial-aware matcher can be described as:
\begin{equation}
v_i^{\prime}=S\left(v_i+\text {MultiHead}\left(Q_i, K_i, V_i\right)\right) \quad i \in\{z, x\}
\end{equation}
\begin{equation}
v_x^{\prime \prime}=S\left(v_x^{\prime}+\text {MultiHead}\left(Q_x^{\prime}, K_z^{\prime}, V_z^{\prime}\right)\right)
\end{equation}
in which, $S$ represents the spatial attention introduced in Eq. \ref{eq-2}. The vectors of $v_x^{\prime \prime}$ outputted by the last spatial-aware matcher is usually viewed as the regression similarity map $s_p$.

\subsection{Dual Alignment Module}
Classification and regression branches are supposed to work in a collaborative manner during tracking an object. If simply regarding them as two completely independent subtasks, there may be severe misalignment problem, decaying the prediction level. In this case, we present a dual alignment module, which can element-wisely model the relationships between two types of similarity vectors to enhance their correspondence. For the initial classification and regression similarity vectors, i.e., $s_c \in \mathbb{R}^{n_{x} \times d}$ and $s_p \in \mathbb{R}^{n_{x} \times d}$, they are first concatenated to generate a modulated vector $s_m \in \mathbb{R}^{2n_{x} \times d}$. A cross-attention regards the vector as the vectors of key $K_m$ and value $V_m$ to update the query vector $Q_c$ (original classification vector $s_c$). Then, the updated classification vector is concatenated with the original regression vector to generate a novel modulated vector of $s_m^{\prime}$, which is used to update the regression vector $s_p$ by the other cross-attention. Lastly, we also introduce a channel attention to highlight the category semantic attributes for classification, as well as employ a spatial attention to capture the local spatial textures for regression, respectively. The role of the proposed alignment module is: 
\begin{equation}
s_c^{\prime}=C\left(s_c+\text {MultiHead}\left(Q_c, K_m, V_m\right)\right)
\end{equation}
\begin{equation}
s_p^{\prime}=S\left(s_p+\text {MultiHead}\left(Q_p, K_m^{\prime}, V_m^{\prime}\right)\right)
\end{equation}
where, $s_c^{\prime}$  and $s_p^{\prime}$ are the aligned classification and regression similarity vectors, respectively. With our feature alignment module, two prediction branches can cooperate in a tighter way.

\section{Multi-attention Associate Tracking}
This section introduces a multi-attention associate Siamese tracker built upon the proposed prediction network, i.e., MAP Net-R. After giving the overall pipeline, we carefully describe the backbone for feature extraction, the heads for state decision, and the losses for model optimization. 
\begin{figure}[t]
	\begin{center}
		\includegraphics[width=0.9\linewidth]{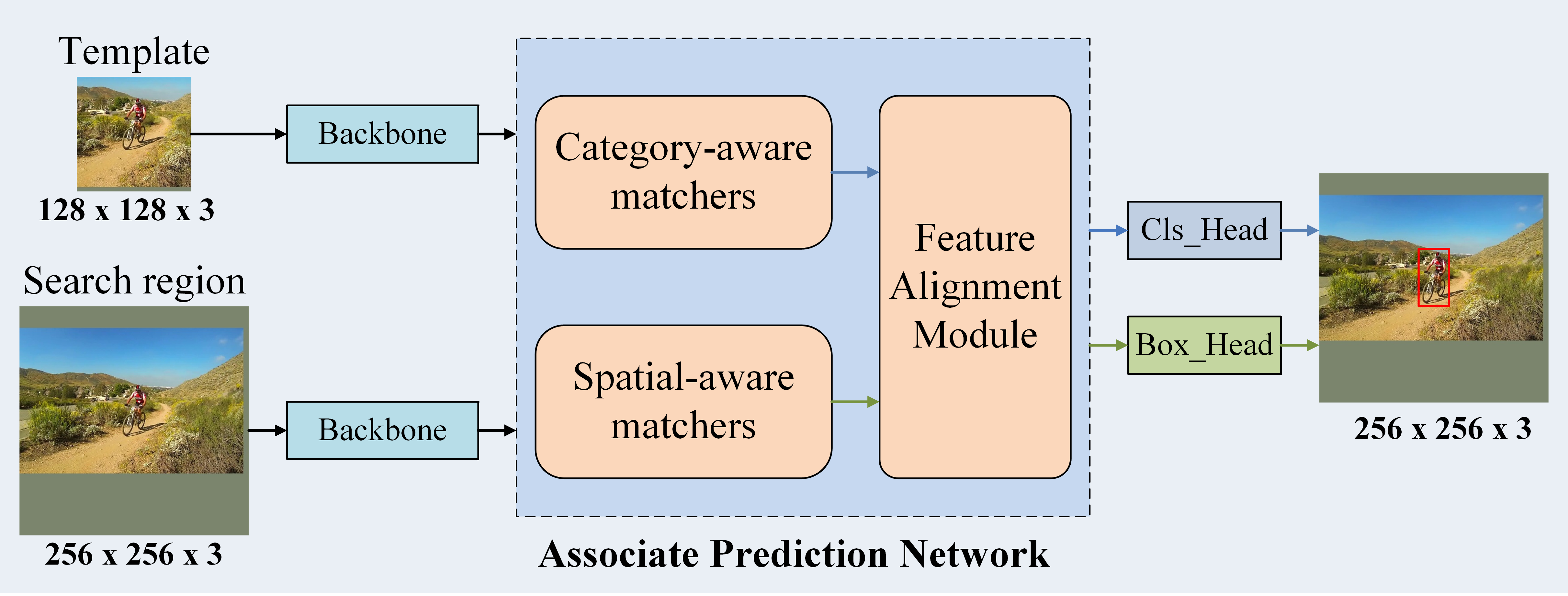}
	\end{center}
	\caption{Pipeline of Siamese tracker based on the proposed prediction network, which is constructed by backbone, prediction network and prediction heads.}
	\label{fourth-fig}
\end{figure}

\subsection{Siamese Pipeline}
The structure of the presented Siamese tracker is illustrated in Fig. \ref{fourth-fig}. It mainly contains three key components: backbone, prediction network and prediction heads. Concretely, a weight-shared backbone is first utilized to extract the features of template and search region patches. Then, these features are compared by both category-aware and spatial-aware matchers to generate the classification and the regression similarity maps, which would be adjusted by the feature alignment module. Finally, two prediction heads perform binary classification and bounding-box regression on the corresponding similarity maps, respectively, outputting the current object location. 

\subsection{Backbone}
Following several previous works \cite{SiamRPN++}, \cite{transt}, we employ the widely-used ResNet-50 \cite{ResNet} for feature extraction, and modify the network carefully to improve its adaptability. Firstly, its last residual block, i.e., the fifth residual block, is removed, and an extra $1\times1$ convolutional layer is appended to decrease the channel quantity of the outputted features from $C$ to $d$. In addition, the convolutional strides in the fourth block are reduced from $2$ to $1$ to enlarge the feature sizes, where the $3\times3$ convolutions are replaced by the dilated convolutions to preserve the receptive fields. For an pair of template image $z \in \mathbb{R}^{H_{z} \times W_{z} \times 3}$ and search region image $x \in \mathbb{R}^{H_{x} \times W_{x} \times 3}$, this backbone can extract their features of $f_{z} \in \mathbb{R}^{\frac{H_{z}}{8} \times \frac{W_{z}}{8} \times d}$ and $f_{x} \in \mathbb{R}^{\frac{H_{x}}{8} \times \frac{W_{x}}{8} \times d}$ ($d=256$). Next, these features would be flatted in spatial dimension, providing the inputted vectors of $v_z \in \mathbb{R}^{n_{z} \times d}$ and $v_x \in \mathbb{R}^{n_{x} \times d}$ for prediction network, in which $n_z = \frac{H_{z}}{8} \times \frac{W_{z}}{8}$ and $n_x = \frac{H_{x}}{8} \times \frac{W_{x}}{8}$.
\subsection{Prediction Heads}
Two parallel prediction heads \cite{transt} are adopted to complete the final classification and regression operations, respectively. Each head is a typical three-layer fully-connected block with hidden channel of 256, where a ReLU activation layer is used to enhance the nonlinearity. Given the classification similarity map $s_c^{\prime}$, classification head computes the confidence score of each element, outputting $n_x$ vectors with lengths of $2$. Regression head estimates the normalized positions relative to the size of search region on every unit of regression similarity map $s_p^{\prime}$, producing $n_x$ coordinate vectors with lengths of $4$. Due to not depend on any anchor-based priors\cite{siamrpn}, \cite{siamban}, this head is more flexible and reliable for state decision.

\subsection{Training Losses}
In previous Siamese trackers \cite{sbt}, \cite{transt}, classification and regression branches are usually optimized using two mutually independent losses, which maybe increase the misalignment probabilities of state decision. Actually, in addition to explore the above alignment module, it is also meaningful to train two branches in a cooperative way. Therefore, we adopt two cross-guided loss functions to optimize the proposed tracker. 

\subsubsection{Precision-guided classification loss}
For classification, if a sample with low regression precision still gets a pretty high classification score, which may defeat other candidates and be viewed as the tracking result, leading to the inferior performance. To avoid this situation, a visible solution is to take the regression precision as the weight of aggregating classification losses, making classification branch to pay more attention to high-precision samples:
\begin{equation}
\mathcal{L}_{p g_{-} c l s}=\frac{\sum_{i \in I_p} \frac{I o U\left(b_i, \widehat{b}\right)}{\overline{I o U}} \mathcal{L}_{c e}\left(y_i, p_i\right)+\beta \sum_{i \in I_n} \mathcal{L}_{c e}\left(y_i, p_i\right)}{N_p+\beta \cdot N_n}
\end{equation}
where, $\mathcal{L}_{ce}$ denotes the binary cross-entropy function, while $y_i$ and $p_i$ are the classification score and the binary ground-truth, respectively. $IoU$ depicts the Intersection over Union between the regression box $b_i$ and the ground-truth box $\widehat{b}$, and $\overline{I o U}$ is the average IoU ratio of all positive samples. $I_p$ or $I_n$ denotes the group of positive or negative samples, where $N_p$ and $N_n$ are the sample quantities in the corresponding groups. $\beta$ is the balancing factor, which is set as $0.0625$ in our implementation.

\subsubsection{Confidence-guided regression loss}
For regression, if a candidate has a high classification score, it is very important to lift its regression precision as much as possible, because it may be regard as the tracking outputs. In this case, we regard the classification confidences as the dynamic weights to compute regression loss:
\begin{equation}
\mathcal{L}_{c g_{-} r e g}=\frac{1}{N_p} \sum_{i \in I_p} \frac{y_i}{\bar{y}}\left(\lambda_1 \mathcal{L}_{\text {giou }}\left(b_i, \hat{b}\right)+\lambda_2 \mathcal{L}_1\left(b_i, \hat{b}\right)\right)
\end{equation}
in which, $\bar{y}$ is the average classification score of all positive samples. $\mathcal{L}_{\text {giou }}$ and $\mathcal{L}_1$ are generalized IoU loss and $l_1$-norm loss, respectively. $\lambda_1$ and $\lambda_2$ denotes the factors for balancing two kinds of losses, which are set to $2$ and $5$, respectively.

\section{Experiments and Results}
In this section, we first introduce the implementation details about offline optimization and online inference, and describe several popular benchmarks. Next, abundant experiments are conducted to test the performance of the presented associate prediction tracker, consisting of ablation studies, quantitative comparisons, qualitative comparisons, etc. Experimental results manifest that the proposed prediction network is more reliable and effective for classification-regression tracking.

\subsection{Implementation Details}
\subsubsection{Offline training}
The presented tracking model is optimized on the data splits of LaSOT \cite{lasot}, TrackingNet \cite{trackingnet}, GOT-10k \cite{GOT-10k} and COCO \cite{coco}. We extract a pair of template and search region samples directly from one video sequence or one still image using diverse data augmentations, whose sizes are set to $128 \times 128$ and $256 \times 256$ respectively, corresponding to $2^2$ and $4^2$ times of the object ground-truth area. The elements within the ground-truth box are labeled as positive samples, while the rest are viewed as negative samples. During optimization, the backbone is first initialized with the parameters pretrained on ImageNet-1k \cite{ImageNet}. The whole tracking model is trained 600 epochs using a AdamW optimizer with a weight decay of 1e-4, in which the iteration and the batch size are 1000 and 84, respectively. We set the initial learning rates as 1e-5 for backbone block, and 1e-4 for other components without initializing, all of which decrease 10 times per 400 epochs. Our network is implemented under Pytorch 1.9.1 on a server with two NVIDIA Tesla A100 GPUs.
\subsubsection{Online inference}
For inference, we first crop the template image in the initial frame and extract its features with backbone, which are kept fixed during tracking process for stability. In each subsequent frame, the search region image is extracted according to the object state in the previous time, whose features are compared with the template features by both category-aware and spatial-aware matchers. After aligning classification and regression similarity vectors, the prediction heads output the confidence scores and normalized coordinates of 1024 candidate elements. Following the assumption of smooth moving, Hanning window penalty is introduced to re-rank the confidence scores, where the penalty factor is set to 0.57. The element with the highest confidence is regarded as the tracking result.

\subsection{Benchmarks and Metrics}
The proposed Siamese tracker is evaluated on five public benchmark datasets, consisting of LaSOT, TrackingNet, GOT-10k, TNL2k and UAV123. Among these, LaSOT \cite{lasot} is a recent large-scale benchmark composed of 280 full-annotated testing sequences, which cover 70 different kinds of objects. The average length of these videos is more than 2500 frames, which is a great challenge to short-term trackers. In addition, the dataset contains 14 types of challenging scenarios, i.e., illumination variation, scale variation, background clutter, etc. TrackingNet \cite{trackingnet} is a recent-released high-diversity dataset, including a large number of short-term sequences collected in the wild environments. For GOT-10k \cite{GOT-10k}, there are 180 sequences in the testing set. To ensure the equality, all participants should be optimized only using its training set, whose object classes have no overlap with the testing set. For the widely-used TNL2k dataset \cite{tnl2k}, it provides 700 challenging video sequences with diverse interference factors. UAV123 \cite{uav} is a typical aerial benchmark, which consists of 123 sequences captured from low-attitude unmanned aerial vehicles.

In the evaluation protocols of the above benchmarks, all of metrics are computed based on center location error and overlap ratio. The former is the pixel distance between the predicted and the ground-truth object centers, while the latter is the Intersection over Union (IoU) of the predicted and the ground-truth bounding boxes. On UAV123 dataset, Success Rate (SR) and Precision Rate (NR) are utilized to evaluate trackers. SR denotes the Area Under Curve (AUC) of success plot which shows the ratios of images when the overlap ratios are larger than a given threshold. NR is the percentage of images when the distance errors are within a given threshold, which is usually set to 20 pixels. For benchmarks of LaSOT, TrackingNet and TNL2k, in addition to SR and NR, Normalized Precision Rate (NPR) is also adopted to quantify tracking performance, which is not sensitive to the image resolution and target size. For GOT-10k, the average overlap rate (AO) and the Success Rates (SR) on two fixed thresholds of 0.5 and 0.75 are employed as evaluation metrics. 

\begin{table}[t]
	\caption{Ablation studies about network components on LaSOT dataset, in which \emph{Base}, \emph{C.A} and \emph{S.A} denote the base, the category-aware, and the spatial-aware matchers, respectively. The best results are highlighted in \textcolor[rgb]{1,0,0}{red} fonts.}	\label{first-tab}	
	\begin{center}
		\setlength{\tabcolsep}{2.5mm}{		
			\begin{tabular}{l|cc|c|cc}
				\toprule[1.5pt] 
				$\#$ & Classification & Regression & Alignment & SR$\uparrow$ & NPR$\uparrow$\\
				\midrule[1pt] 
				1  & \multicolumn{2}{c}{\emph{Base}(shared)} & -- & 0.632 & 0.716\\
				2  & \emph{Base} & \emph{Base} & \ding{53} & 0.641 & 0.723\\
				3  & \emph{Base} & \emph{S.A} & \ding{53} & 0.646 & 0.731\\
				4  & \emph{C.A} & \emph{S.A} & \ding{53} & 0.652 & 0.737\\
				\midrule[1pt] 
				5  & \emph{C.A} & \emph{S.A} & \checkmark & \textcolor[rgb]{1,0,0}{0.661} & \textcolor[rgb]{1,0,0}{0.749}\\
				\bottomrule[1.5pt]
		\end{tabular}}
	\end{center}
\end{table}

\subsection{Ablation Studies}
\subsubsection{Network components}
Initially, a base matcher is designed by combining two self-attentions and one cross-attention. Its structure is similar to our presented matcher in Fig. \ref{third-fig}, and the only difference is that there are no channel or spatial attentions to execute feature selection. Next, we construct four ablation variants to manifest the necessity of exploring our feature matchers and feature alignment module, i.e.,\emph{Variant \#1}, \emph{\#2}, \emph{\#3} and \emph{\#4}. In detail, classification and regression branches share several base matchers in \emph{Variant \#1}, in which the matching results are adopted by classification and regression heads simultaneously. In contrast, \emph{Variant \#2} embeds two base matching blocks into classification and regression branches, respectively. In \emph{Variant \#3} and \emph{\#4}, the base matchers in two branches are gradually replaced by our category-aware and spatial-aware matchers. \emph{Variant \#5} furtherly incorporates the proposed dual alignment module, formatting the MAPNet-R tracker.

The tracking results of these variations are shown in Table \ref{first-tab}. Compared to \emph{Variant \#1}, \emph{Variant \#2} obtains great increments of 0.9\% on Success and 0.7\% on Normalized precision, which demonstrates that it is meaningful to deploy diverse matching blocks in two prediction branches. In addition, by comparing \emph{Variant \#3}, \emph{Variant \#4} with \emph{Variant \#2}, we observe that the proposed category-aware and spatial-aware matchers are more effective for classification and regression, respectively. As last, \emph{Variant \#5}, i.e., MAPNet-R tracker, surpasses \emph{Variant \#4} by 0.9\% on Success and 1.2\% on Normalized precision, declaring that the dual alignment module is very important to improve tracking performance.
\begin{table}[h]
	\caption{Ablation studies about the quantities of matchers on LaSOT dataset, in which both performance and speed are considered.The best results are highlighted in \textcolor[rgb]{1,0,0}{red} fonts.}	\label{second-tab}	
	\begin{center}
		\setlength{\tabcolsep}{4.5mm}{		
			\begin{tabular}{l|cc|c}
				\toprule[1.5pt] 
				$\#$  & SR$\uparrow$ & NPR$\uparrow$ & FPS $\uparrow$\\
				\midrule[1pt] 
				1  & 0.643 & 0.724 & \textcolor[rgb]{1,0,0}{33.4}\\
				2  & 0.655 & 0.740 & 29.2\\
				3  & 0.661 & 0.749 & 25.7\\
				4  & \textcolor[rgb]{1,0,0}{0.663} & \textcolor[rgb]{1,0,0}{0.752}&22.3\\
				\bottomrule[1.5pt]
		\end{tabular}}
	\end{center}
\end{table}  
\subsubsection{Quantities of feature matchers}
The number of matchers directly influences the capability of the prediction network. We implement the network with diverse quantities of matchers and compare their performance in Table \ref{second-tab}.  It is rational that the tracking performance improves along with the increments of feature matchers. However, in contrast to employ 3 feature matchers, it does not lift tracking performance obviously when introducing 4 matchers, and the speed is not real-time. Hence, we utilize 3 feature matchers for each matching module in our prediction network.

\begin{table}[t]
	\caption{Ablation studies about diverse classification and regression losses on LaSOT dataset, in which $\mathcal{L}_{ce}$, $\mathcal{L}_{\text {giou }}$ and $\mathcal{L}_{1}$ denotes binary cross-entropy loss, generalized IoU loss and $l_1$-norm loss, respectively.The best results are highlighted in \textcolor[rgb]{1,0,0}{red} fonts.}	\label{thrid-tab}	
	\begin{center}
		\setlength{\tabcolsep}{2.5mm}{		
			\begin{tabular}{l|cc|cc}
				\toprule[1.5pt] 
				$\#$ & Classification & Regression & SR$\uparrow$ & NPR$\uparrow$\\
				\midrule[1pt] 
				1  & $\mathcal{L}_{ce}$ & $\mathcal{L}_{1}+\mathcal{L}_{\text {giou }}$ & 0.650 & 0.736\\
				2  & $\mathcal{L}_{p g_{-} c l s}$ & $\mathcal{L}_{1}+\mathcal{L}_{\text {giou }}$ &  0.655 & 0.744\\
				3  & $\mathcal{L}_{ce}$ & $\mathcal{L}_{c g_{-} r e g}$  & 0.656 & 0.741\\
				4  & $\mathcal{L}_{p g_{-} c l s}$ & $\mathcal{L}_{c g_{-} r e g}$  & \textcolor[rgb]{1,0,0}{0.661} & \textcolor[rgb]{1,0,0}{0.749}\\
				\bottomrule[1.5pt]
		\end{tabular}}
	\end{center}
\end{table}
\subsubsection{Optimization losses}
In this part, we train the presented MAPNet-R tracker with diverse combinations of classification and regression losses, and report the tracking results in Table \ref{thrid-tab}. It is easy to find that the proposed precision-guided classification loss ($\mathcal{L}_{p g_{-} c l s}$)  and confidence-guided regression loss ($\mathcal{L}_{c g_{-} r e g}$) are more appropriate for optimizing classification-regression tracking model. Compared with \emph{Combination \#1}, they lift the performance by 1.1\% on Success and 1.3\% on Normalized precision.
\begin{figure*}[t]
	\begin{center}
		\includegraphics[width=0.4\linewidth]{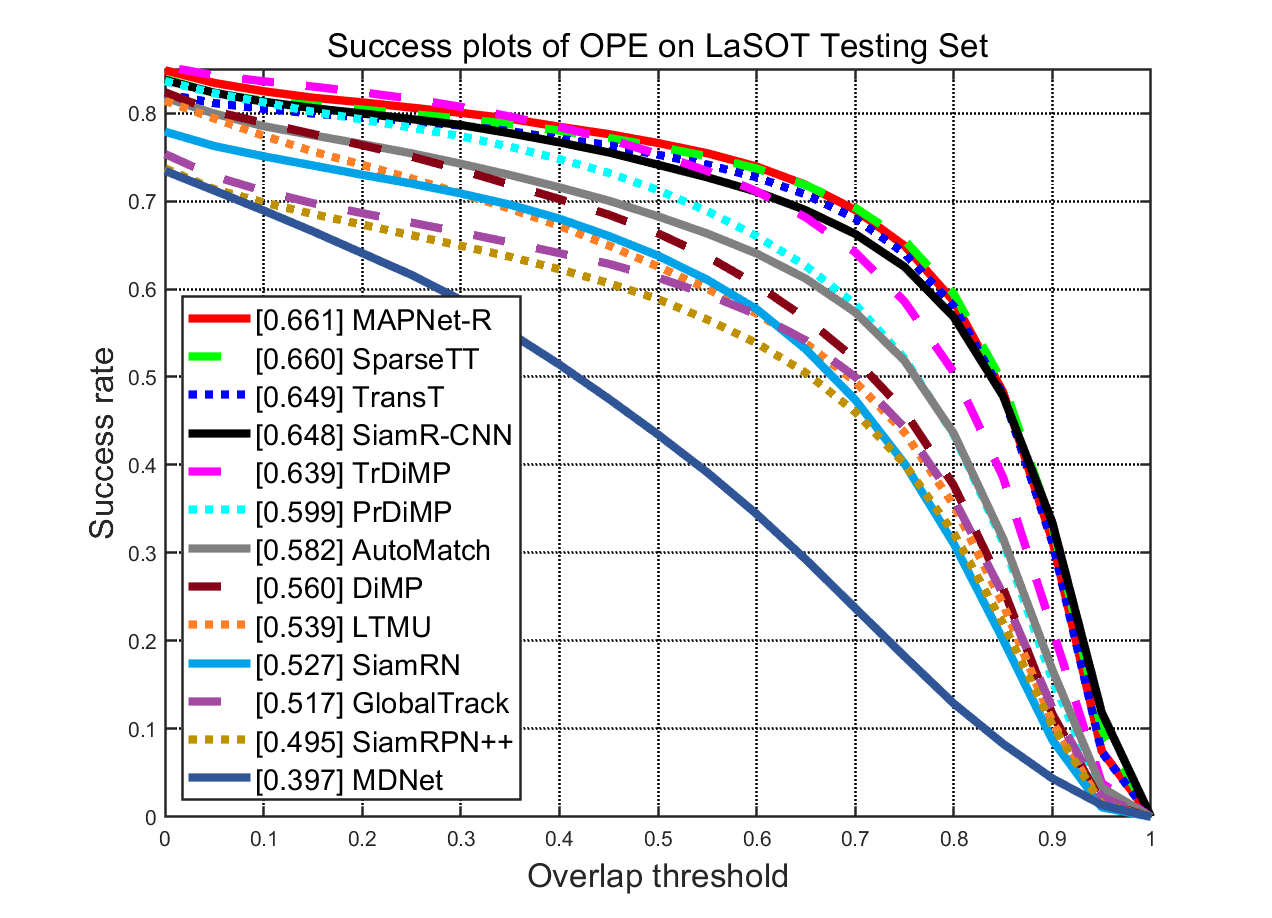}
		\includegraphics[width=0.4\linewidth]{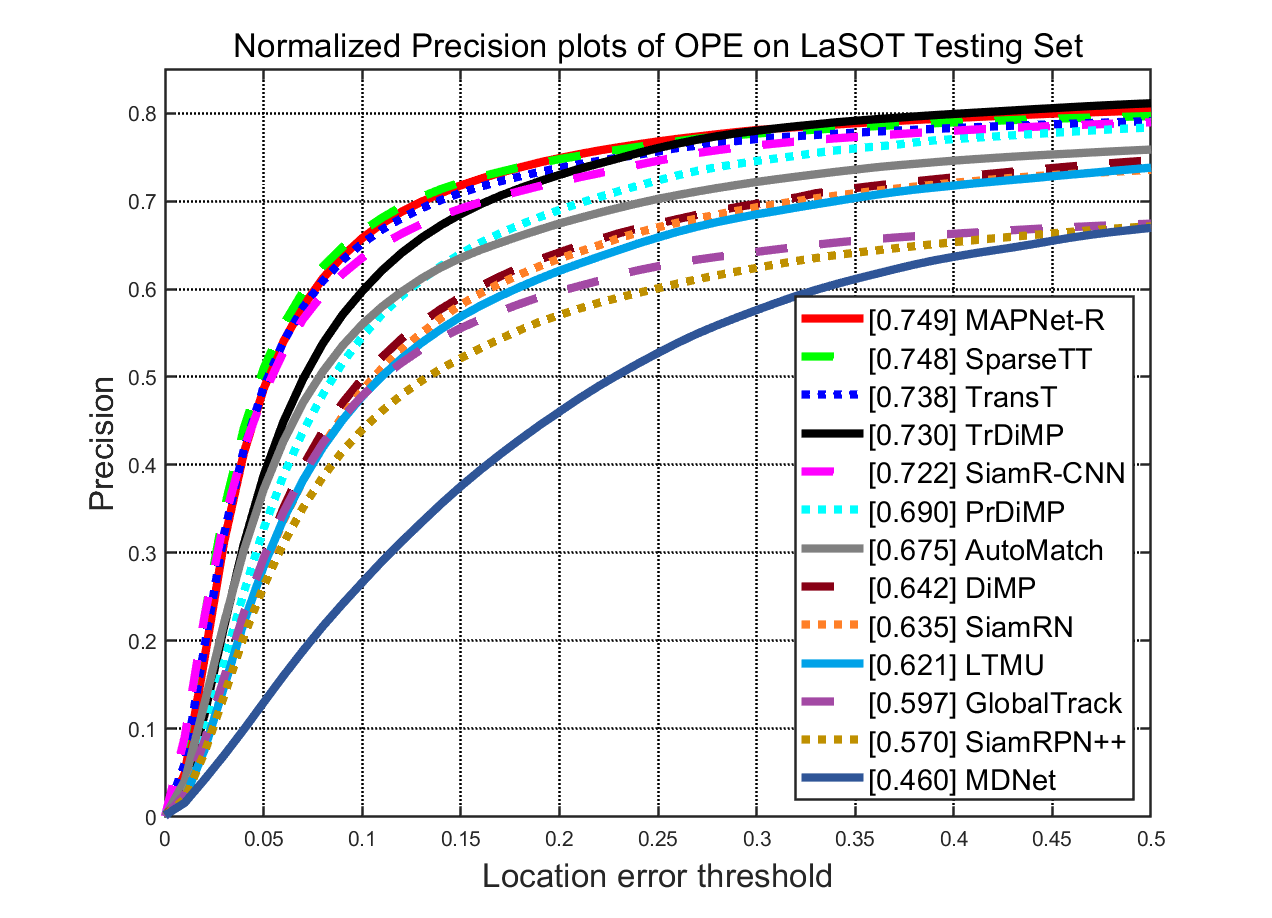}
	\end{center}
	\caption{Success and Normalized precision plots of all trackers in OPE formulation on LaSOT. These trackers are ranked according to their performance scores.}
	\label{fifth-fig}
\end{figure*}

\subsection{Quantitative Comparisons }
\subsubsection{LaSOT}
On this benchmark, we compare the proposed tracker, i.e., MAPNet-R, with twelve state-of-the-art algorithms, including SparseTT \cite{sparsett}, TransT \cite{transt}, TrDiMP \cite{trsiam}, SiamR-CNN \cite{siamr-cnn}, PrDiMP \cite{prdimp}, AutoMatch \cite{automatch}, DiMP \cite{dimp}, GlobalTrack \cite{globaltrack}, LTMU \cite{ltmu}, SiamRN \cite{siamrn}, SiamRPN++ \cite{SiamRPN++} and MDNet \cite{mdnet}. The overall Success and Normalized precision plots of these methods are displayed in Fig. 5. We observe that our network ranks first with the highest Success and Normalized precision scores of 66.1\% and 74.9\%. Compared to the typical TransT model \cite{transt}, the proposed work exceeds it by 1.2\% on Success and 1.1\% on Normalized precision, although it designed a large-scale matching network for  classification and regression with more self-attentions and cross-attentions. In addition, our method is superior to another outstanding tracker of TrDiMP \cite{trsiam} by 2.2\% on Success and 1.9\% on Normalized precision.

\begin{figure*}[t]
	\begin{flushleft}
		\includegraphics[width=0.245\linewidth]{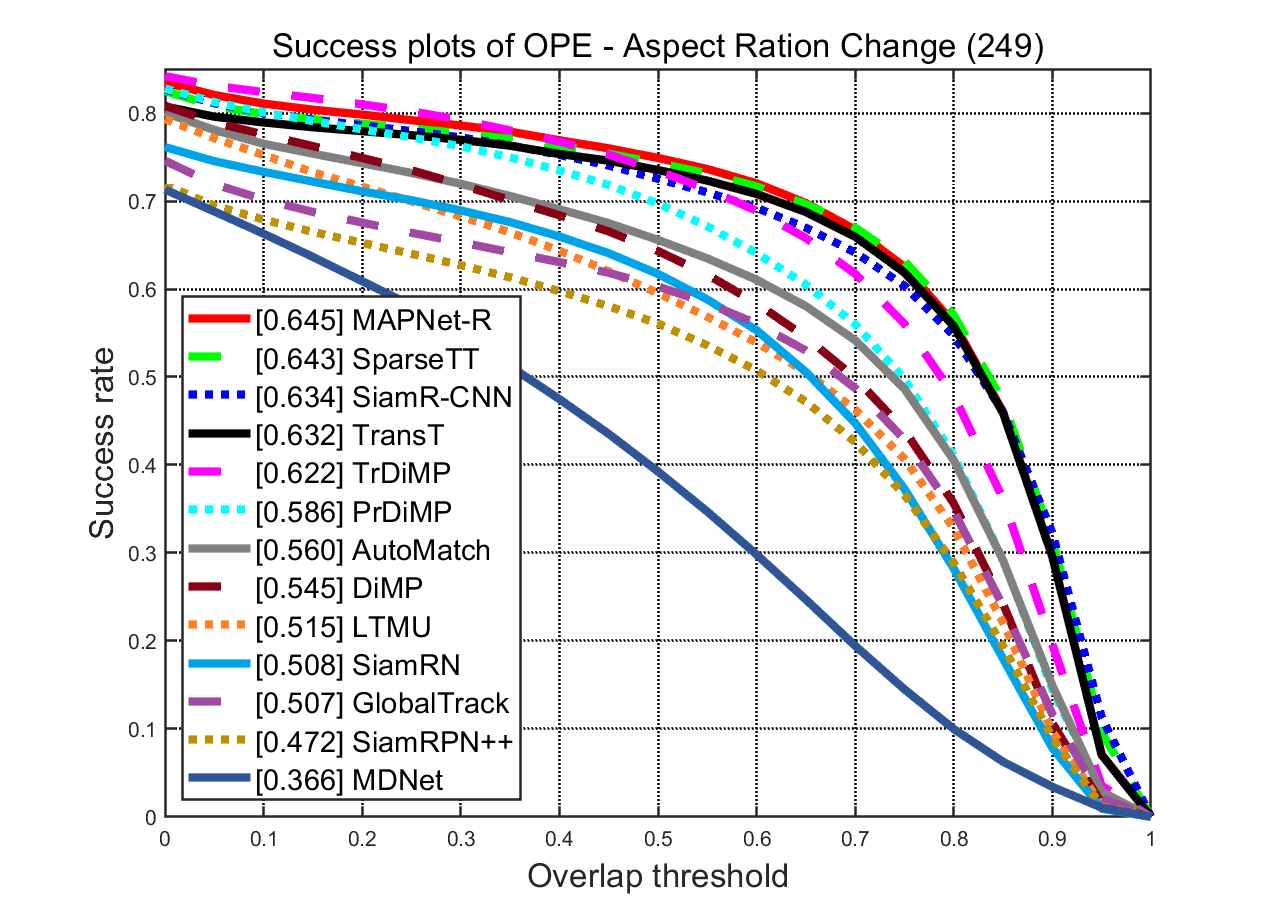}
		\includegraphics[width=0.245\linewidth]{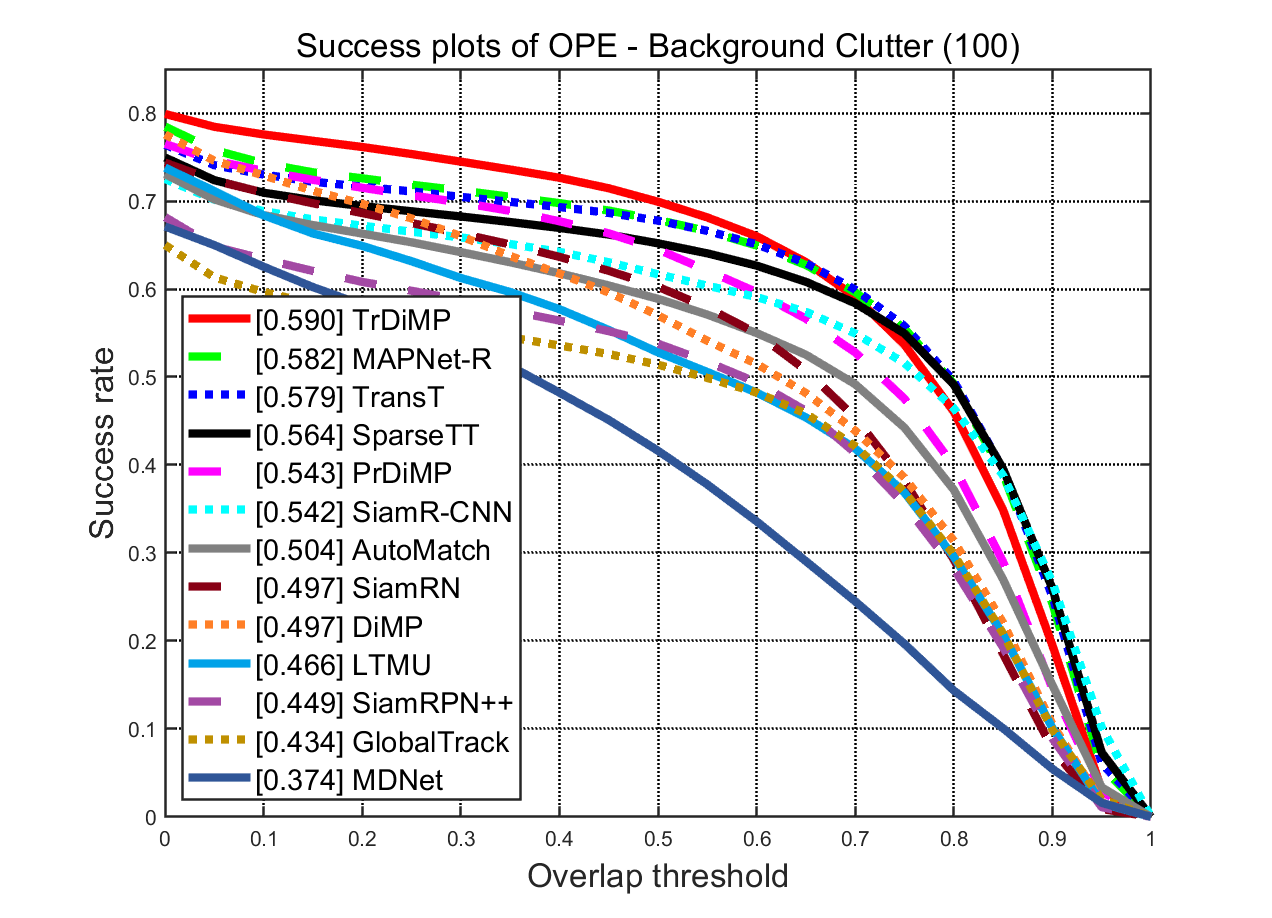}
		\includegraphics[width=0.245\linewidth]{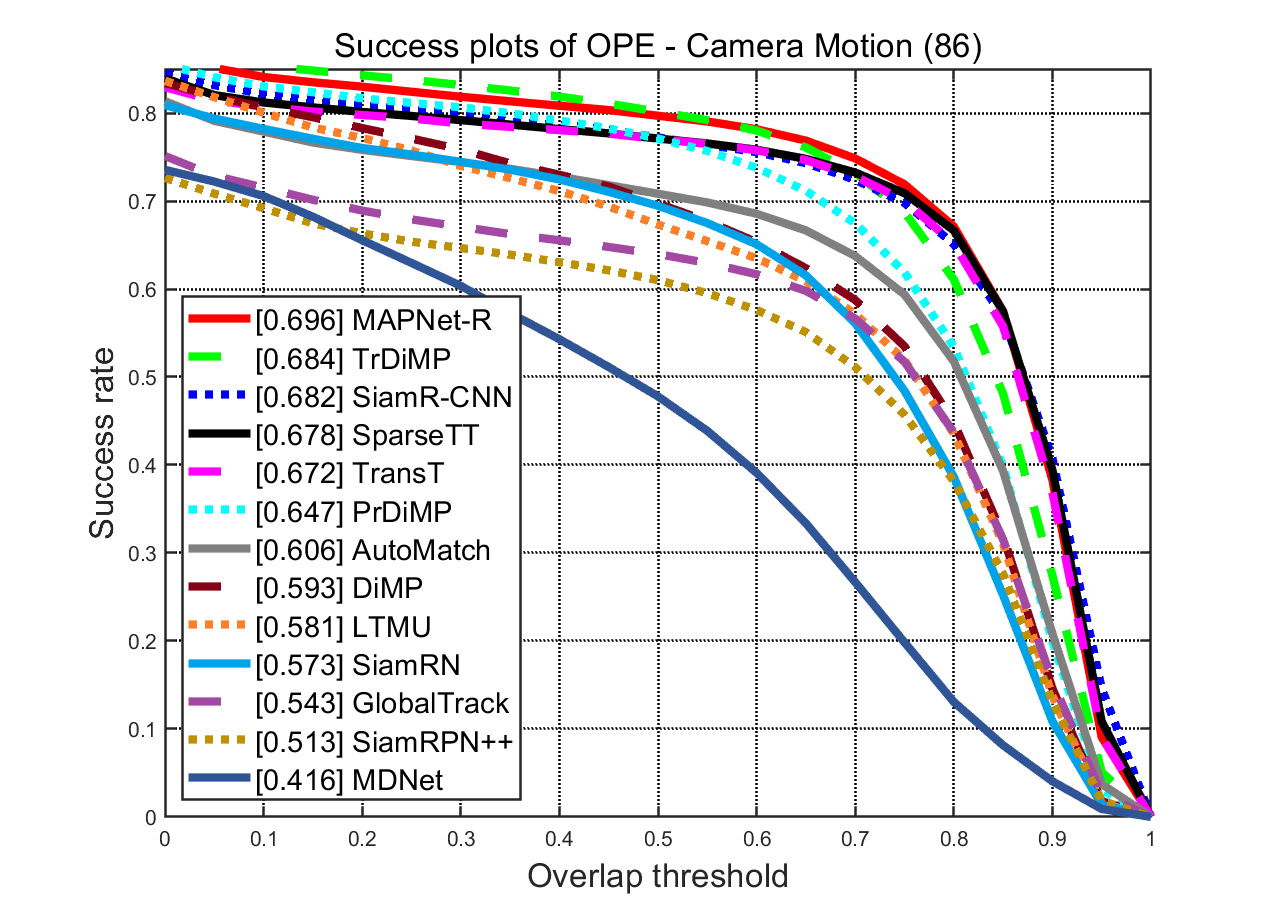}
		\includegraphics[width=0.245\linewidth]{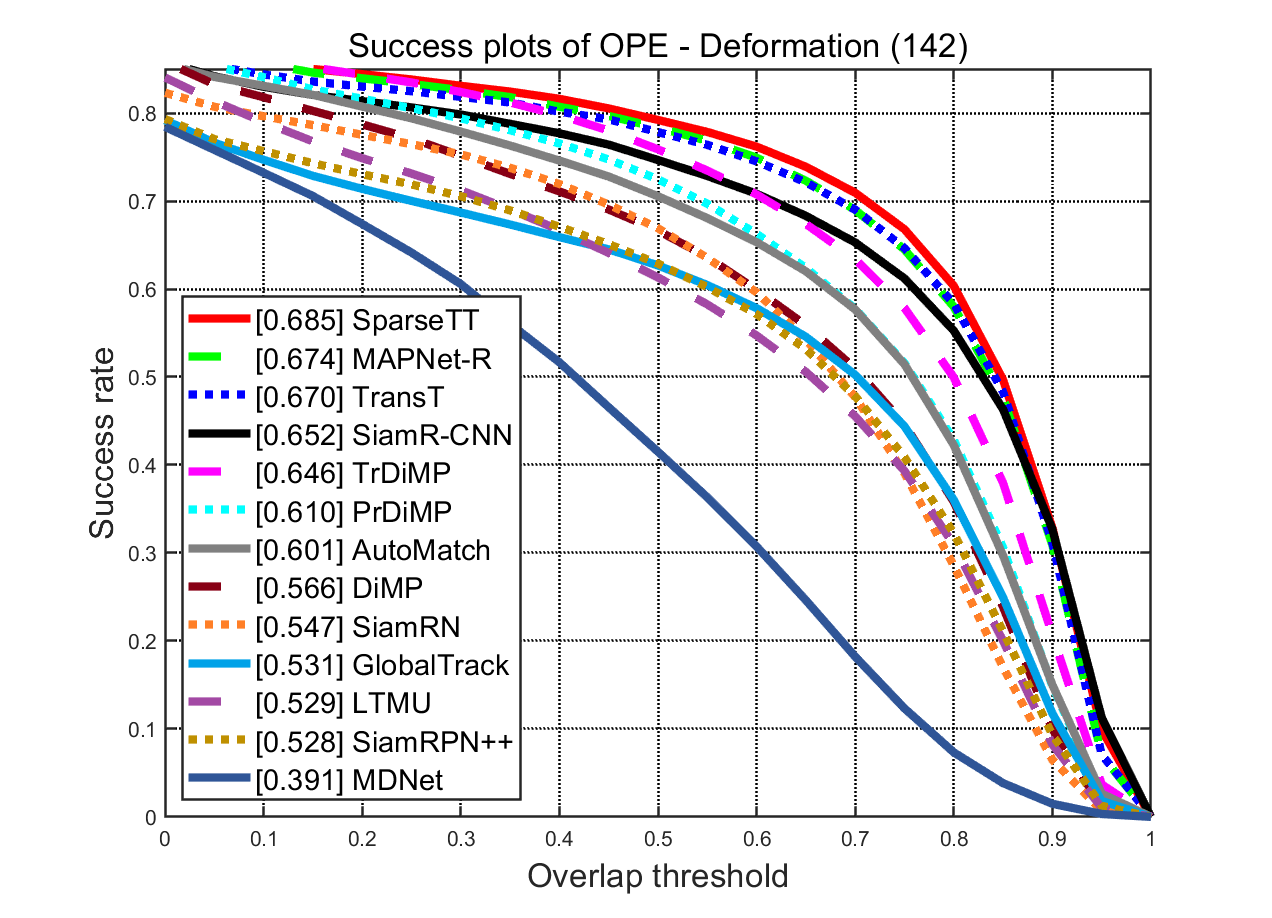}
		
		\includegraphics[width=0.245\linewidth]{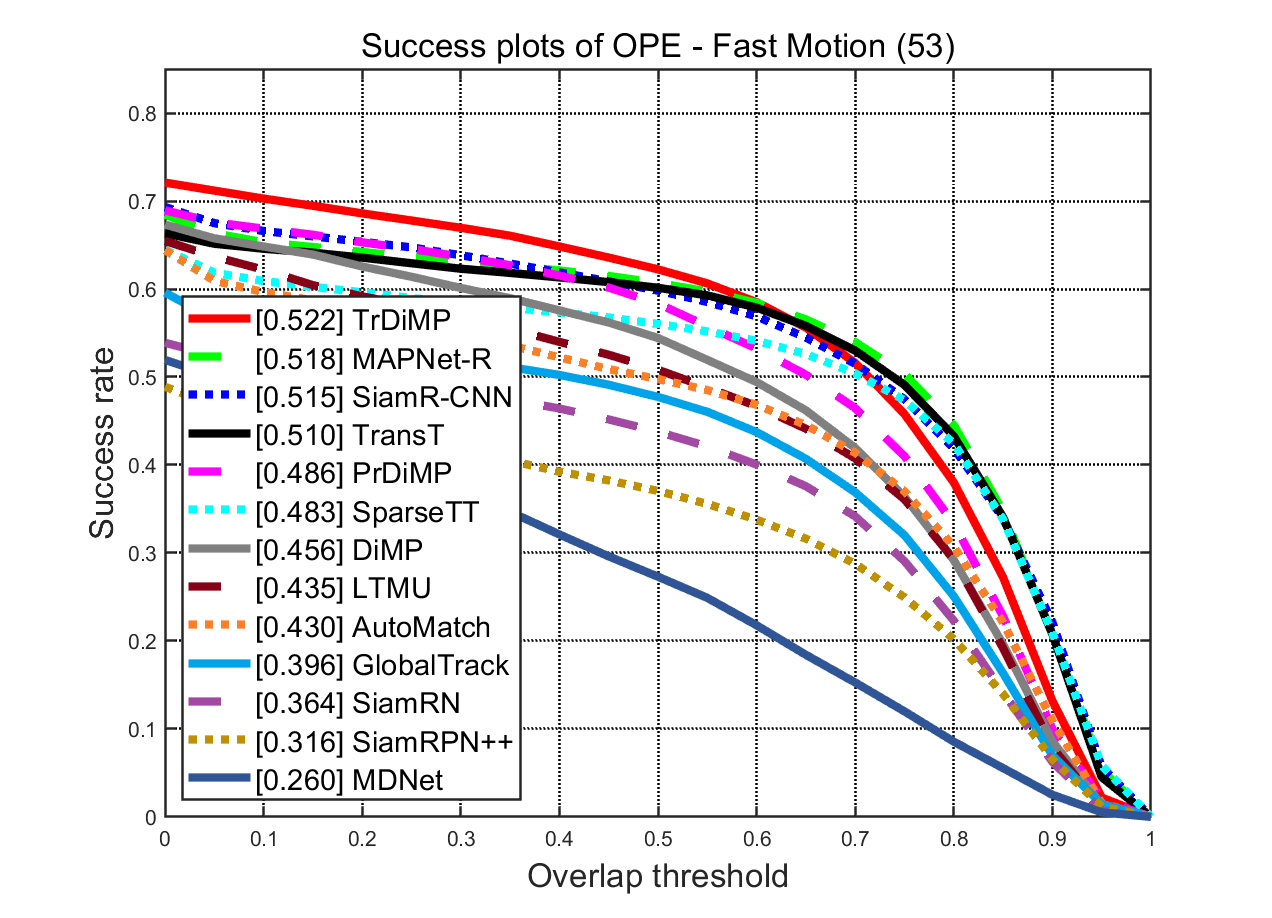}
		\includegraphics[width=0.245\linewidth]{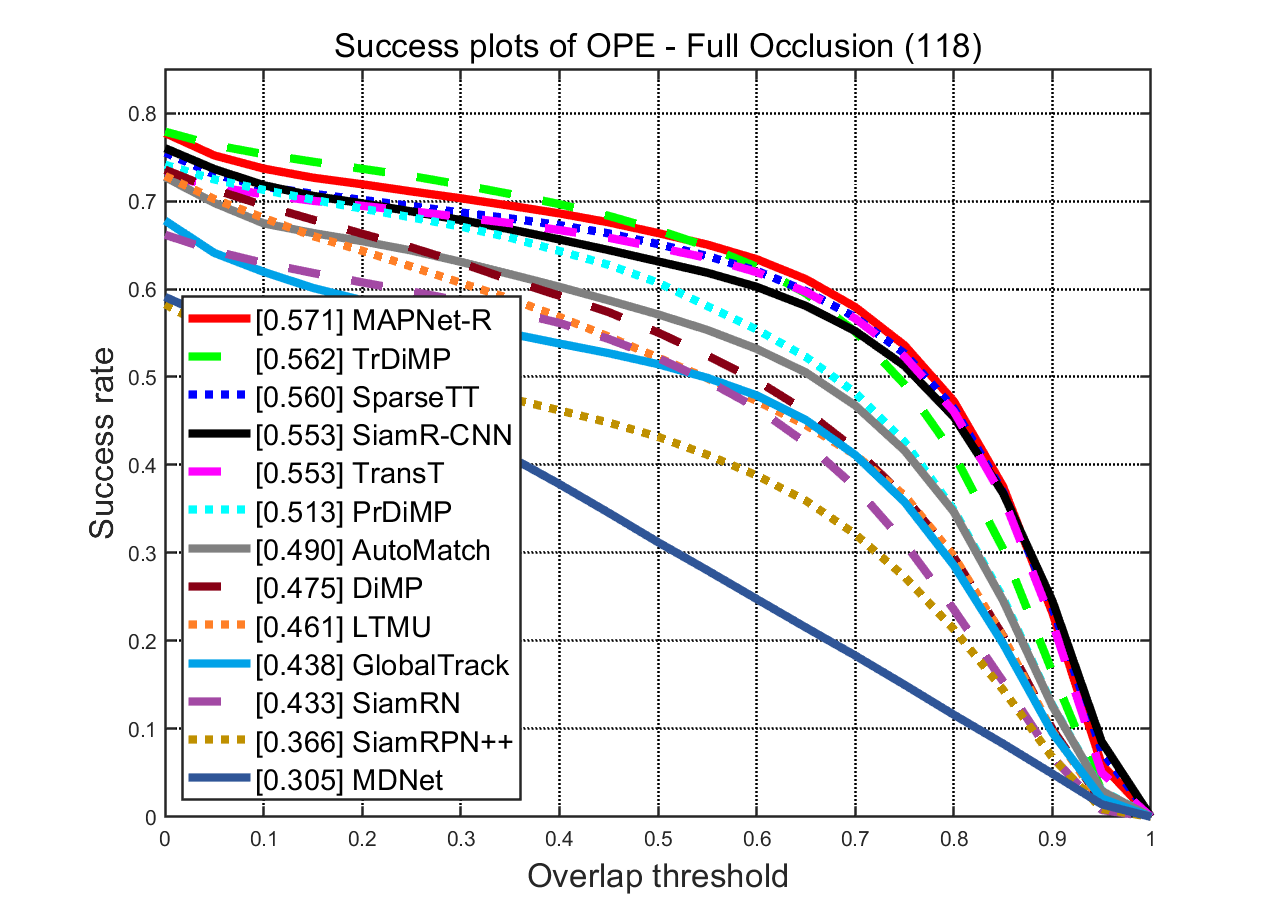}
		\includegraphics[width=0.245\linewidth]{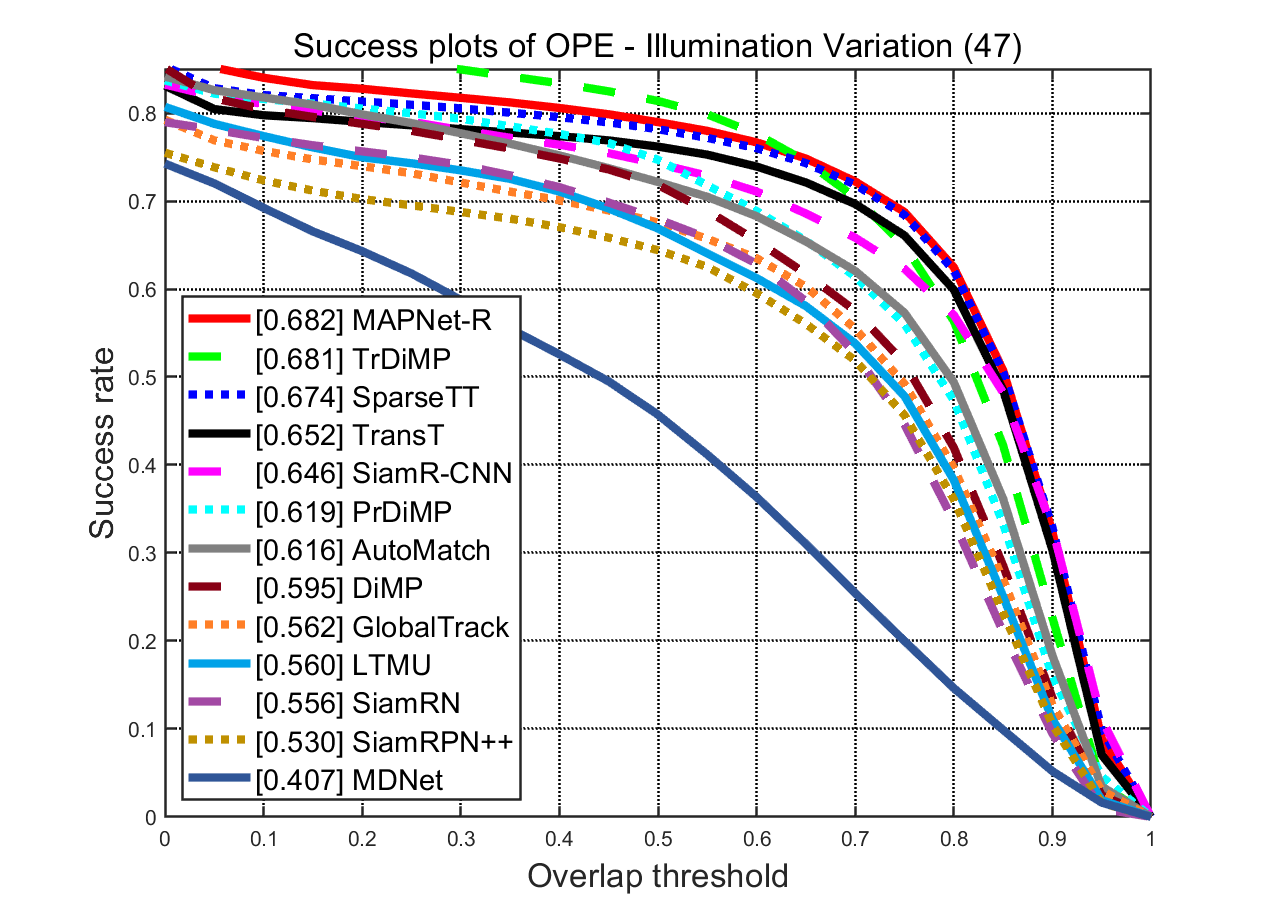}
		\includegraphics[width=0.245\linewidth]{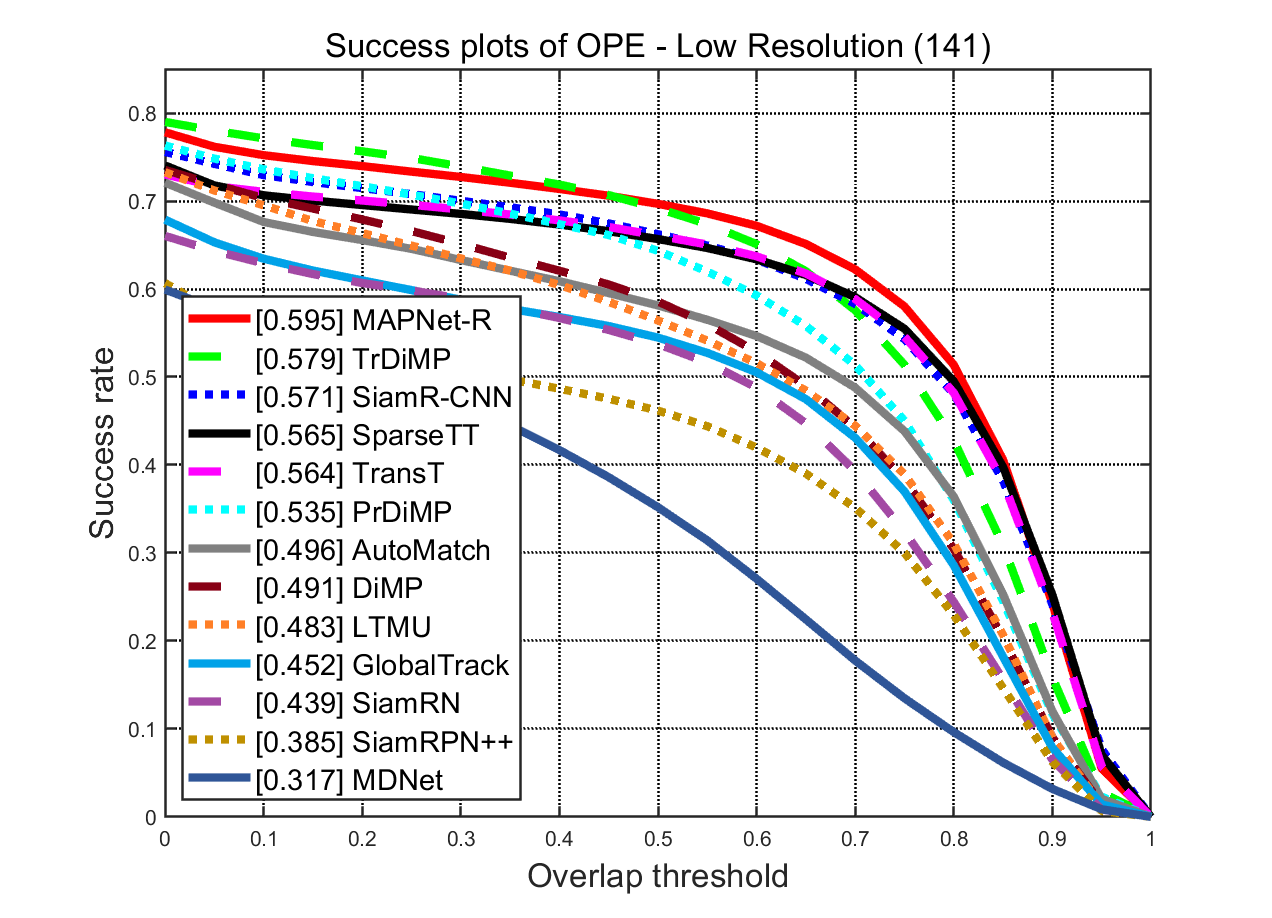}
		
		\includegraphics[width=0.245\linewidth]{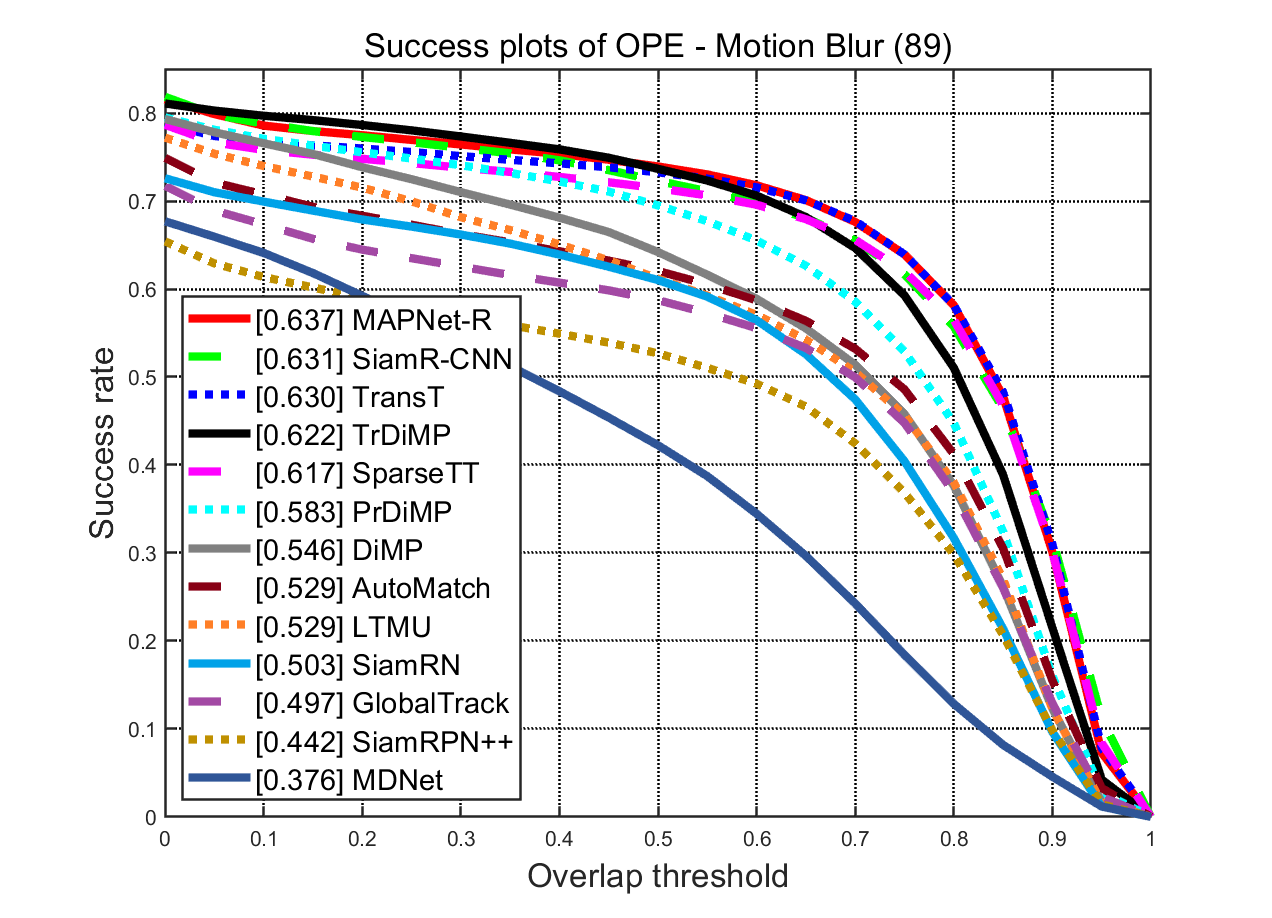}
		\includegraphics[width=0.245\linewidth]{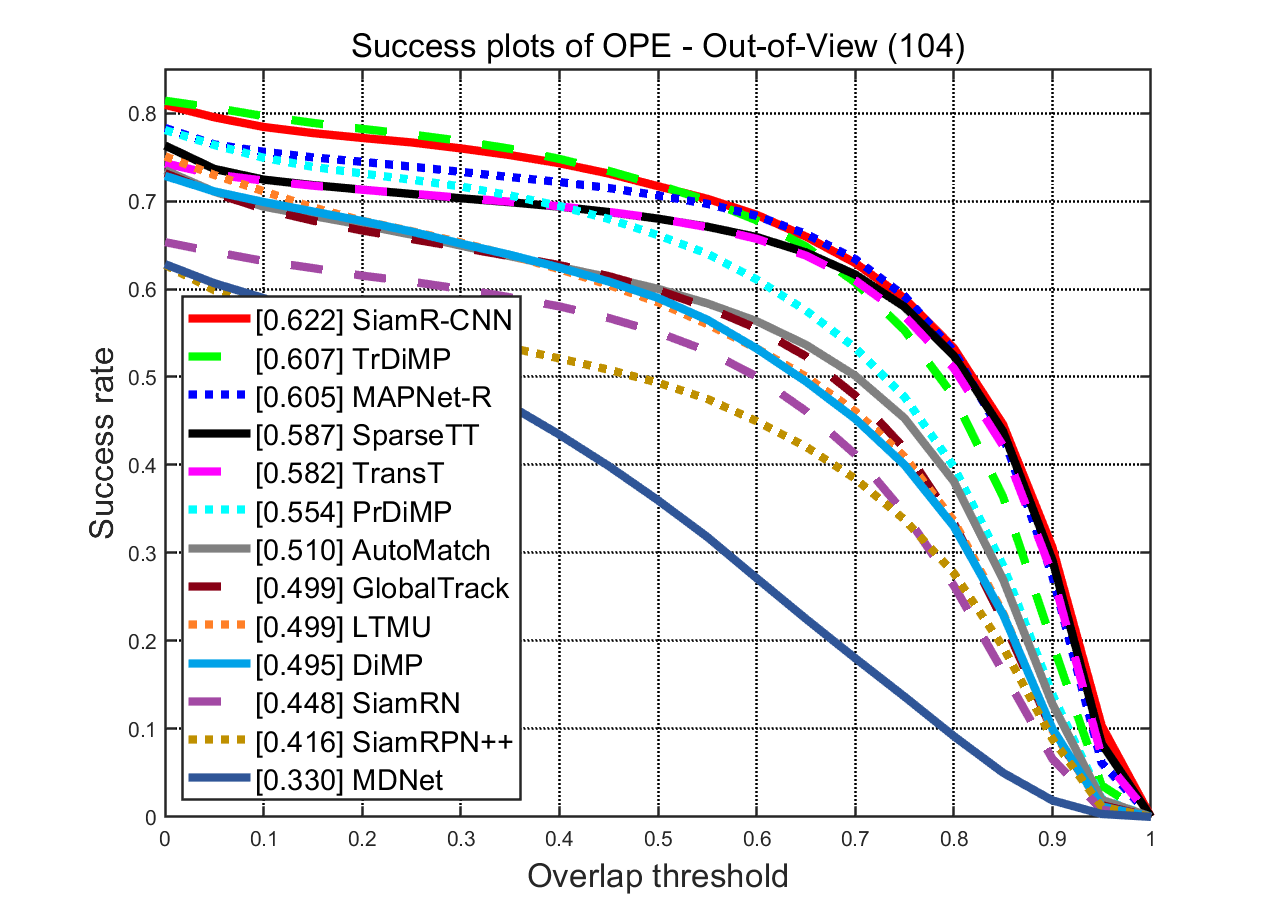}
		\includegraphics[width=0.245\linewidth]{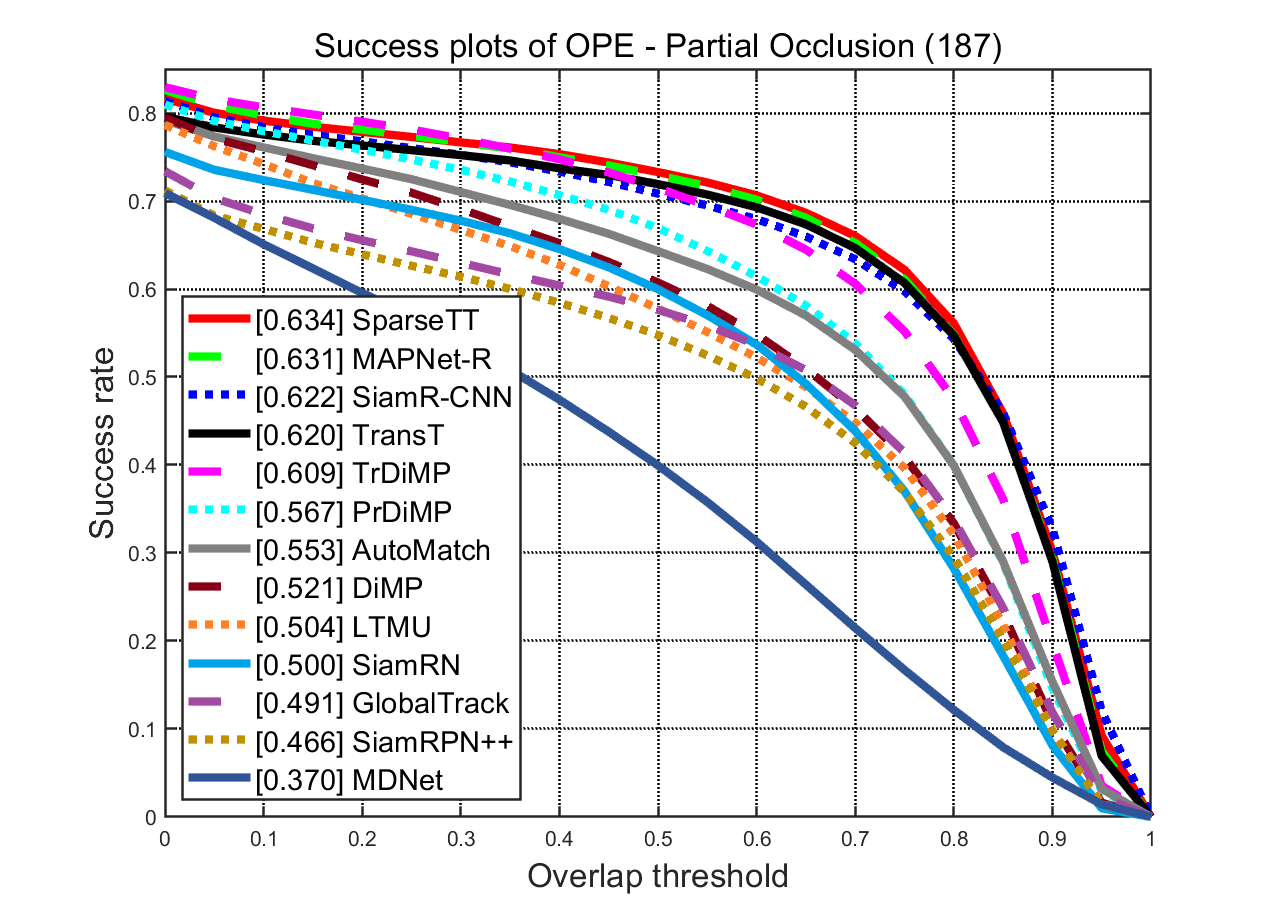}		
		\includegraphics[width=0.245\linewidth]{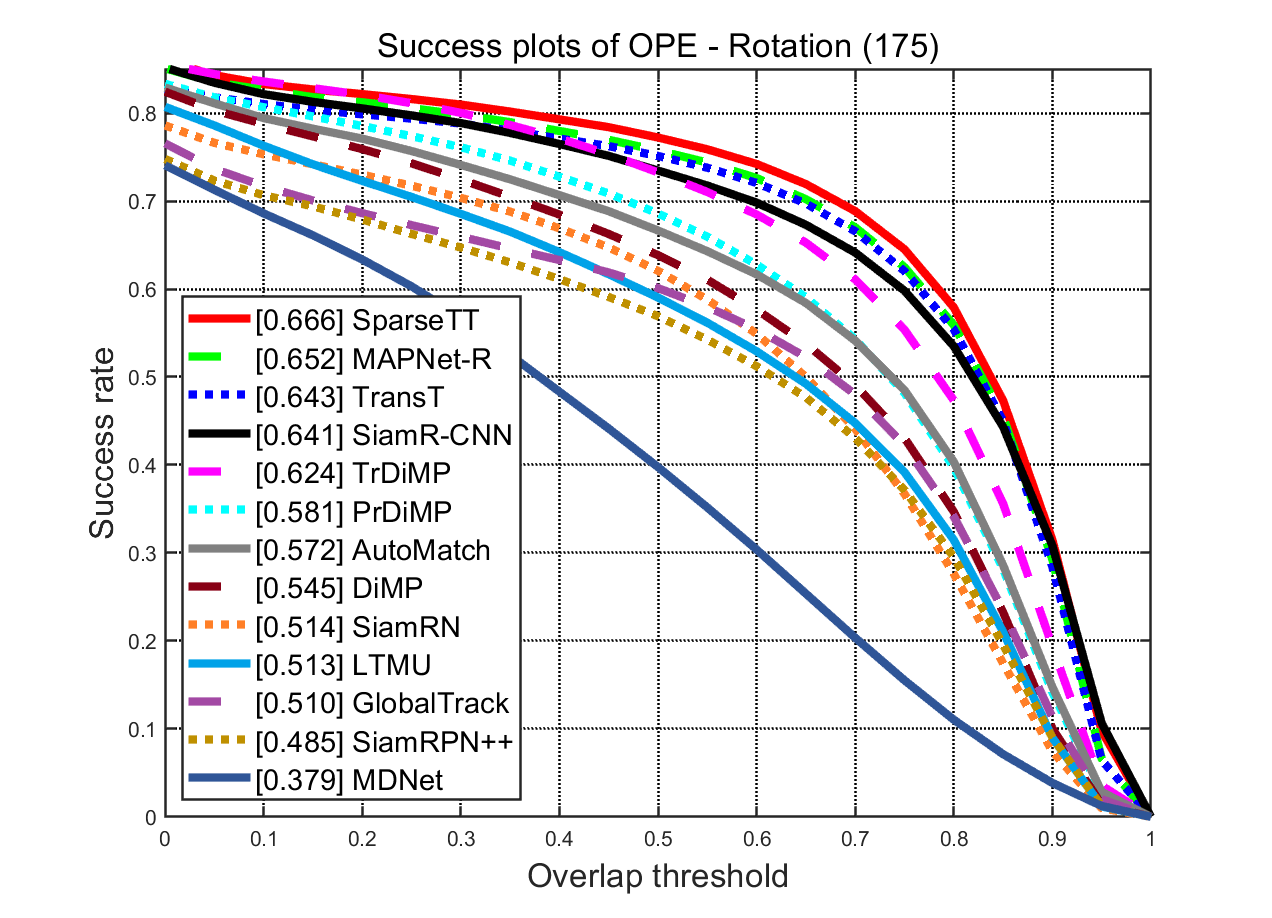}
		
		\includegraphics[width=0.245\linewidth]{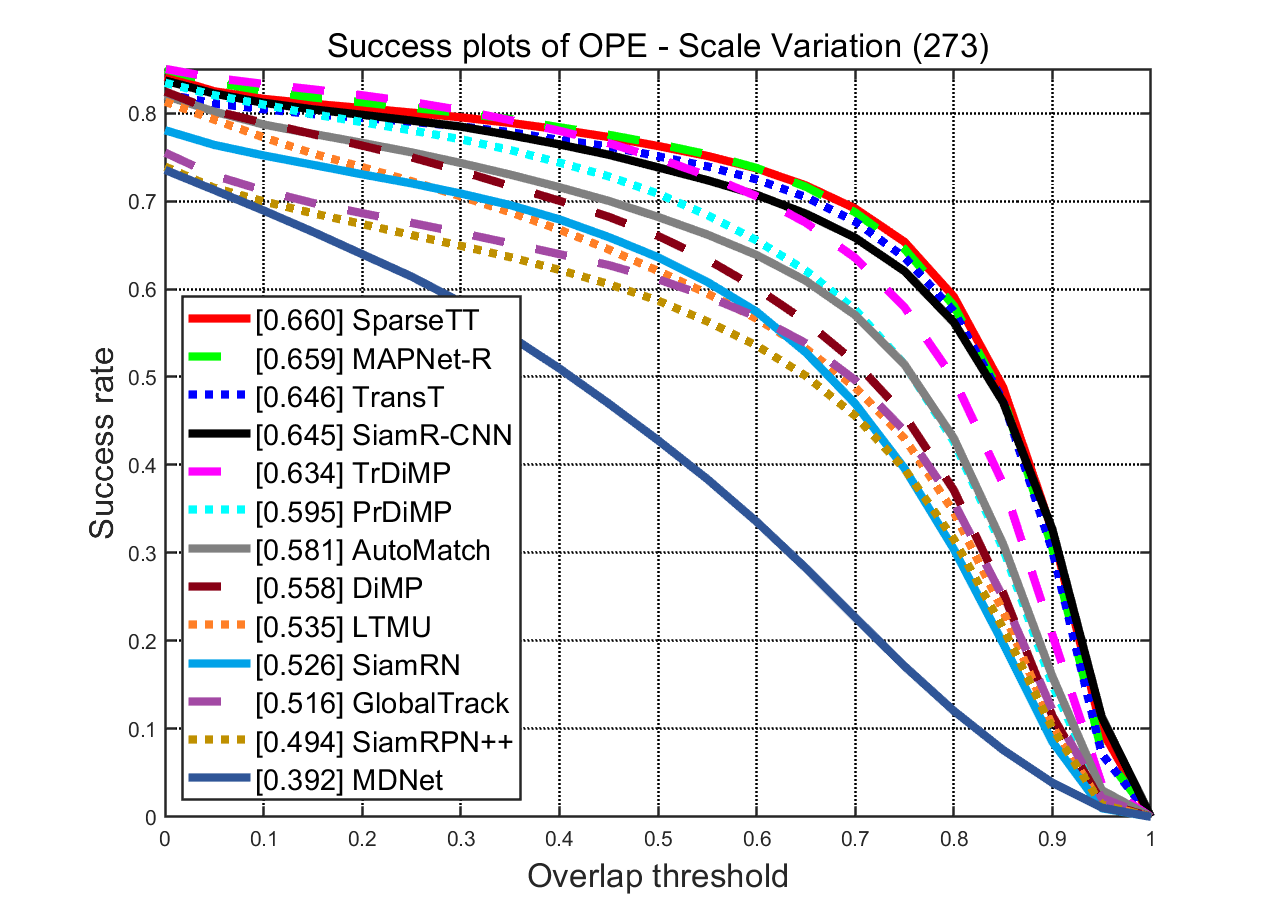}
		\includegraphics[width=0.245\linewidth]{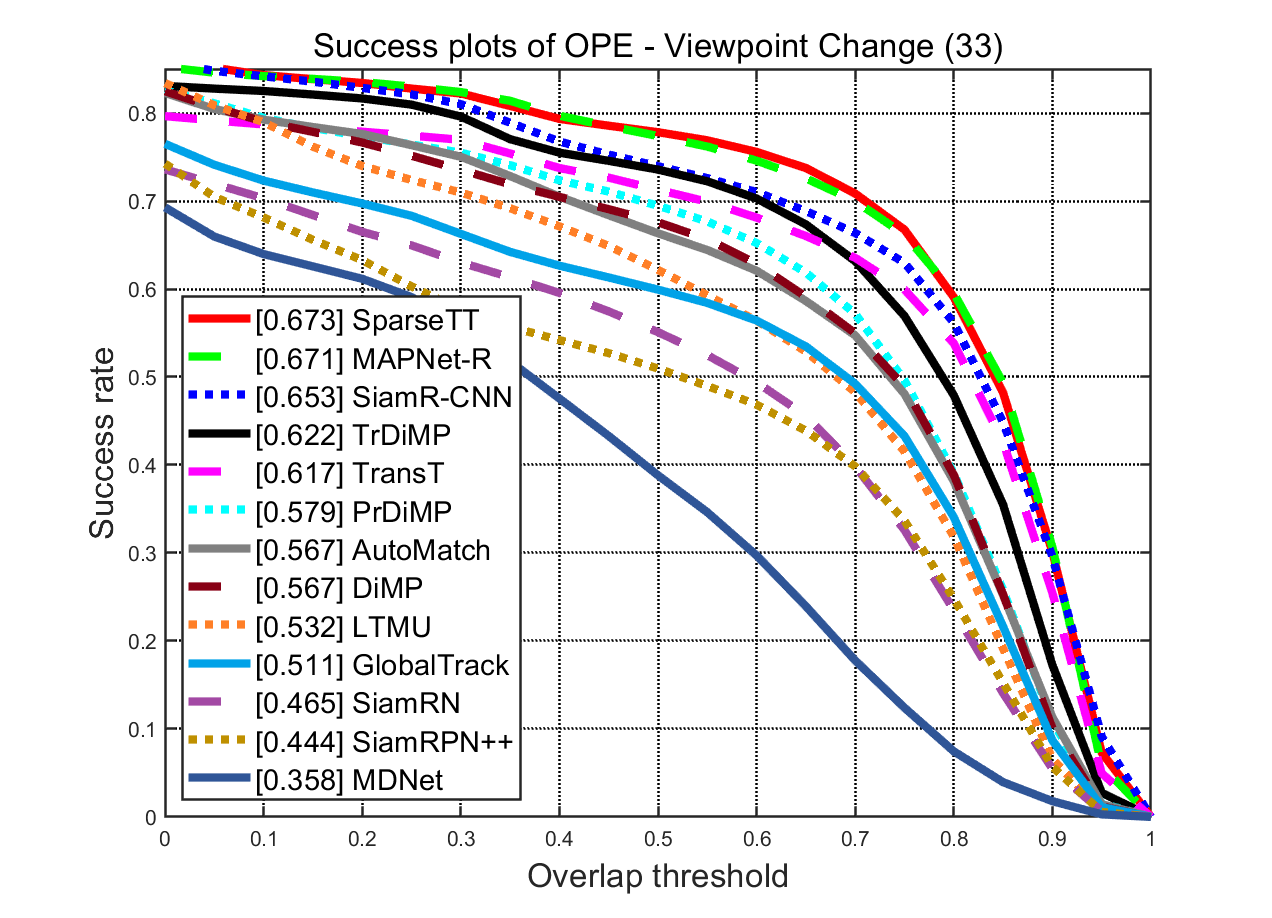}
	\end{flushleft}
	\caption{Success plots of different attributes in OPE formulation on LaSOT. The number in the parenthesis denotes the number of sequences within the attribute. All comparison methods are ranked according to their success scores.}
	\label{sixth-fig}
\end{figure*}
 
To demonstrate the concrete performance of the presented prediction network, we also provide the Success plots of all trackers on 14 kinds of challenging attributes, as shown in Fig. \ref{sixth-fig}. These plots manifest that our MAPNet-R tracker is able to achieve satisfactory tracking results on all of attributes, which yields the best performance on 6 diverse attributes in term of Success. Especially on the attributes of Camera Motion (CM), Full Occlusion (FO) and Low Resolution (LR), our approach obtains nearly or more than 1.0\% improvements on Success compared to the second-ranked algorithms. These phenomena declare that MAPNet-R is stronger in different complicated scenarios, proving that our prediction network is efficient to extract more sufficient and suitable similarity maps for both category classification and coordinate regression.

\begin{table}[t]
	\caption{Comparison with state-of-the-art trackers on TrackingNet. The best three results are highlighted in \textcolor[rgb]{1,0,0}{red}, \textcolor[rgb]{0,1,0}{green} and \textcolor[rgb]{0,0,1}{blue} fonts.}	\label{fourth-tab}	
	\begin{center}
		\setlength{\tabcolsep}{2.5mm}{		
			\begin{tabular}{lccc}
				\toprule[1.5pt] 
				Trackers & SR$\uparrow$ & NPR$\uparrow$ & 	PR$\uparrow$ \\
				\midrule[1pt] 
				SiamFC \cite{siamfc} & 0.571 & 0.663 & 0.533 \\
				Ocean \cite{ocean} &	0.692 &	0.794 &	0.687 \\
				SiamRPN++ \cite{SiamRPN++} &  0.733 & 0.800 & 0.694 \\
				DiMP	\cite{dimp} & 0.740 & 0.801 & 0.687 \\
				AutoMatch \cite{automatch} &	0.760 &	-- &	0.726 \\
				PrDiMP \cite{prdimp} &	0.758 &	0.816 &	0.704 \\
				KeepTrack \cite{keeptrack} & 0.781 &	0.835 &	0.738 \\
				TREG \cite{treg} & 0.785 & 0.838 & 0.750 \\
				TrDiMP \cite{trsiam} & 0.784 & 0.833 & 0.731 \\
				Stark \cite{stark} & \textcolor[rgb]{0,1,0}{0.820} & \textcolor[rgb]{1,0,0}{0.869} & \textcolor[rgb]{0,1,0}{0.791} \\
				UTT \cite{utt} & 0.797 & -- & 0.770 \\
				ToMP \cite{tomp} & \textcolor[rgb]{0,0,1}{0.815} & \textcolor[rgb]{0,1,0}{0.864} & \textcolor[rgb]{0,0,1}{0.789} \\				
				\midrule[1pt] 
				MAPNet-R  & \textcolor[rgb]{1,0,0}{0.823} &  \textcolor[rgb]{0,1,0}{0.864} & \textcolor[rgb]{1,0,0}{0.796} \\
				\bottomrule[1.5pt]
		\end{tabular}}
	\end{center}
\end{table} 
\subsubsection{TrackingNet} 
The presented tracking method is evaluated on the dataset by comparing it with other popular participants, consisting of ToMP \cite{tomp}, UTT \cite{utt}, Stark \cite{stark}, TrDiMP \cite{trsiam}, TREG \cite{treg}, KeepTrack \cite{keeptrack}, PrDiMP \cite{prdimp}, Ocean \cite{ocean}, AutoMatch \cite{automatch}, DiMP \cite{dimp}, SiamRPN++ \cite{SiamRPN++} and SiamFC \cite{siamfc}. According to the results shown in Table \ref{fourth-tab}, our MAPNet-R tracker realizes very excellent performance on all evaluation metrics. Concretely, in comparison with the state-of-the-art ToMP \cite{tomp}, our approach produces substantial gains of 0.8\% on Success and 0.7\% on Normalized precision. Moreover, the proposed model surpasses another remarkable tracker, i.e., TrDiMP \cite{trsiam}, by 3.9\% on Success, 3.1\% on Normalized precision and 6.5\% on Precision, which also predicts the object state in a classification-regression parallel manner.

\begin{table}[t]
	\caption{Comparison with state-of-the-art trackers on GOT-10k. The best three results are highlighted in \textcolor[rgb]{1,0,0}{red}, \textcolor[rgb]{0,1,0}{green} and \textcolor[rgb]{0,0,1}{blue} fonts.}	\label{fifth-tab}	
	\begin{center}
		\setlength{\tabcolsep}{3mm}{		
			\begin{tabular}{lccc}
				\toprule[1.5pt] 
				Trackers & AO$\uparrow$ & SR$_{0.5}$$\uparrow$ & SR$_{0.75}$$\uparrow$ \\
				\midrule[1pt] 
				ECO	\cite{eco} & 0.316 & 0.309 & 0.111 \\
				SiamRPN++  \cite{SiamRPN++} &  0.517 & 0.616 &  0.325 \\
				ATOM \cite{ATOM} & 0.556 &	0.635 & 0.402 \\
				SiamFC++ \cite{siamfc++} & 0.595 &	0.695 & 0.479 \\
				OCEAN \cite{ocean} & 0.611 &	0.721 & 0.473 \\
				PrDiMP \cite{prdimp} & 0.634 &	0.738 & 0.543 \\
				SiamR-CNN \cite{siamr-cnn} & 0.649 &	0.728 & 0.597 \\
				
				TrDiMP \cite{trsiam} & 0.671 &	0.777 & 0.583 \\
				TransT \cite{trsiam} & 0.671 &	0.768 & 0.609 \\
				Stark \cite{stark} & \textcolor[rgb]{0,1,0}{0.688} &  0.781 & \textcolor[rgb]{0,1,0}{0.641} \\
				MATTrack \cite{mattrack} & 0.677 &	\textcolor[rgb]{0,1,0}{0.784} & -- \\
				SimTrack-B/16 \cite{simtrack} & \textcolor[rgb]{0,0,1}{0.686} &  \textcolor[rgb]{1,0,0}{0.789} & \textcolor[rgb]{0,0,1}{0.624} \\
				\midrule[1pt] 
				MAPNet-R & \textcolor[rgb]{1,0,0}{0.695} &  \textcolor[rgb]{0,1,0}{0.784} & \textcolor[rgb]{1,0,0}{0.649} \\
				\bottomrule[1.5pt]
		\end{tabular}}
	\end{center}
\end{table}
\subsubsection{GOK-10k}
We compare the presented tracker with a few outstanding algorithms on GOT-10k dataset, such as SimTrack-B/16 \cite{simtrack}, MATTrack \cite{mattrack}, TransT \cite{transt}, Stark \cite{stark} and so on, and report their tracking results on Table \ref{fifth-tab}. It is worth noting that the proposed tracker performs better than all comparison methods in term of overall performance. Compared with the state-of-the-art SimTrack-B/16 \cite{simtrack}, our tracker is inferior to it slightly on SR$_{0.5}$, but yields great increments on the rest two metrics, i.e., 0.9\% on AO and 2.5\% on SR$_{0.75}$. For fully-transformer tracker of MATTrack \cite{mattrack}, our method lifts the performance by 1.8\% on AO. In addition, our algorithm exceeds TransT by 2.4\% on AO, 1.6\% on SR$_{0.5}$ and 4.0\% on SR$_{0.75}$.

\begin{figure*}[t]
	\begin{center}
		\includegraphics[width=0.4\linewidth]{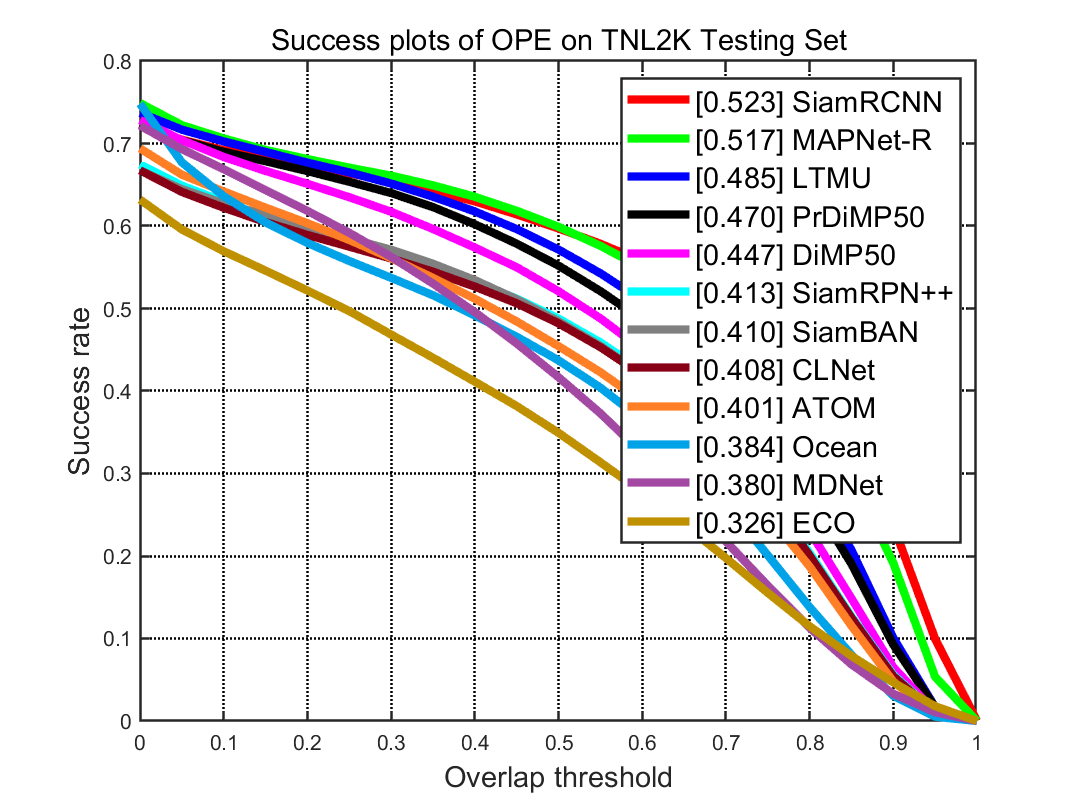}
		\includegraphics[width=0.4\linewidth]{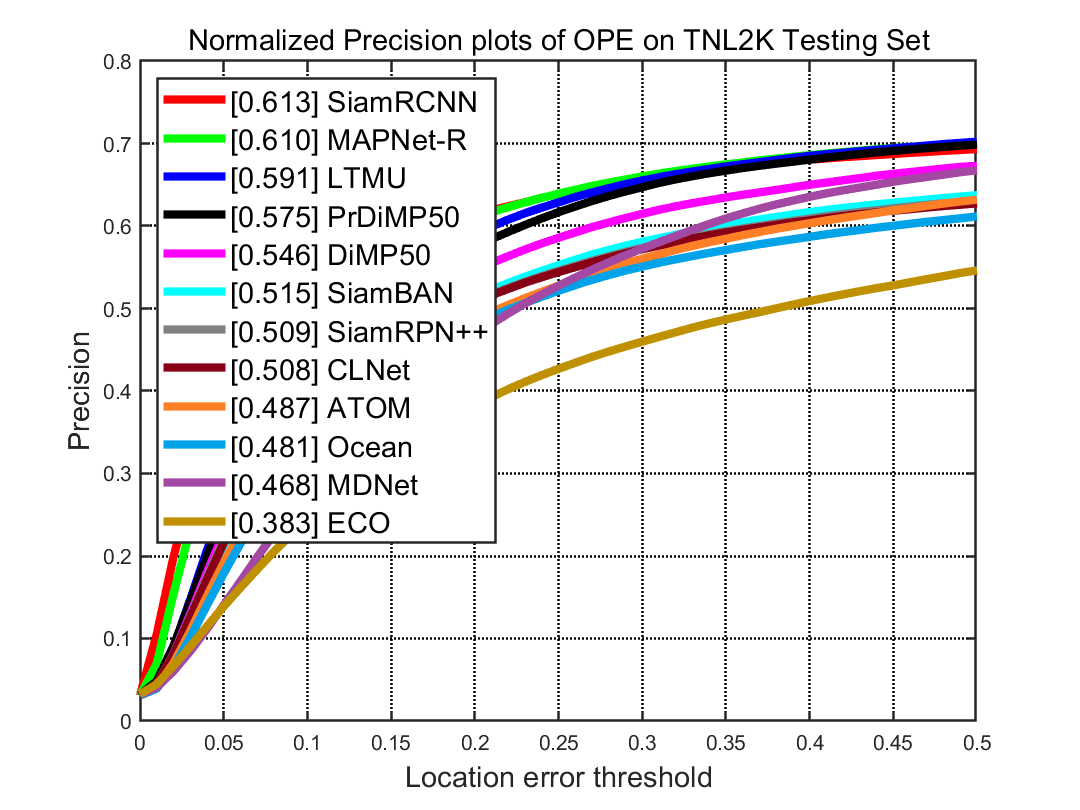}
	\end{center}
	\caption{Success and Normalized precision plots of all trackers on TNL2k. These trackers are ranked according to their performance scores.}
	\label{seventh-fig}
\end{figure*}

\begin{figure*}[t]
	\begin{center}
		\includegraphics[width=0.4\linewidth]{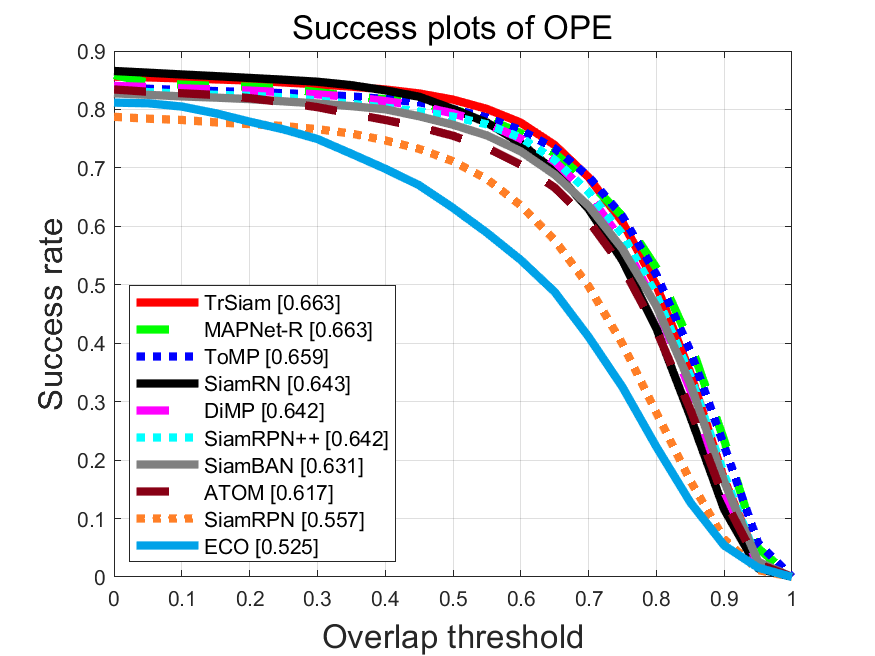}
		\includegraphics[width=0.4\linewidth]{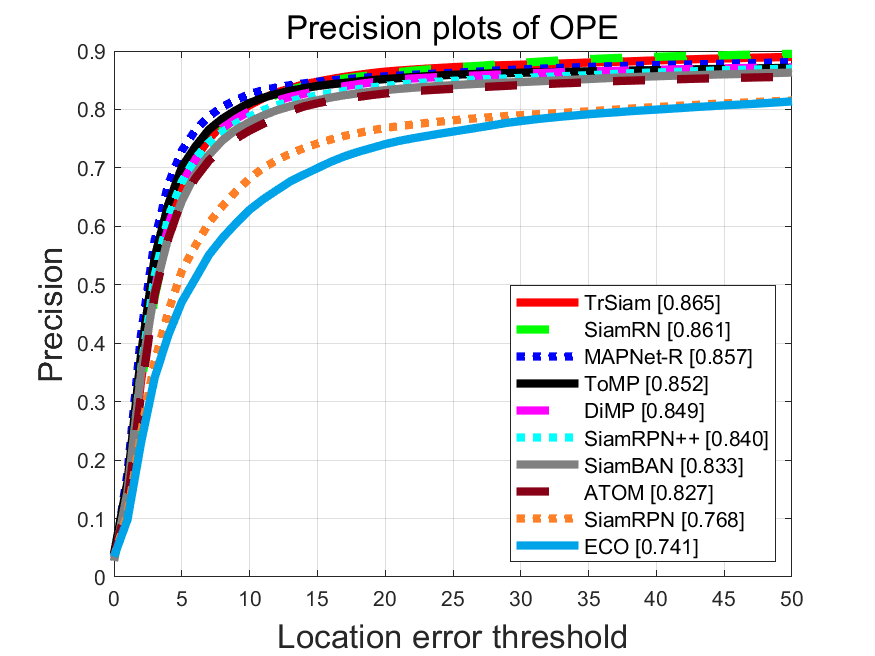}
	\end{center}
	\caption{Success and Precision plots of all comparison algorithms on UAV123. These algorithms are ranked according to the performance scores.}
	\label{eighth-fig}
\end{figure*}
 
\subsubsection{TNL2k}
We conduct quantitative experiments on TNL2k dataset by comparing our tracker with several representative works, i.e., SiamR-CNN \cite{siamr-cnn}, LTMU \cite{ltmu}, PrDiMP \cite{prdimp}, DiMP \cite{dimp}, SiamRPN++ \cite{SiamRPN++}, SiamBAN \cite{siamban}, CLNet \cite{clnet}, ATOM \cite{ATOM}, Ocean \cite{ocean}, MDNet \cite{mdnet} and ECO \cite{eco}. As shown in Fig. \ref{seventh-fig}, the proposed approach realizes very satisfactory performance on both Success and Normalized precision, which only has a small gap with SiamR-CNN \cite{siamr-cnn}. Compared with the typical long-term tracker of LTMU \cite{ltmu}, our MAPNet-R obtains significant improvements of 3.2\% on Success and 1.9\% on Normalized precision.

\subsubsection{UAV123}
We produce the Success and Precision plots of our method on the benchmark in Fig. \ref{eighth-fig}, in which some recently proposed trackers are adopted for comparison, including TrSiam \cite{trsiam}, ToMP \cite{tomp}, SiamRN \cite{siamrn}, SiamRPN++ \cite{SiamRPN++}, SiamBAN \cite{siamban}, DiMP \cite{dimp}, ATOM \cite{ATOM}, SiamRPN \cite{siamrpn} and ECO \cite{eco}. The proposed MAPNet-R gains very outstanding performance and performs favorably against most of recent advanced trackers on both Success and Precision metrics. The only exception is the TrSiam \cite{trsiam}, which outperforms our model slightly. The main reason is that TrSiam exploited a temporal-spatial transformer to model the dependencies between multi-stage object samples, which is very valuable for adapting to the severe appearance variations of object.

\begin{figure*}[t]
	\begin{center}
		\includegraphics[width=0.195\linewidth]{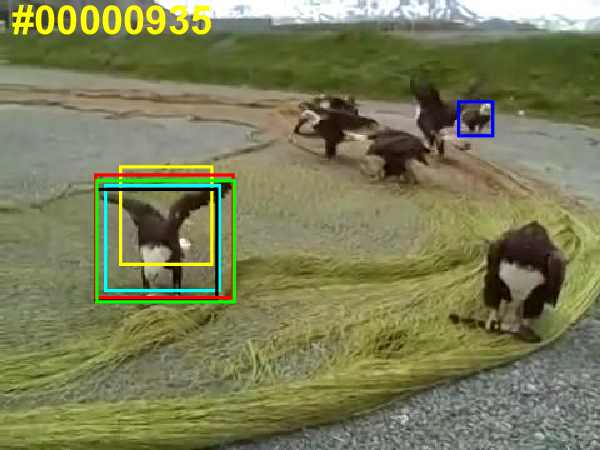}
		\includegraphics[width=0.195\linewidth]{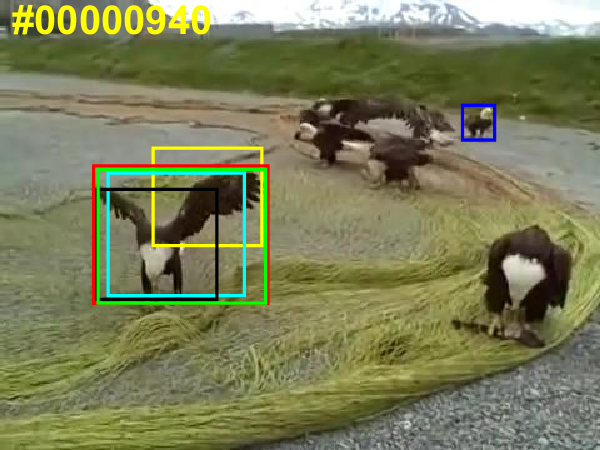}
		\includegraphics[width=0.195\linewidth]{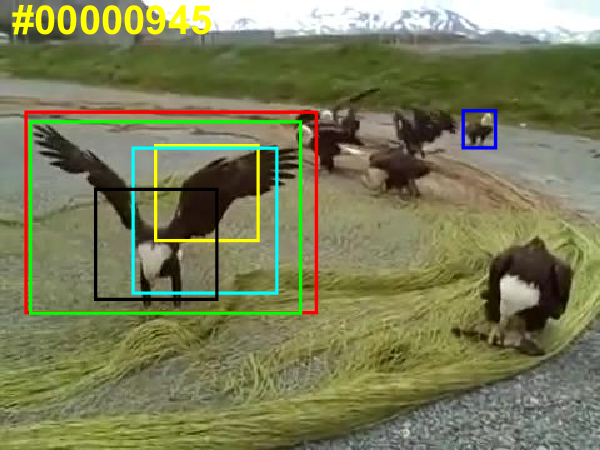}
		\includegraphics[width=0.195\linewidth]{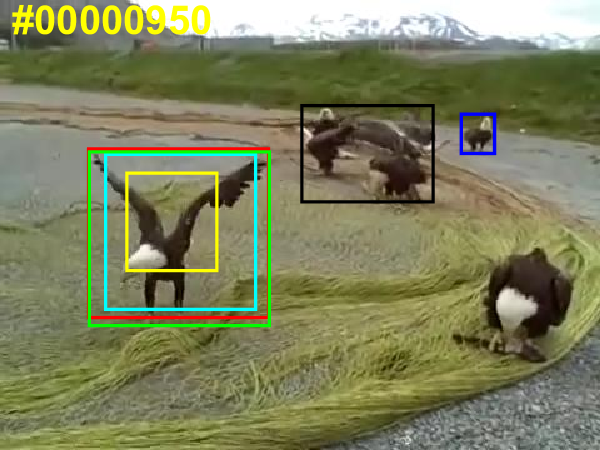}
		\includegraphics[width=0.195\linewidth]{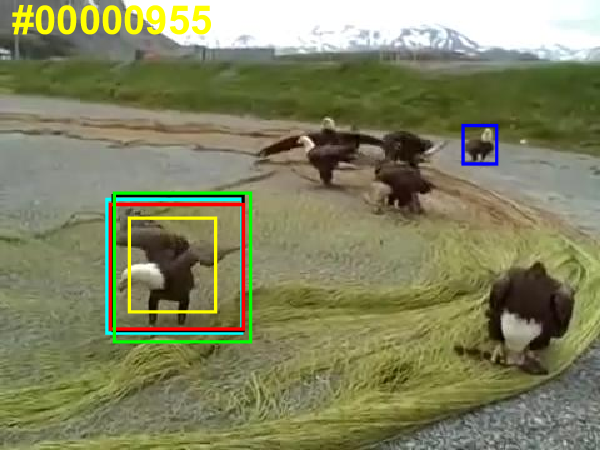}
		
		\includegraphics[width=0.195\linewidth]{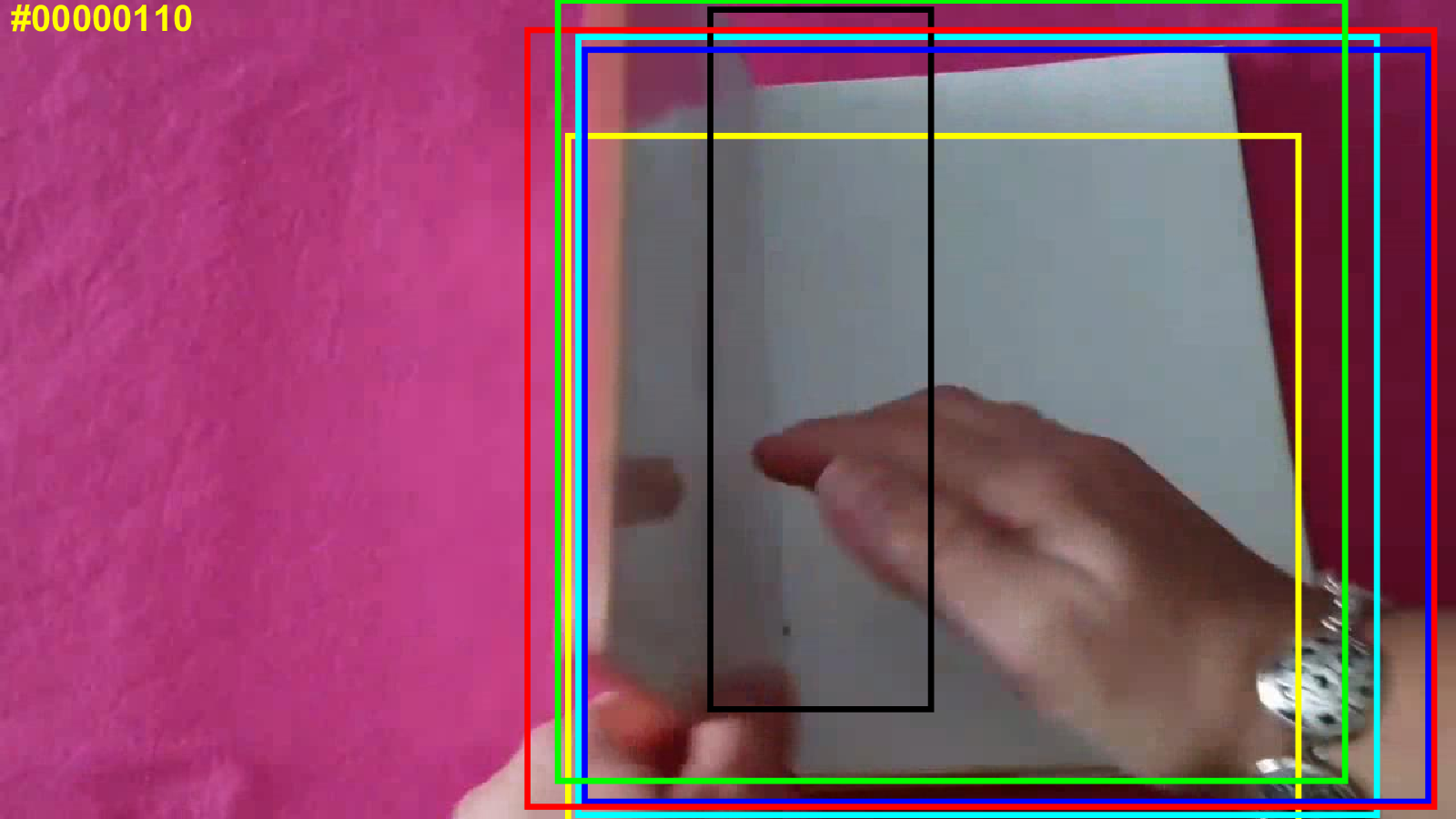}
		\includegraphics[width=0.195\linewidth]{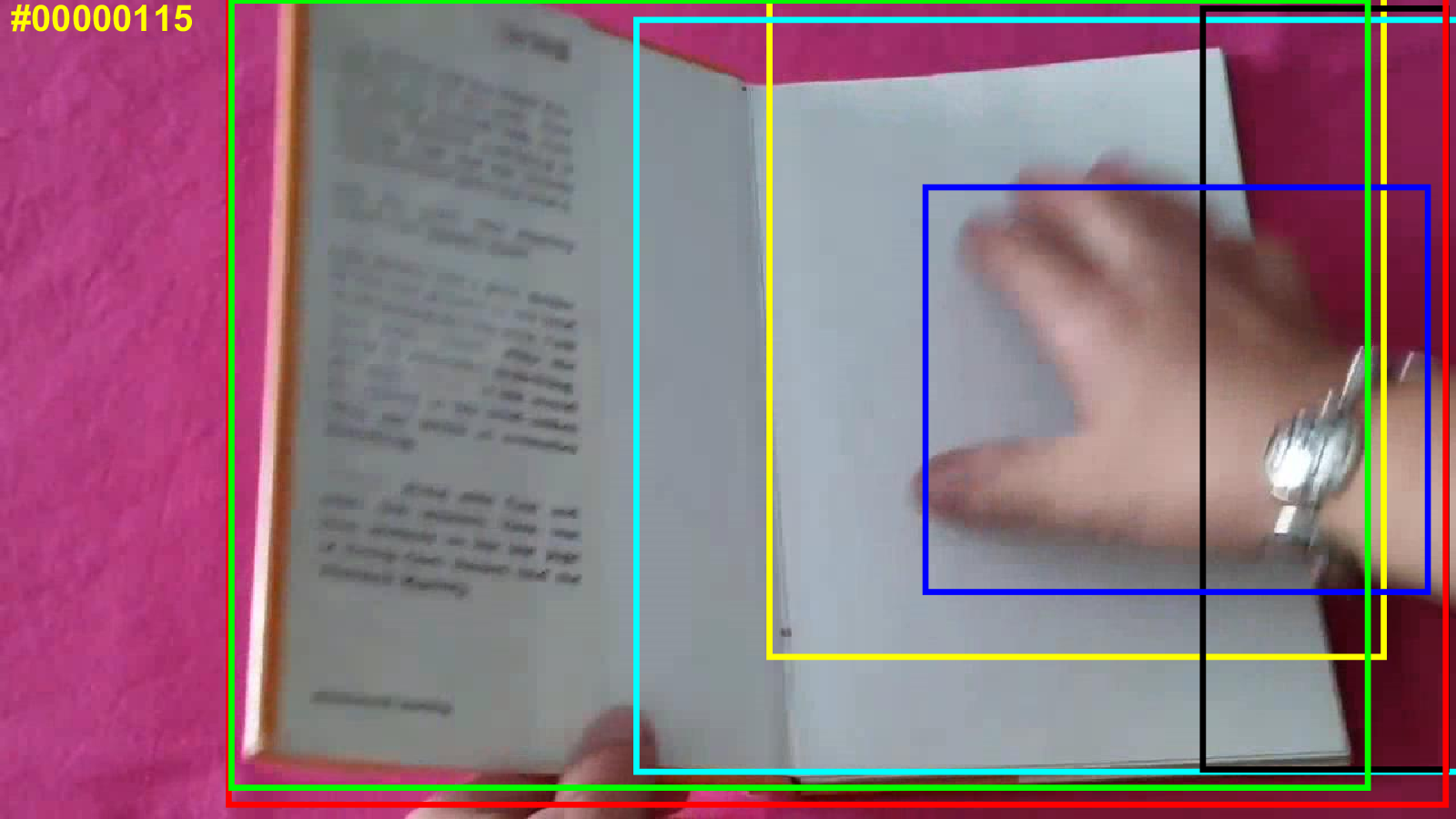}
		\includegraphics[width=0.195\linewidth]{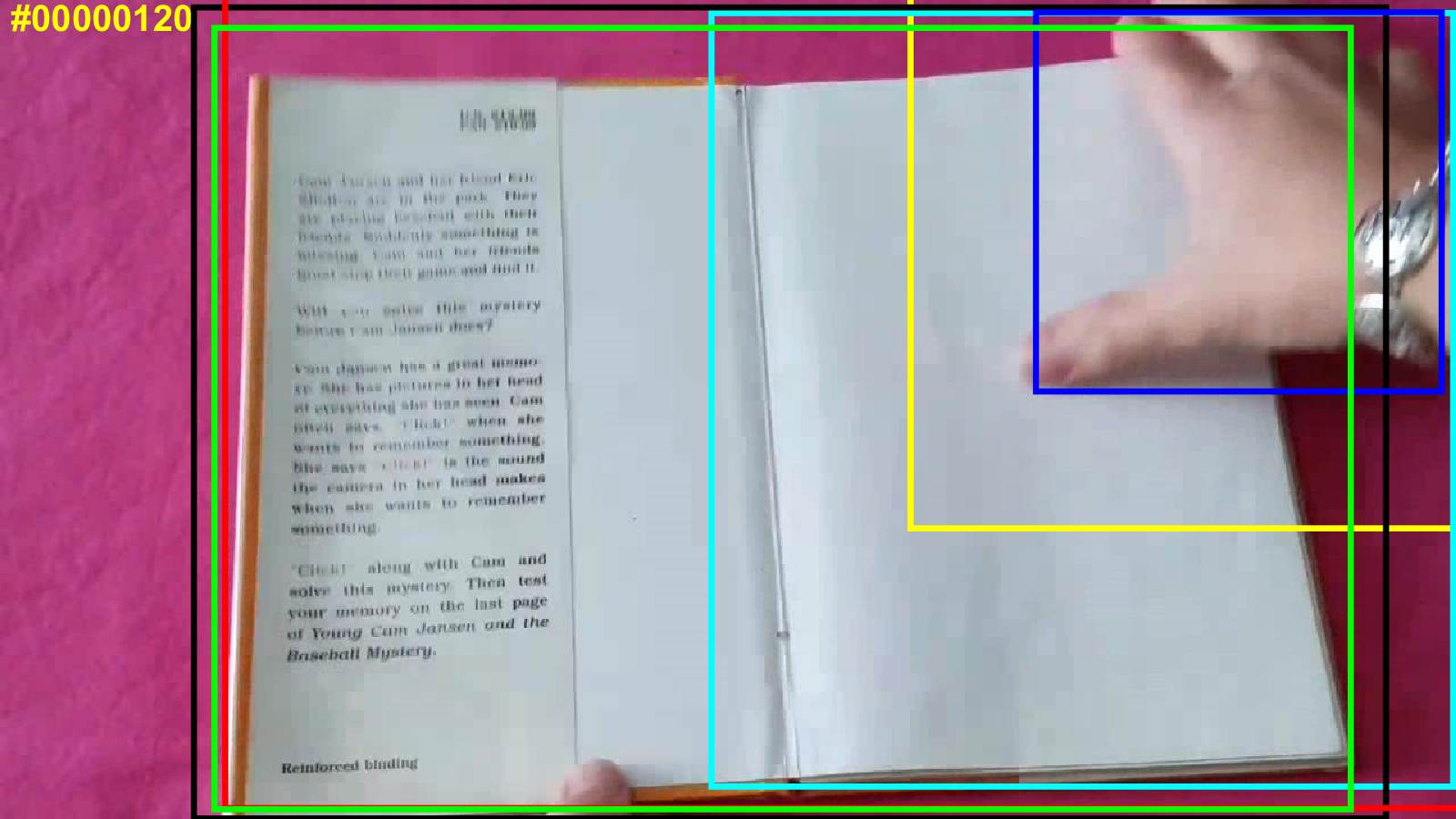}
		\includegraphics[width=0.195\linewidth]{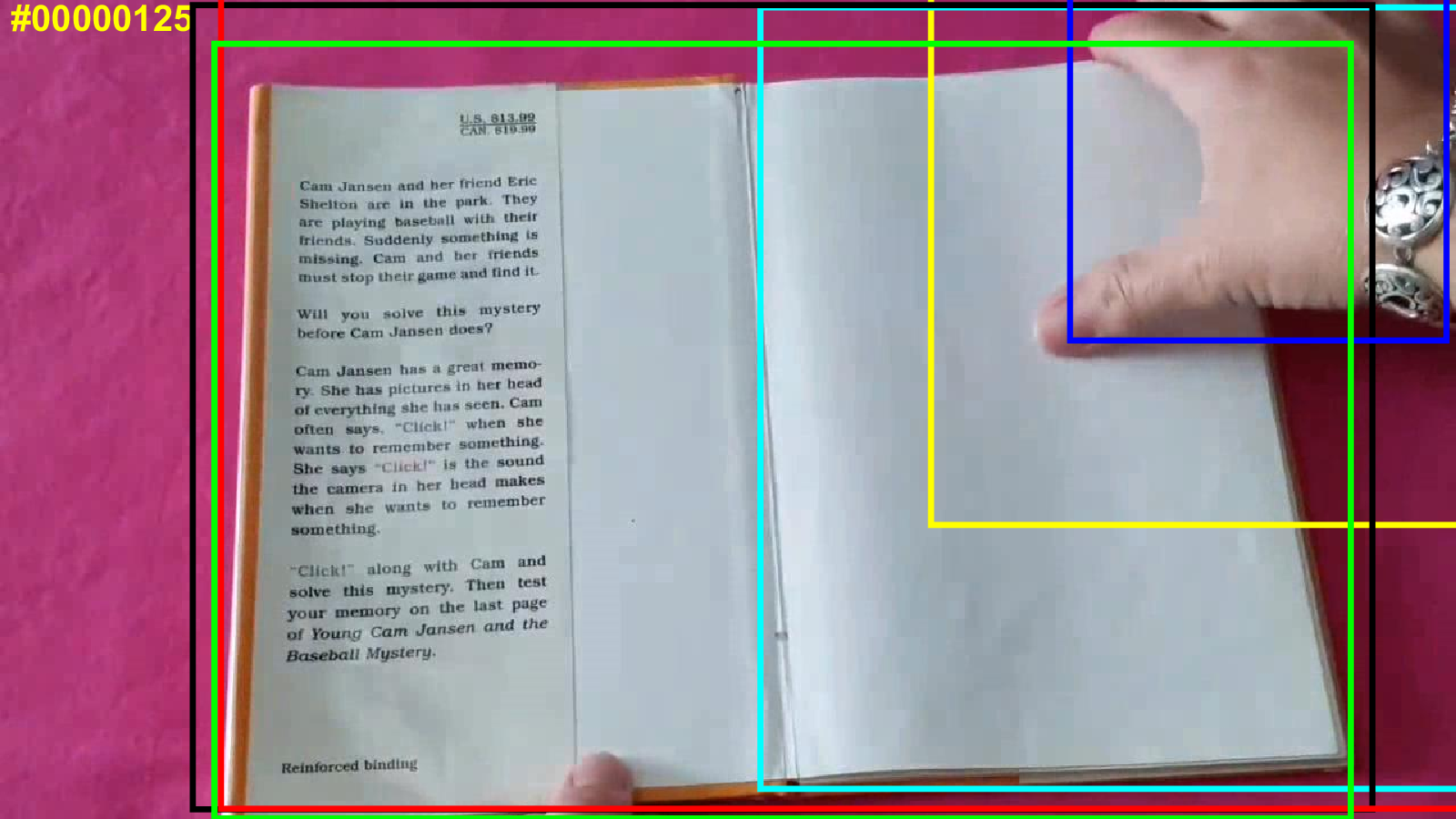}
		\includegraphics[width=0.195\linewidth]{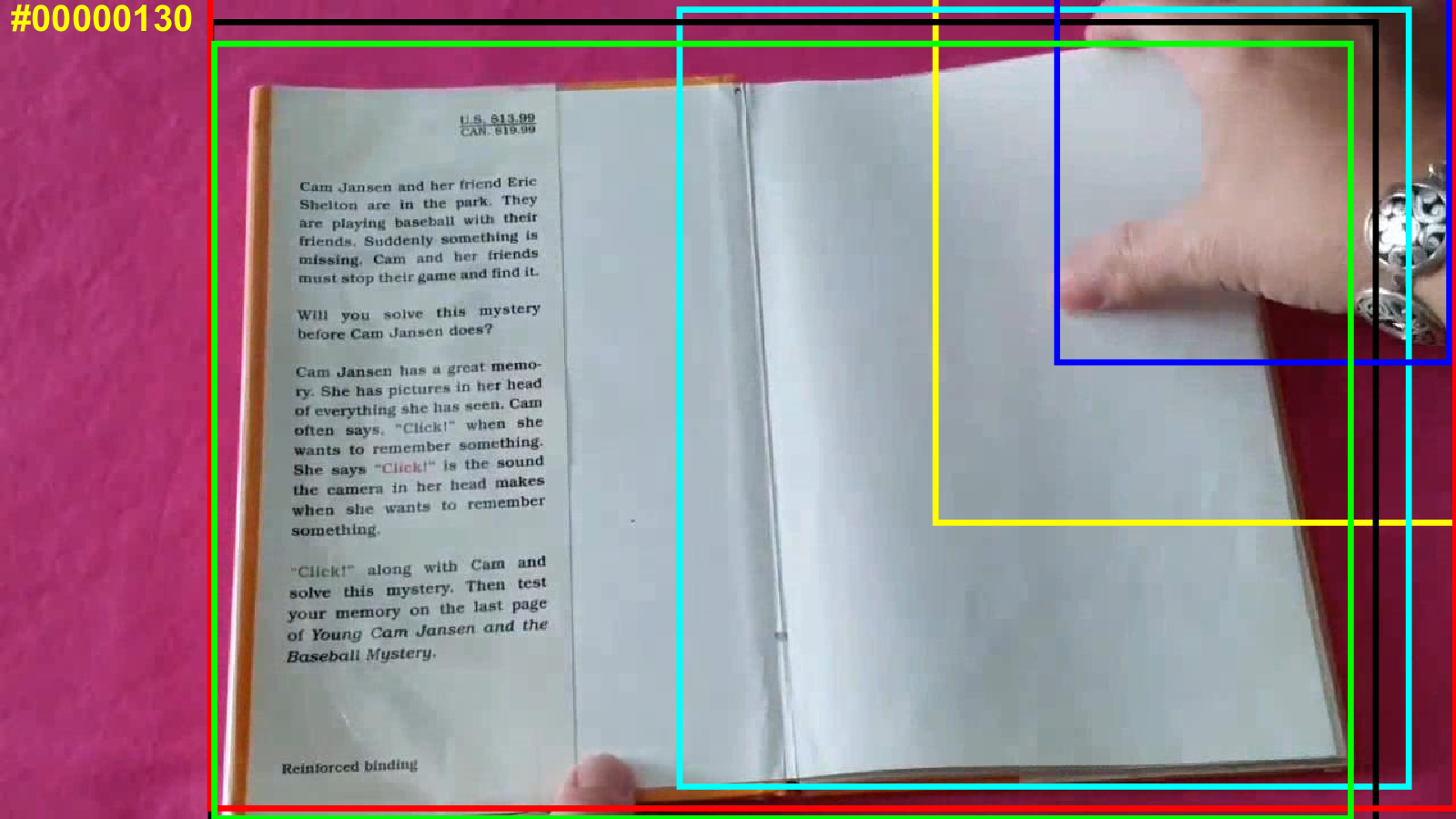}
		
		\includegraphics[width=0.195\linewidth]{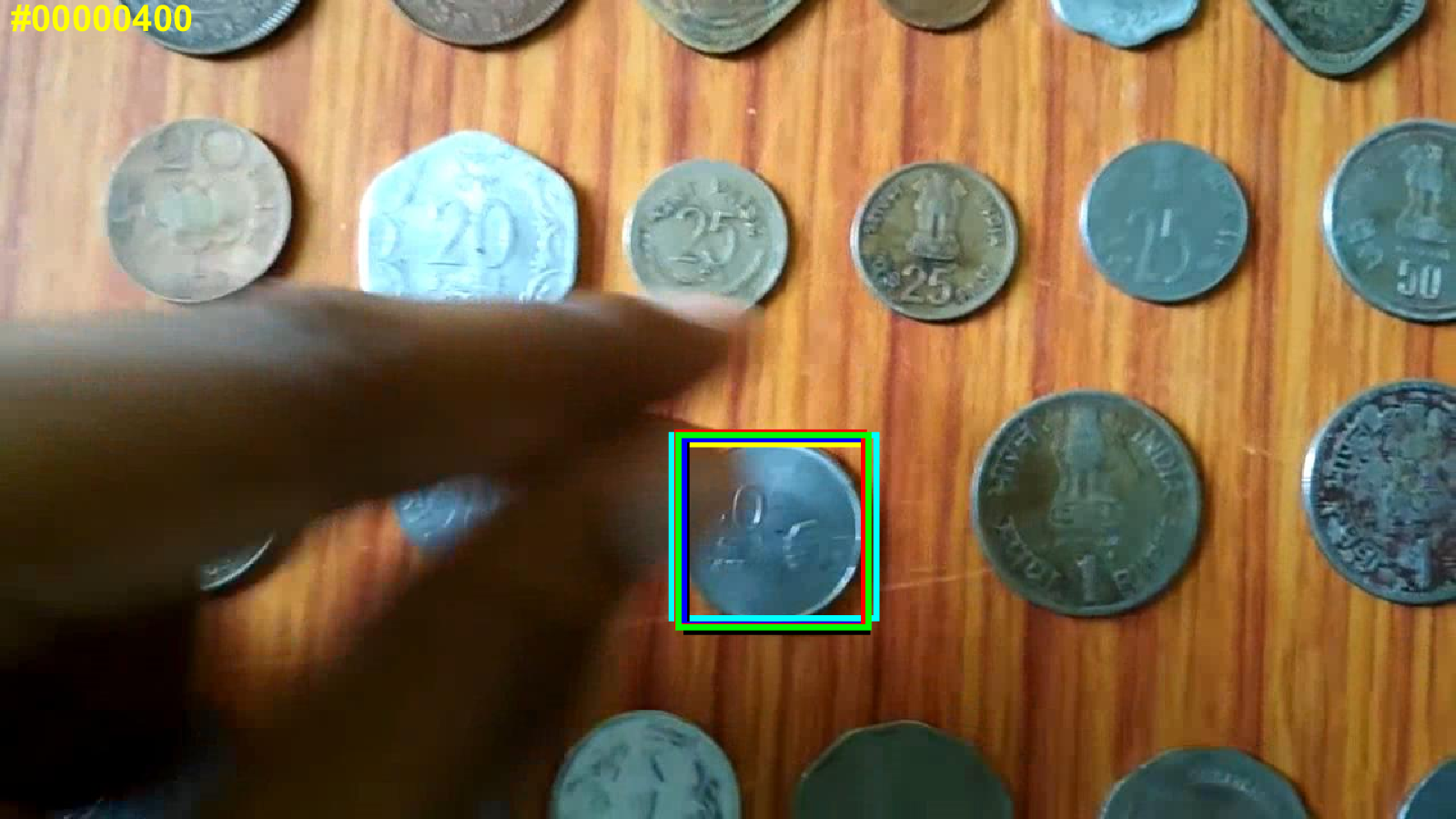}
		\includegraphics[width=0.195\linewidth]{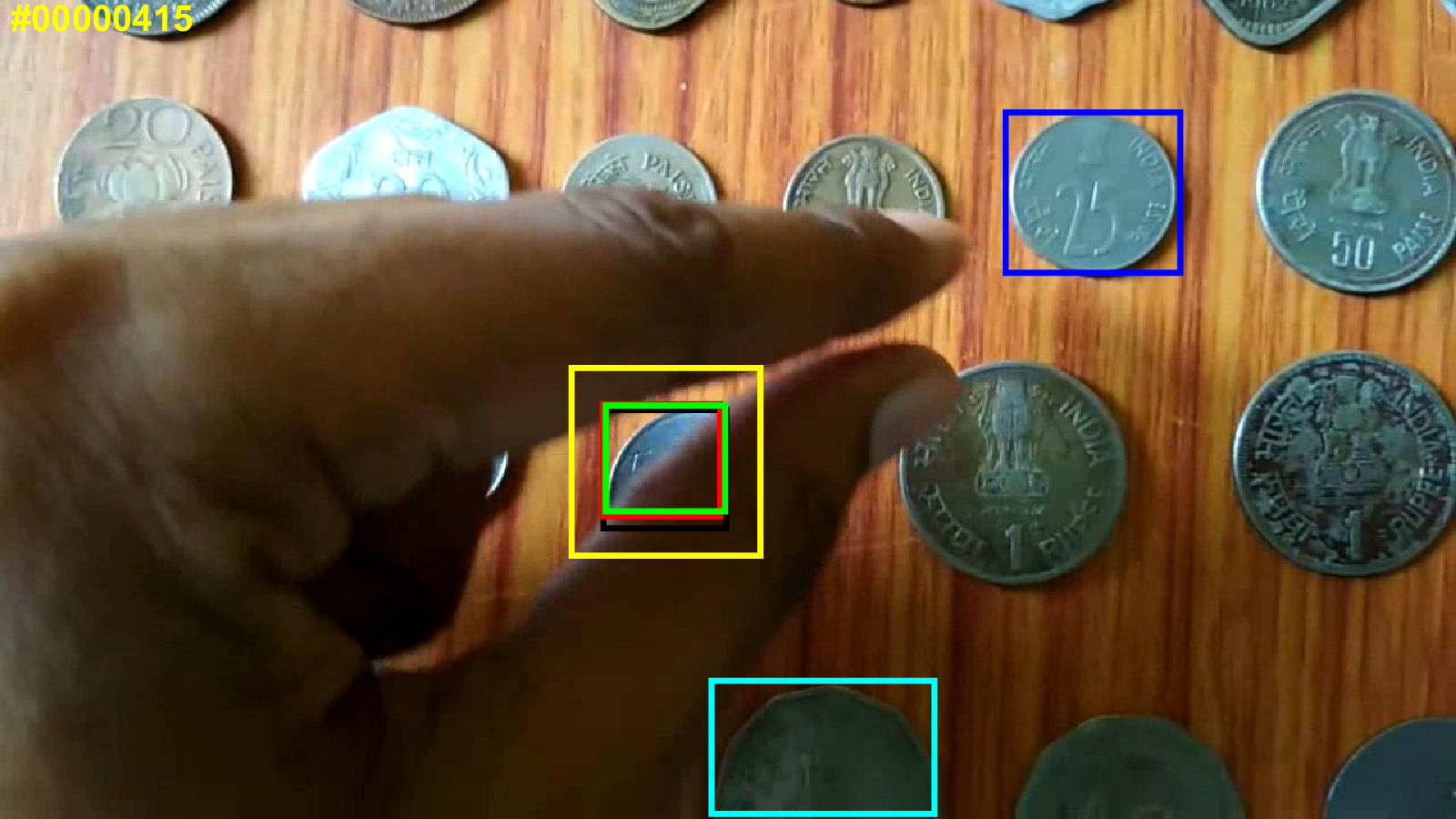}
		\includegraphics[width=0.195\linewidth]{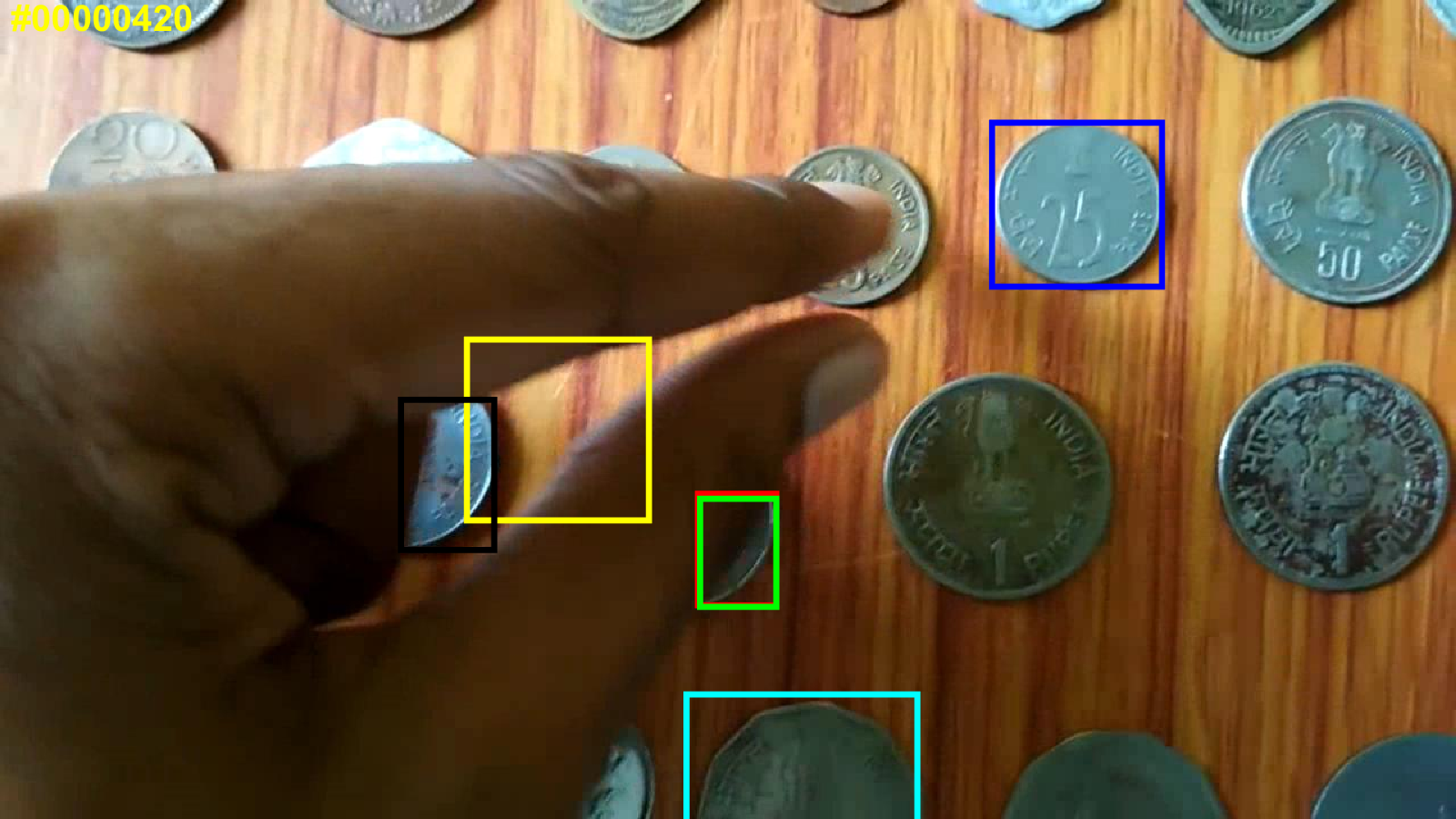}
		\includegraphics[width=0.195\linewidth]{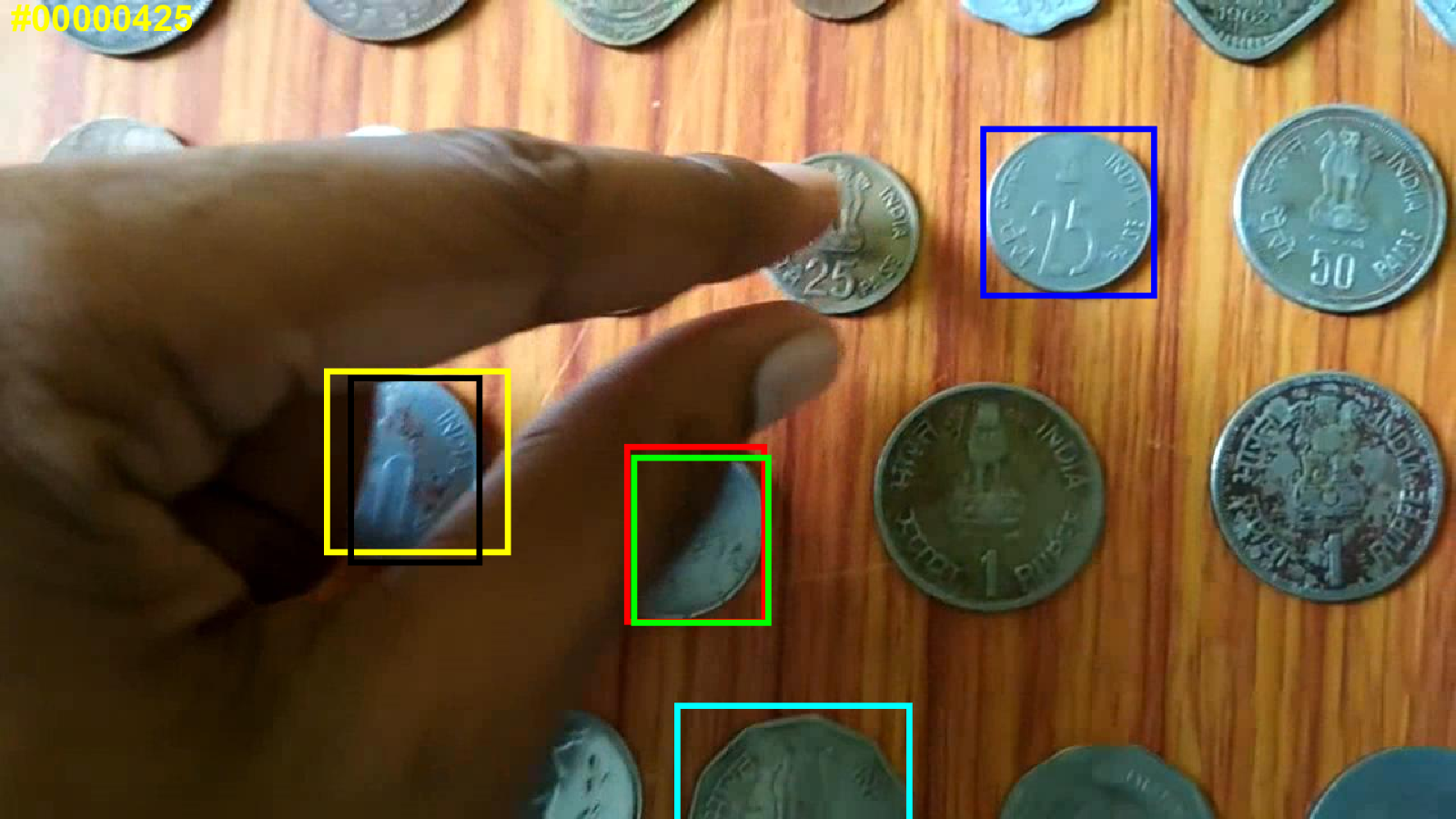}
		\includegraphics[width=0.195\linewidth]{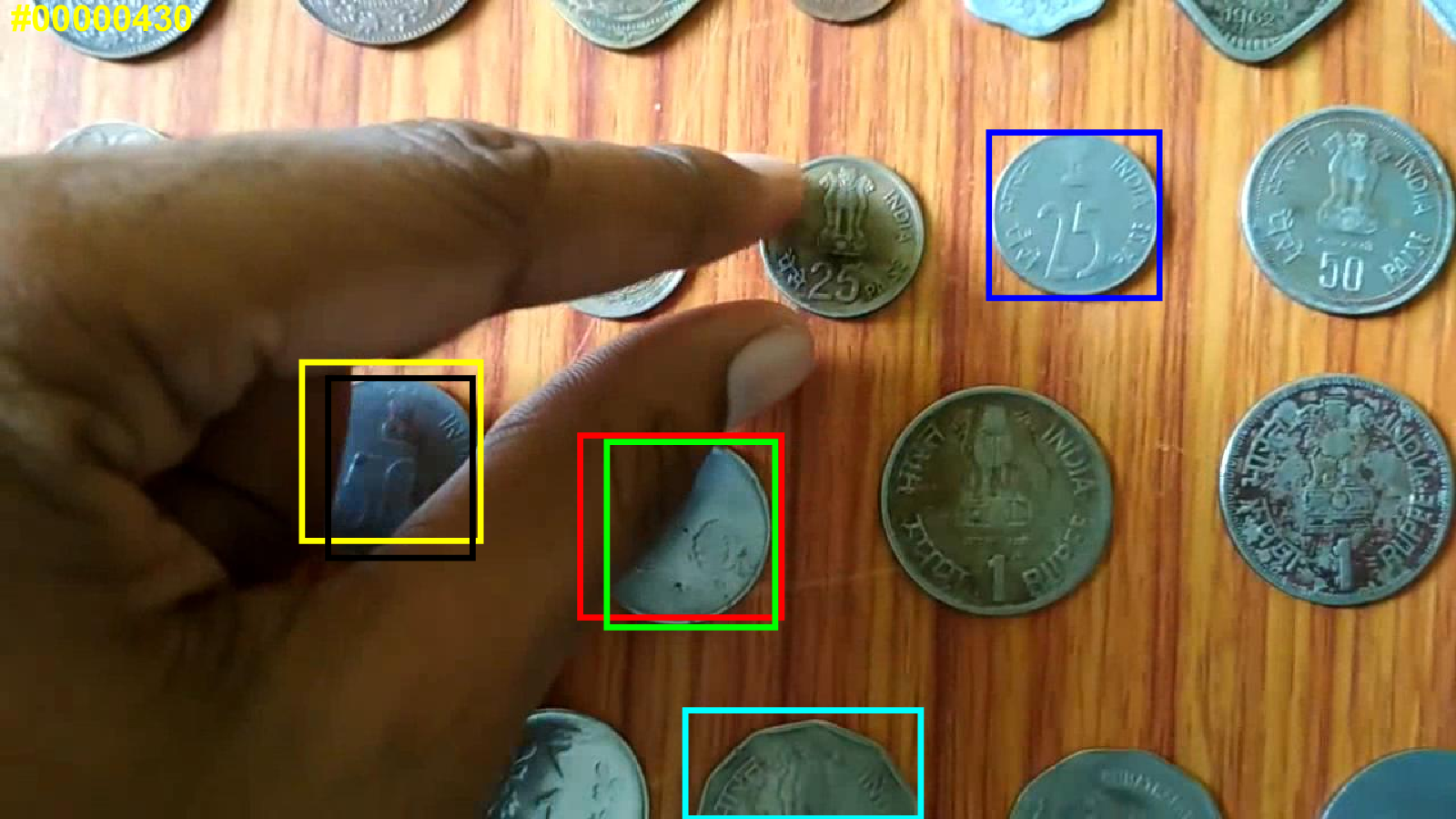}
		
		\includegraphics[width=0.195\linewidth]{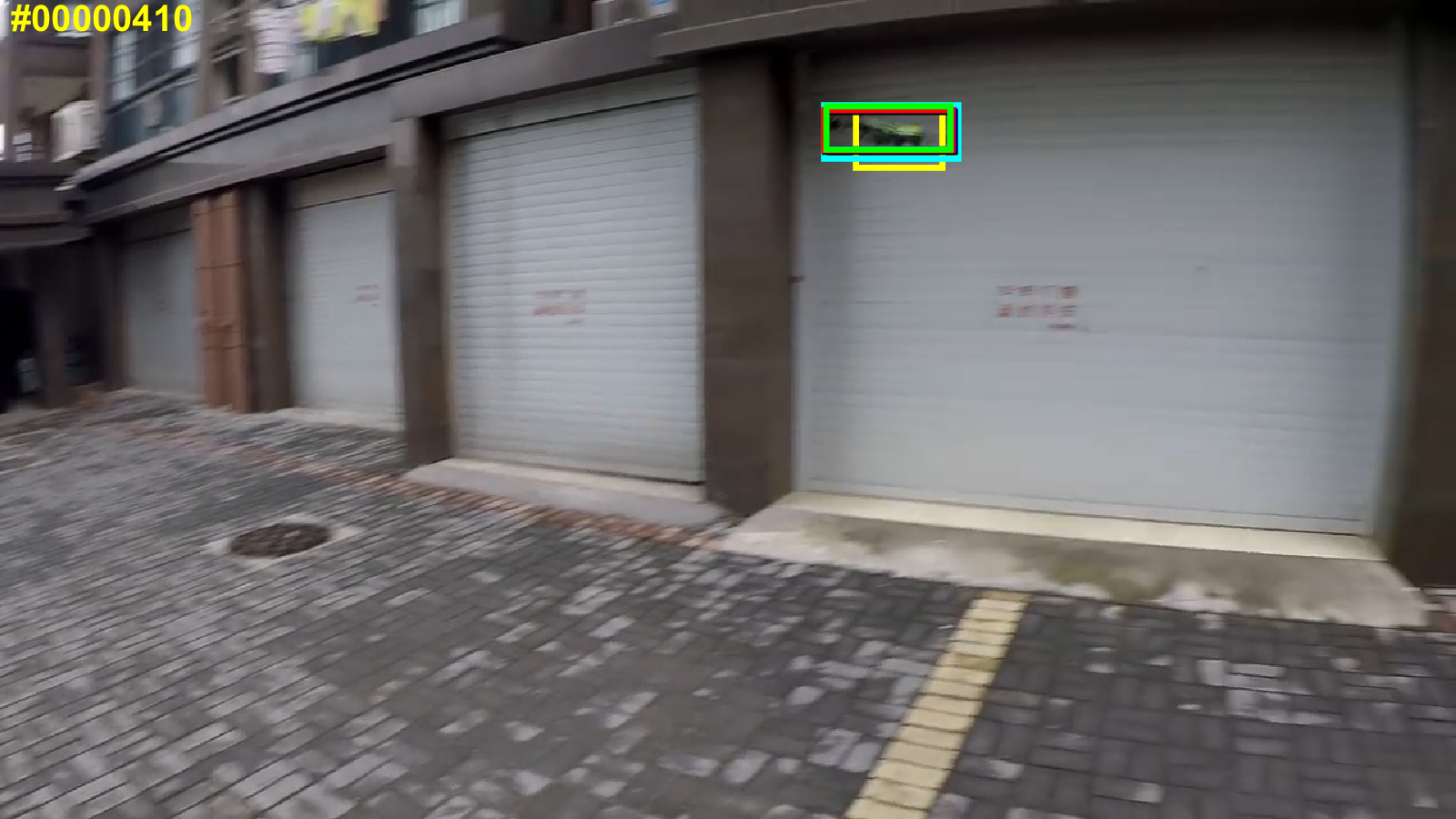}
		\includegraphics[width=0.195\linewidth]{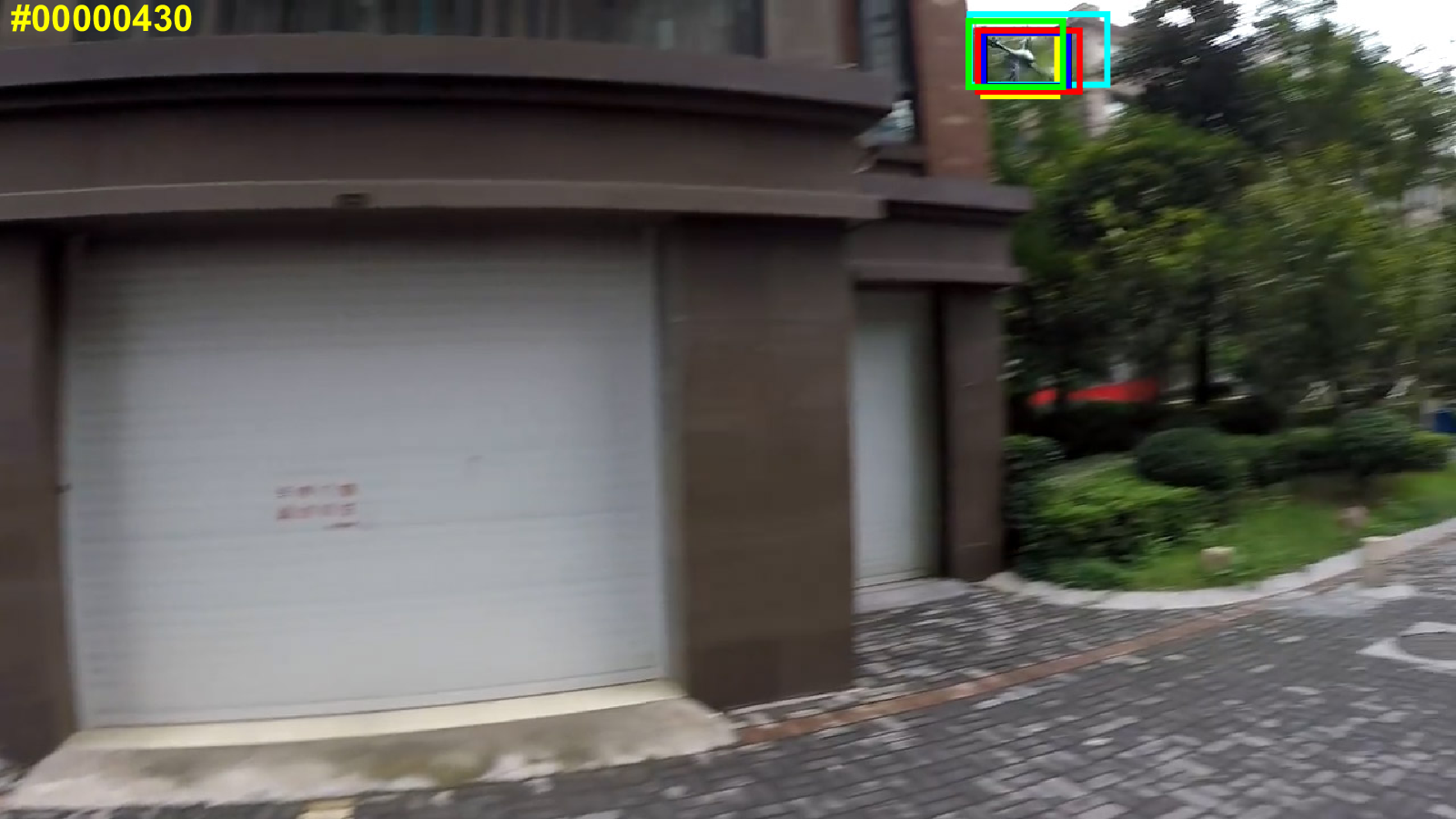}
		\includegraphics[width=0.195\linewidth]{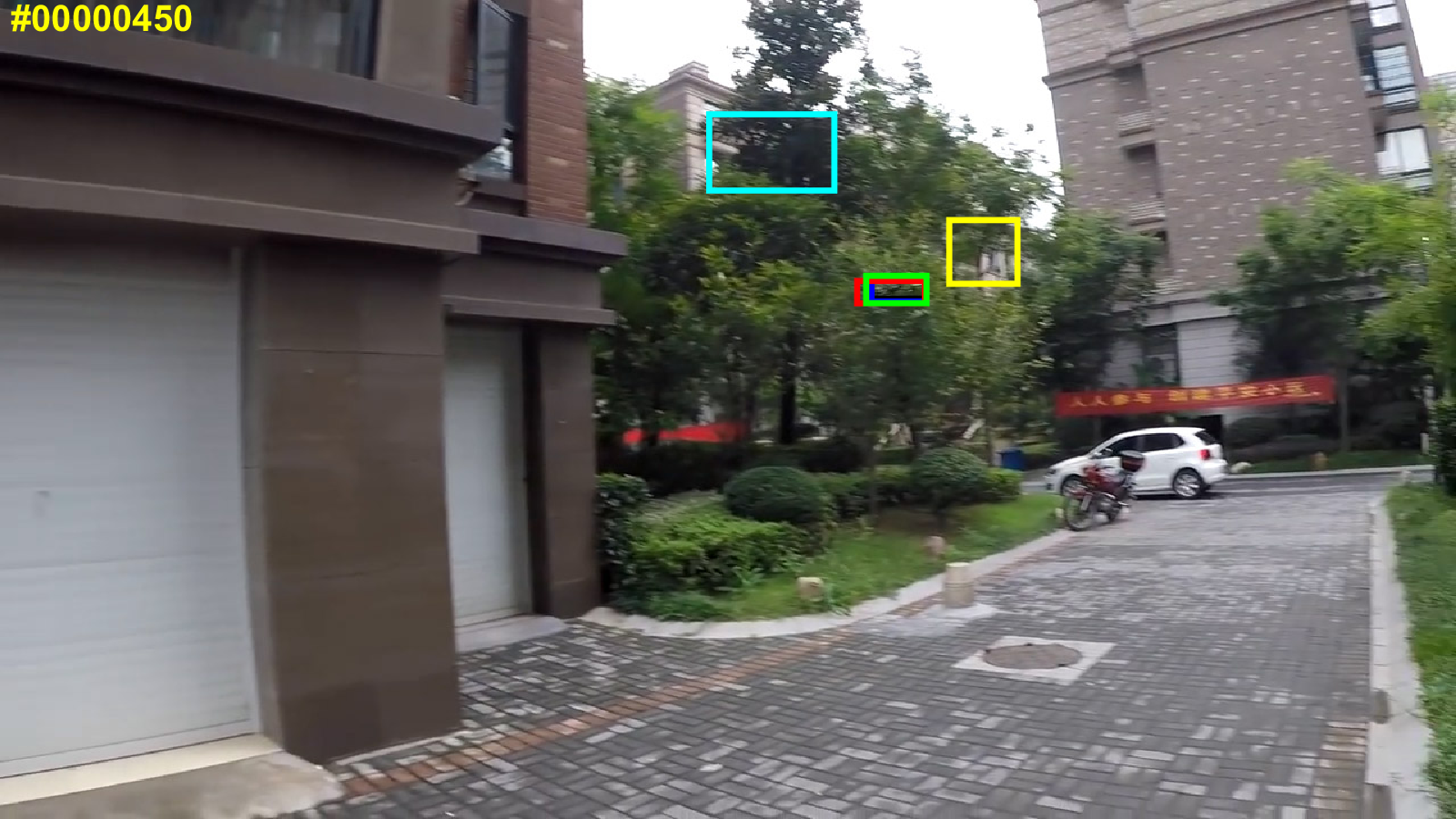}
		\includegraphics[width=0.195\linewidth]{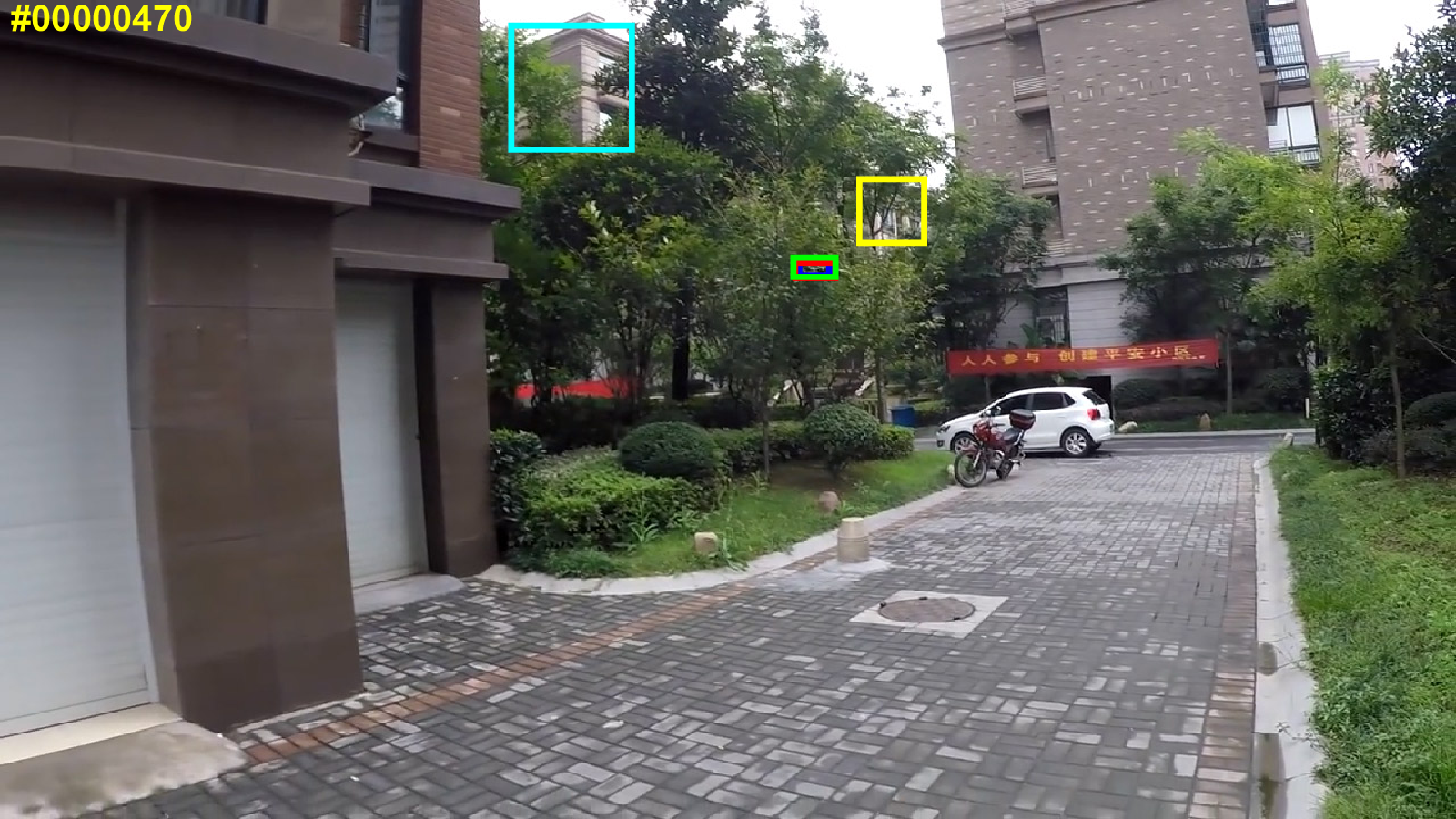}
		\includegraphics[width=0.195\linewidth]{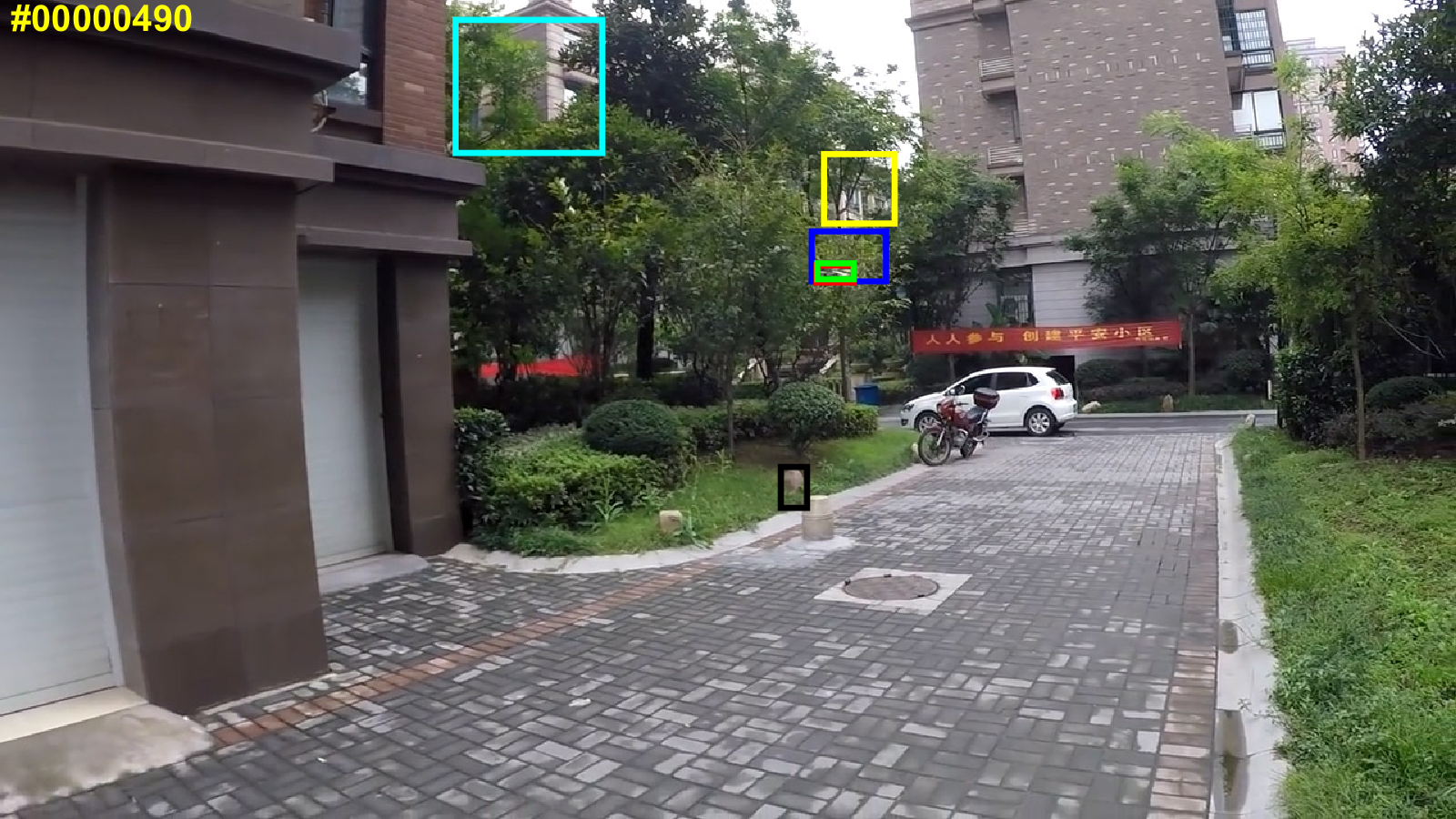}
		
		\includegraphics[width=0.195\linewidth]{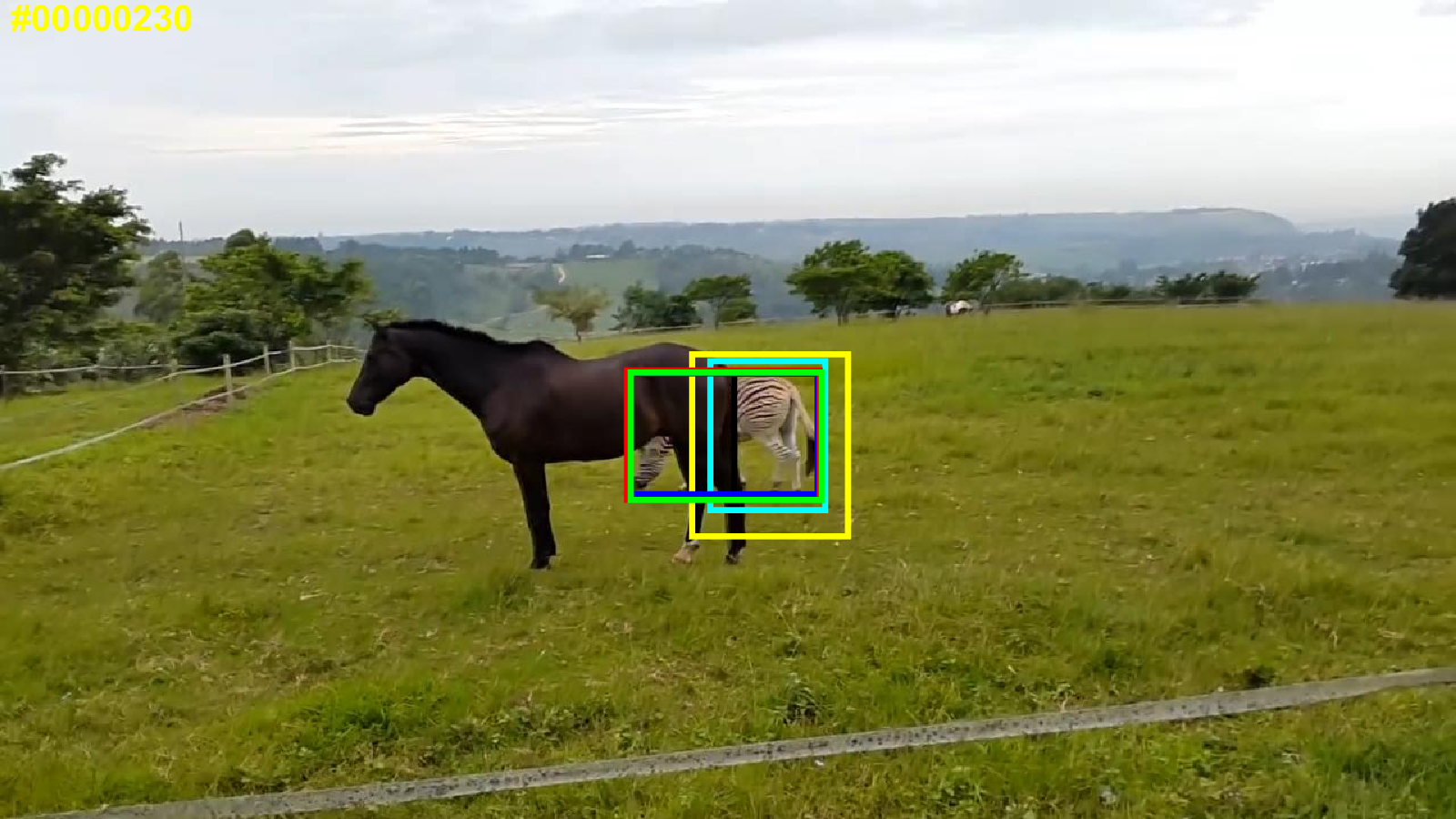}
		\includegraphics[width=0.195\linewidth]{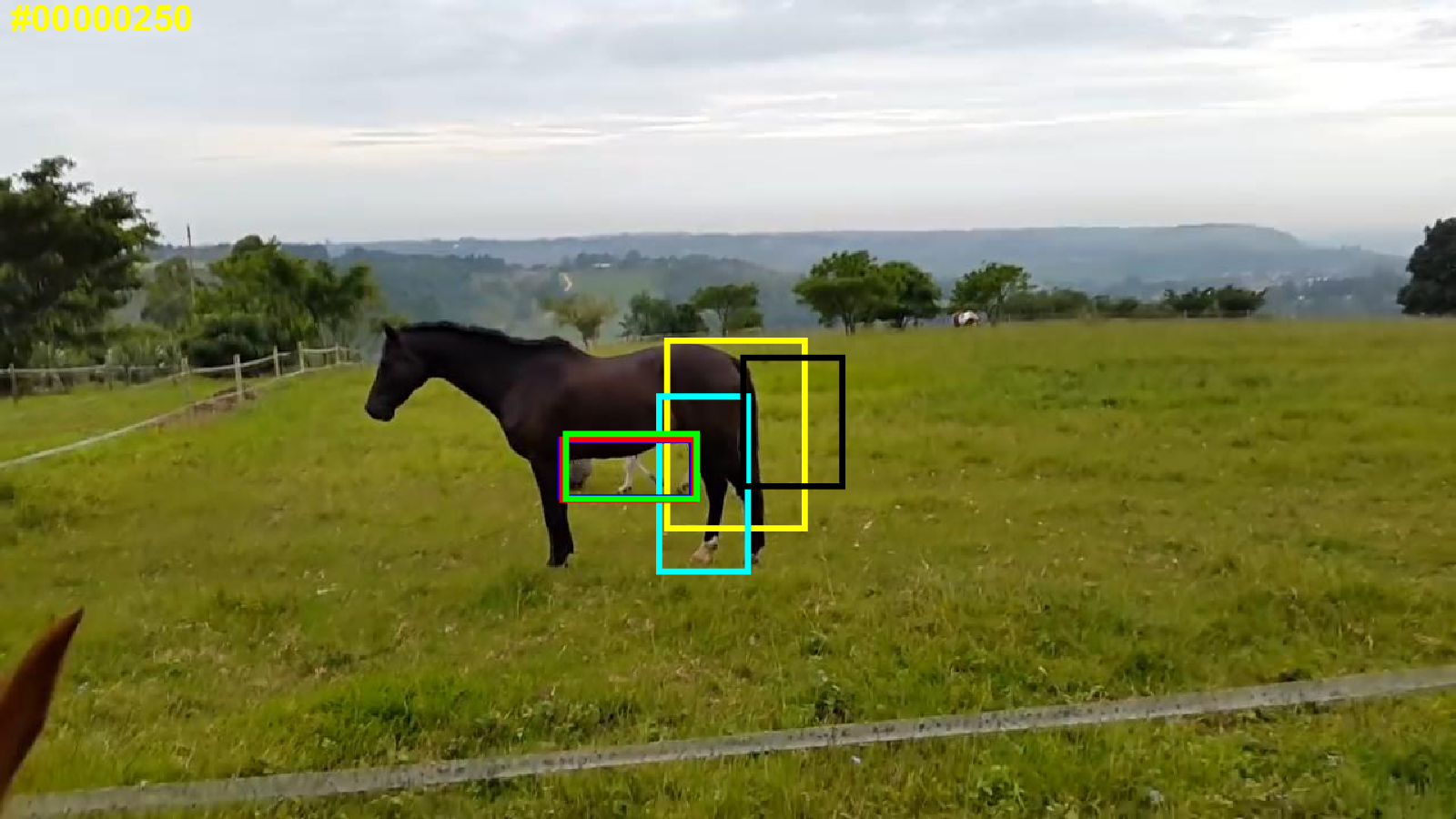}
		\includegraphics[width=0.195\linewidth]{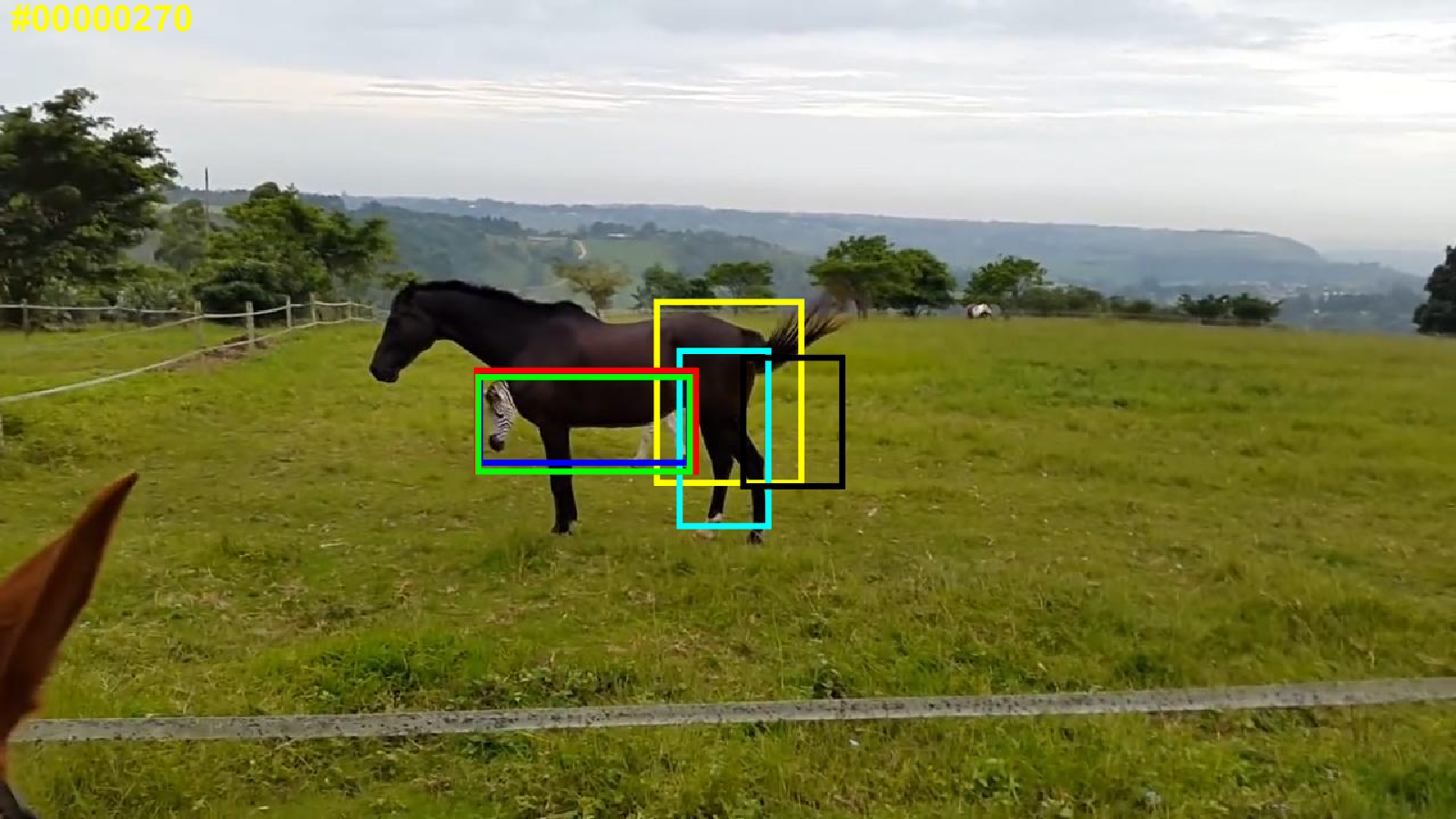}
		\includegraphics[width=0.195\linewidth]{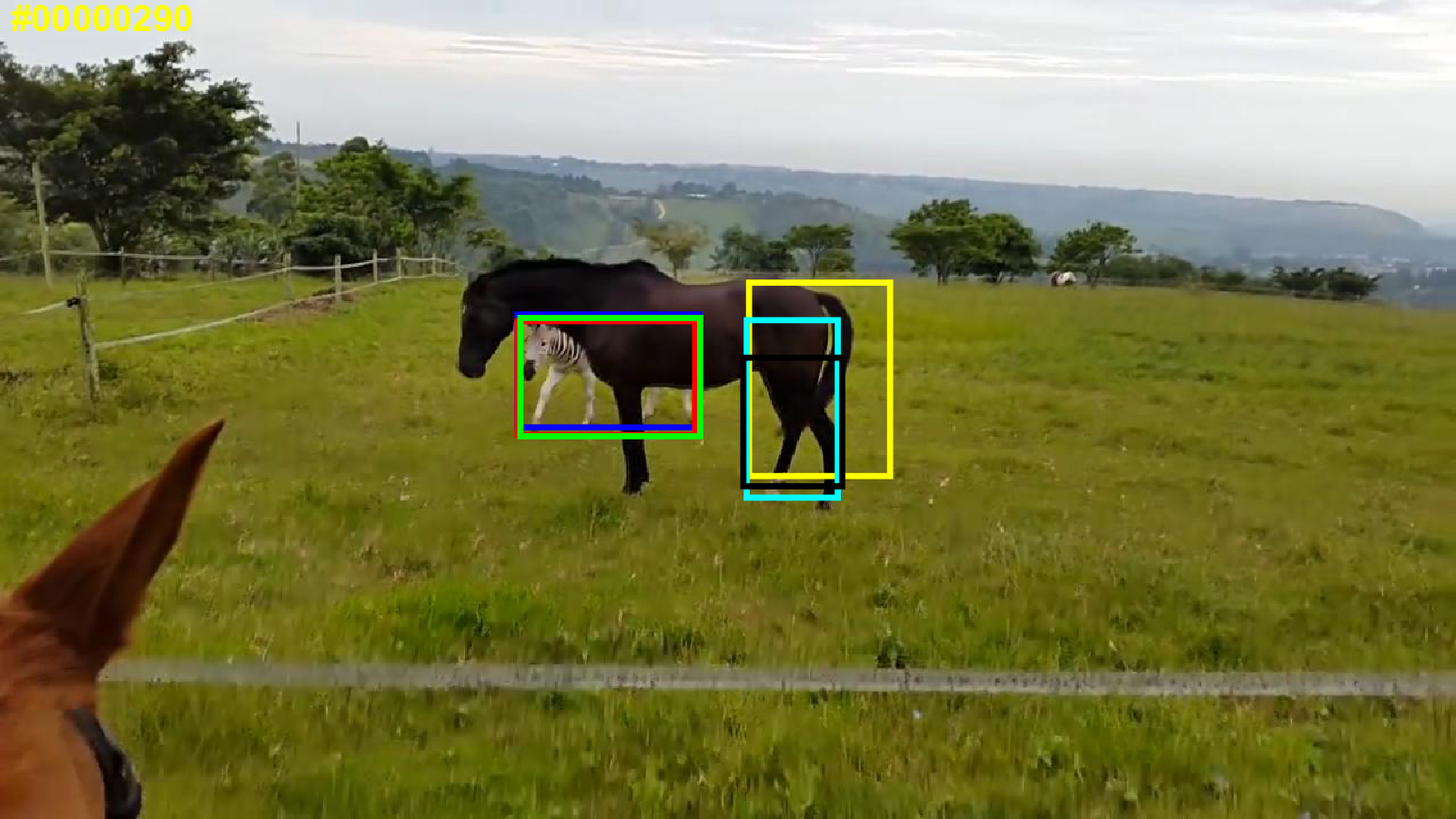}
		\includegraphics[width=0.195\linewidth]{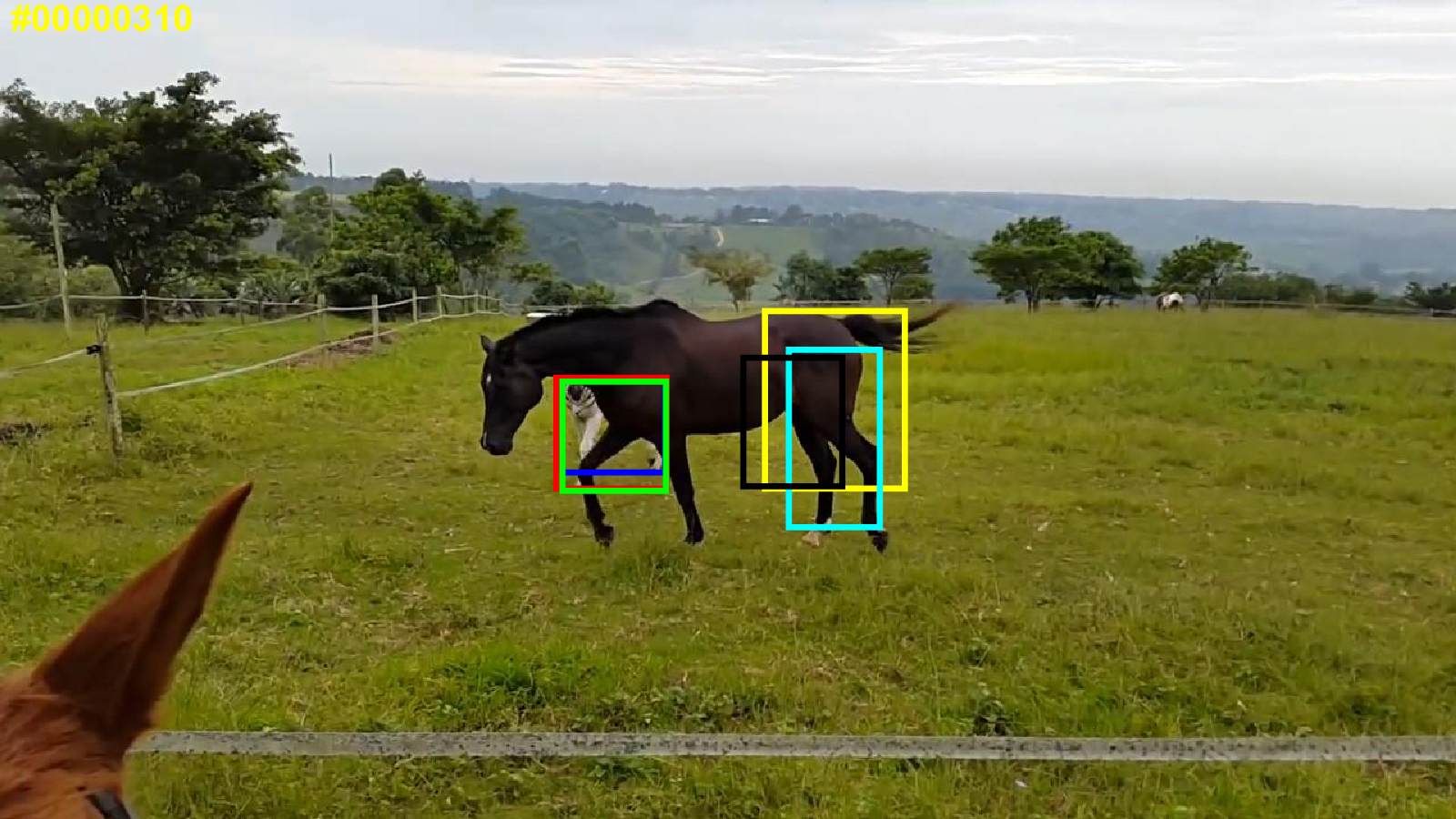}
		
		\includegraphics[width=0.8\linewidth]{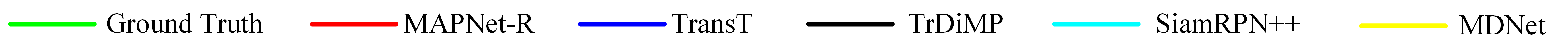}
	\end{center}
	\caption{Qualitative comparisons with four state-of-the-art trackers on several challenging sequences of LaSOT dataset (bird-5, book-10, coin-18, drone-13, zebra-17).}
	\label{ninth-fig}
\end{figure*}

\begin{figure*}[t]
	\begin{center}
		\includegraphics[width=0.195\linewidth]{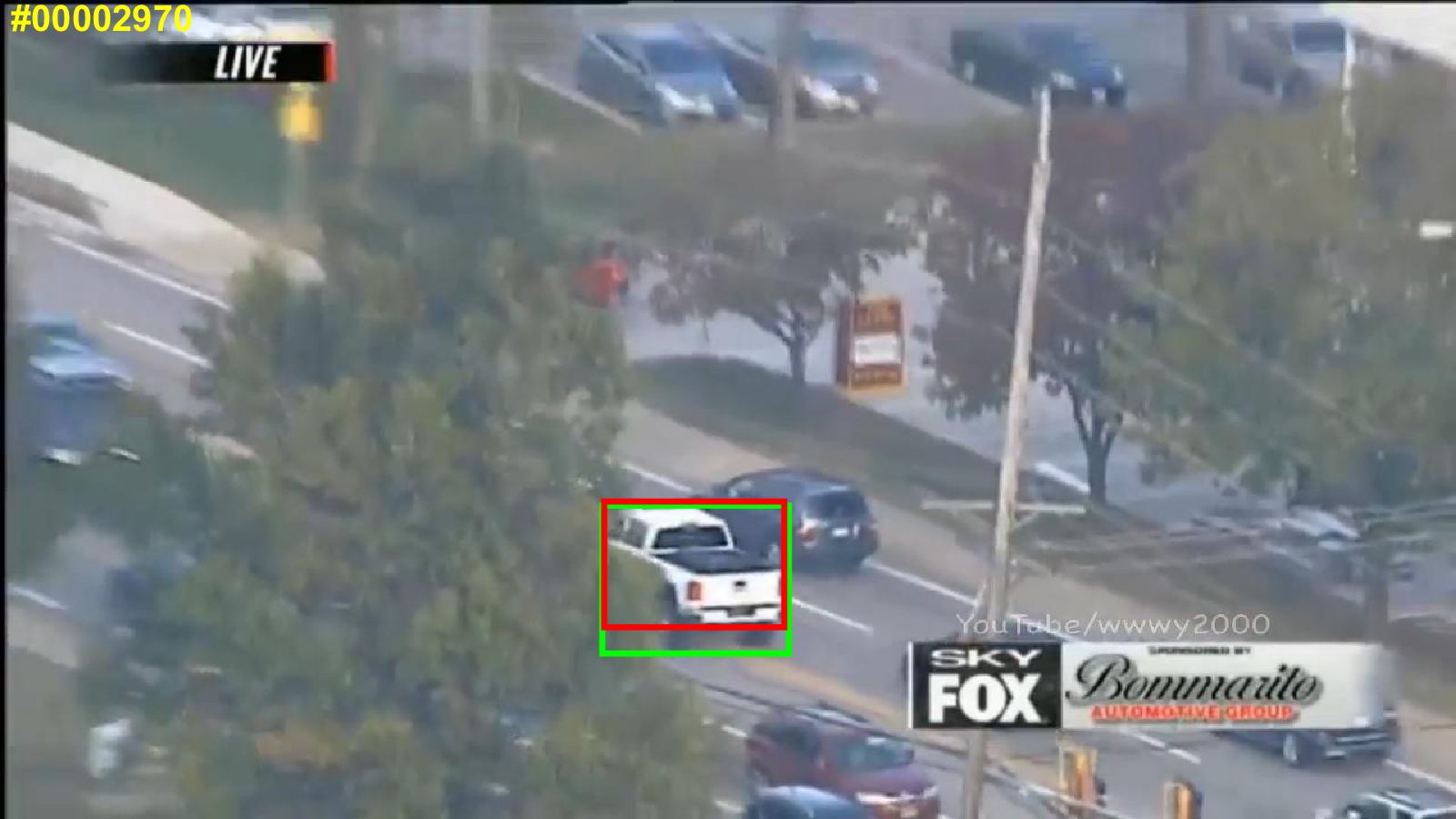}
		\includegraphics[width=0.195\linewidth]{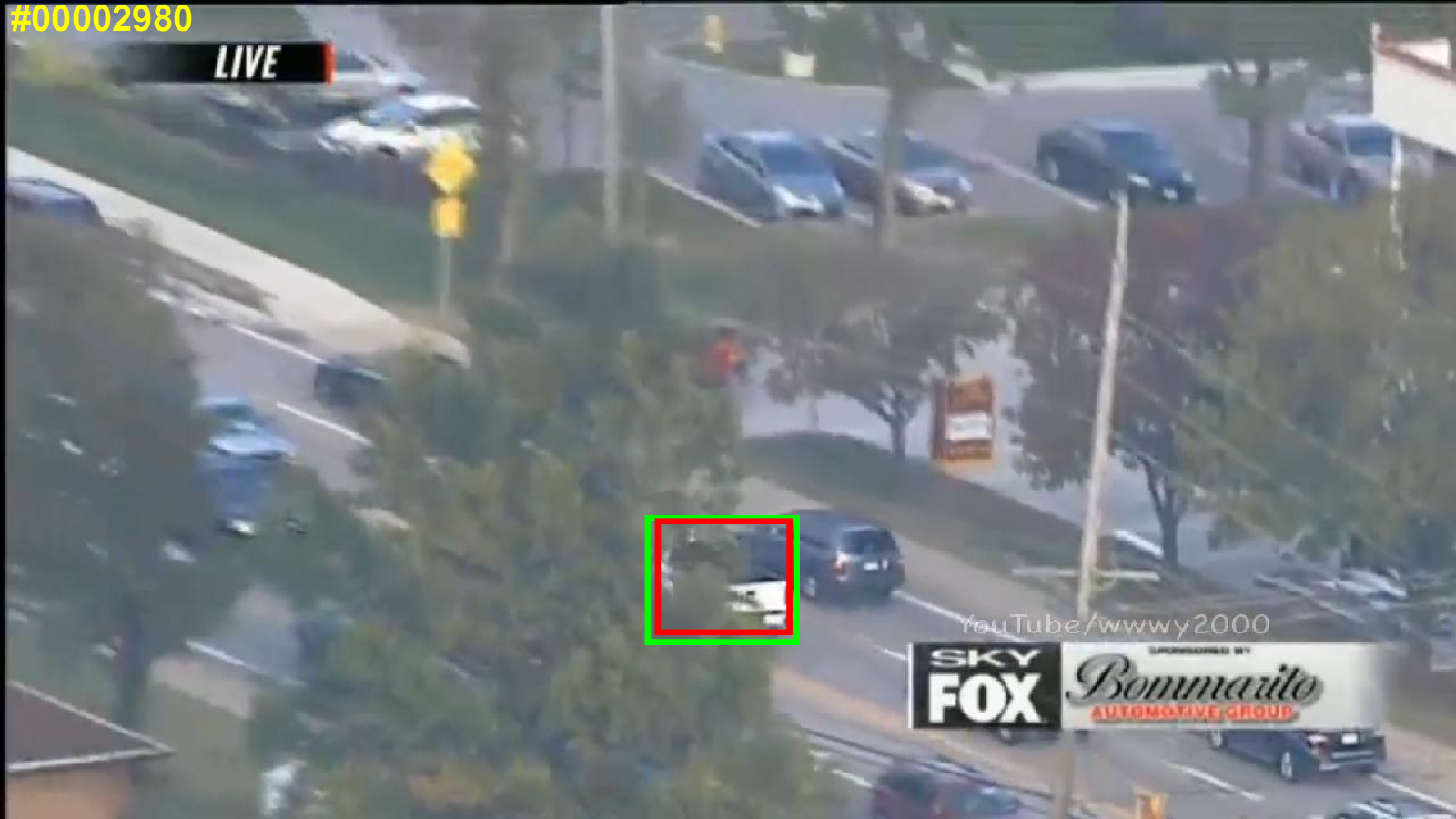}
		\includegraphics[width=0.195\linewidth]{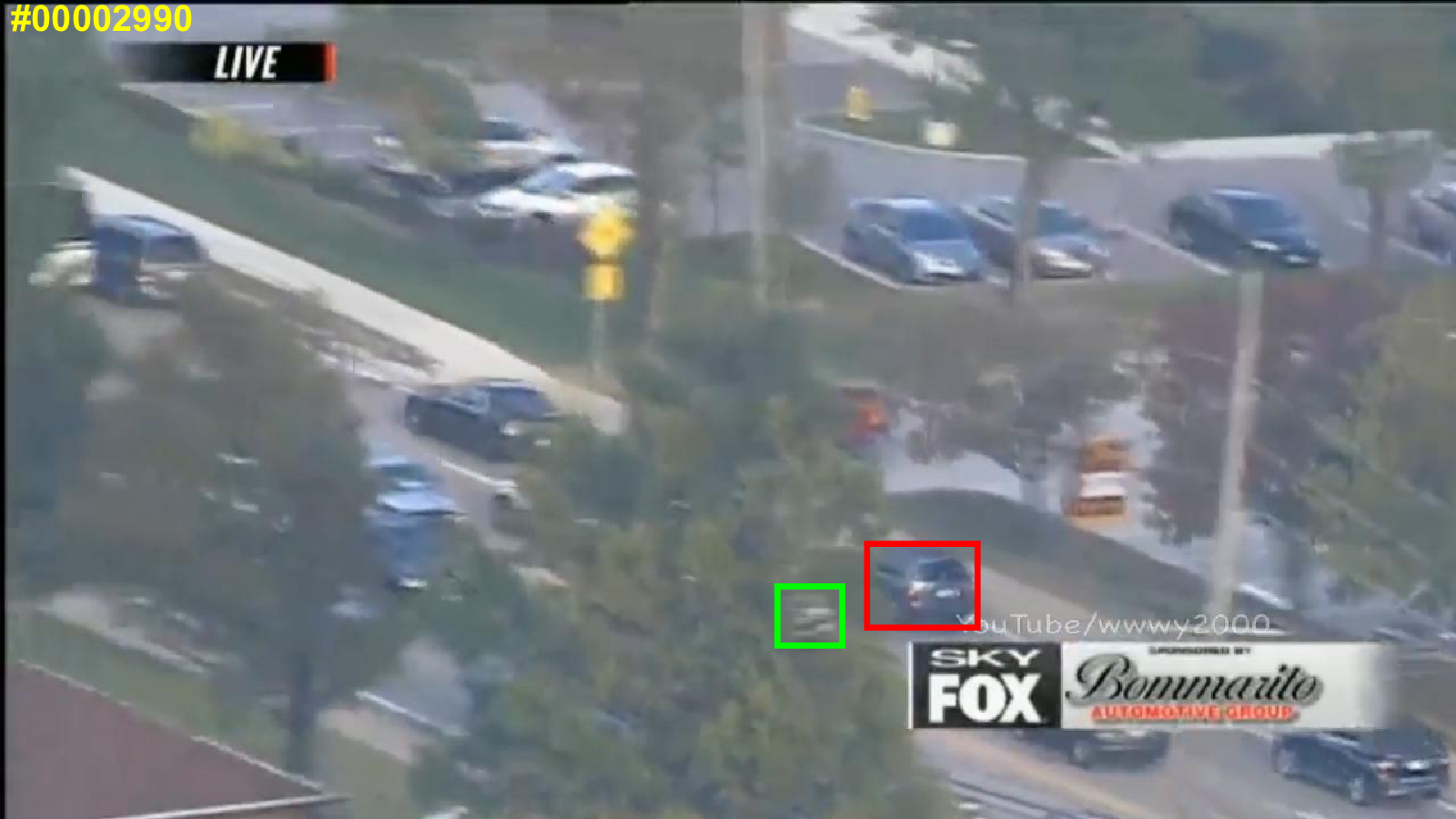}
		\includegraphics[width=0.195\linewidth]{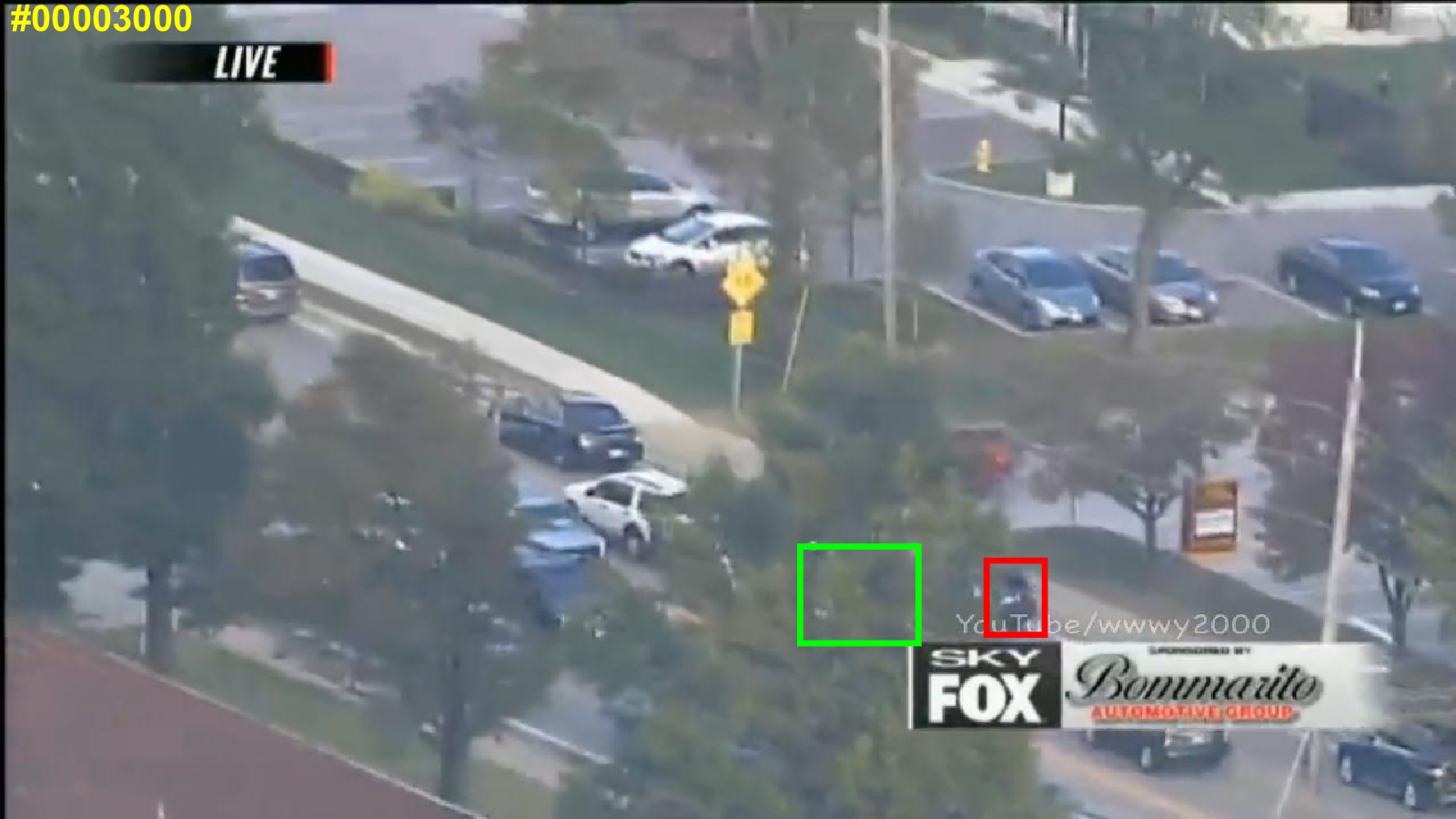}
		\includegraphics[width=0.195\linewidth]{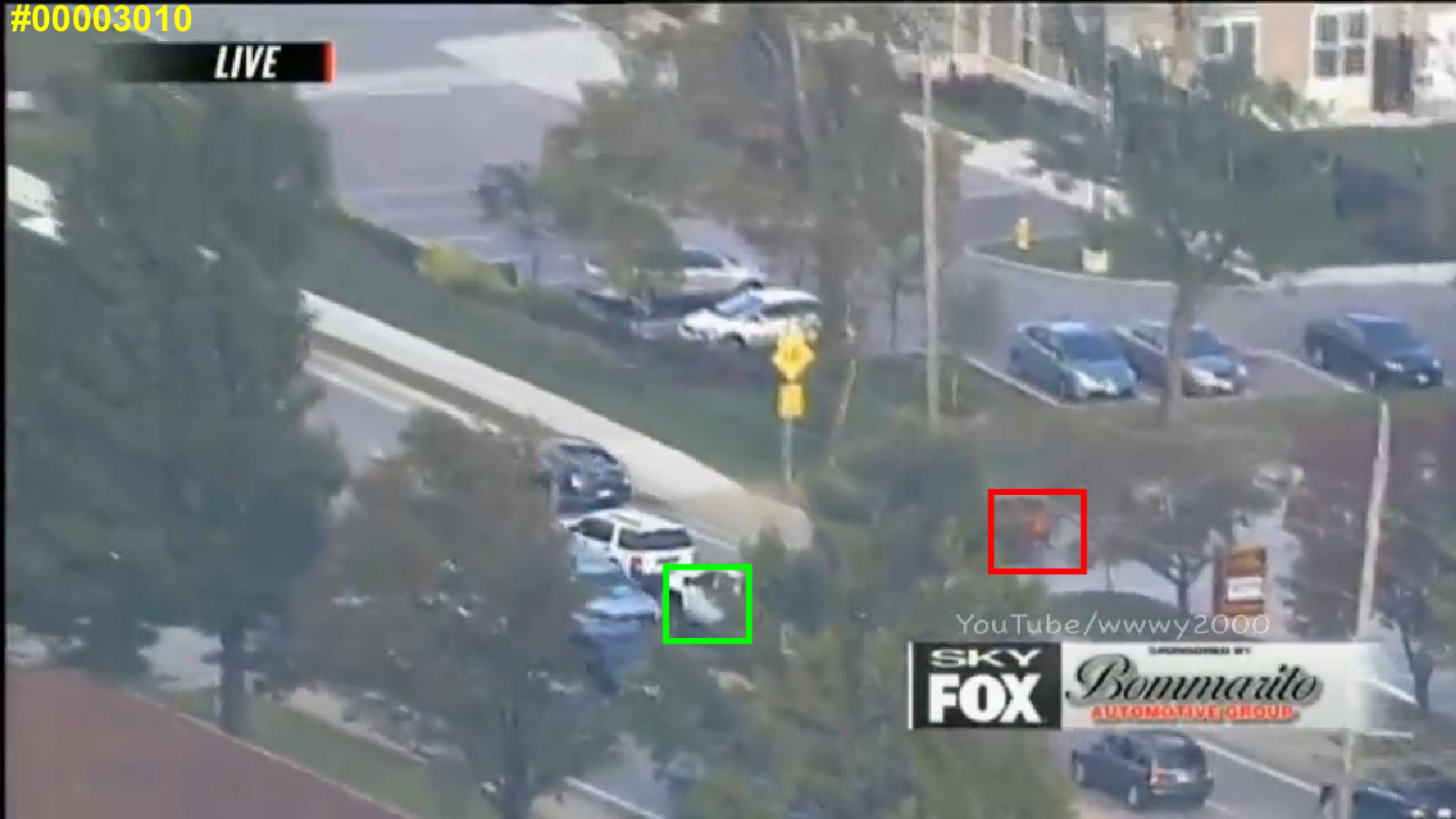}
		
		\includegraphics[width=0.195\linewidth]{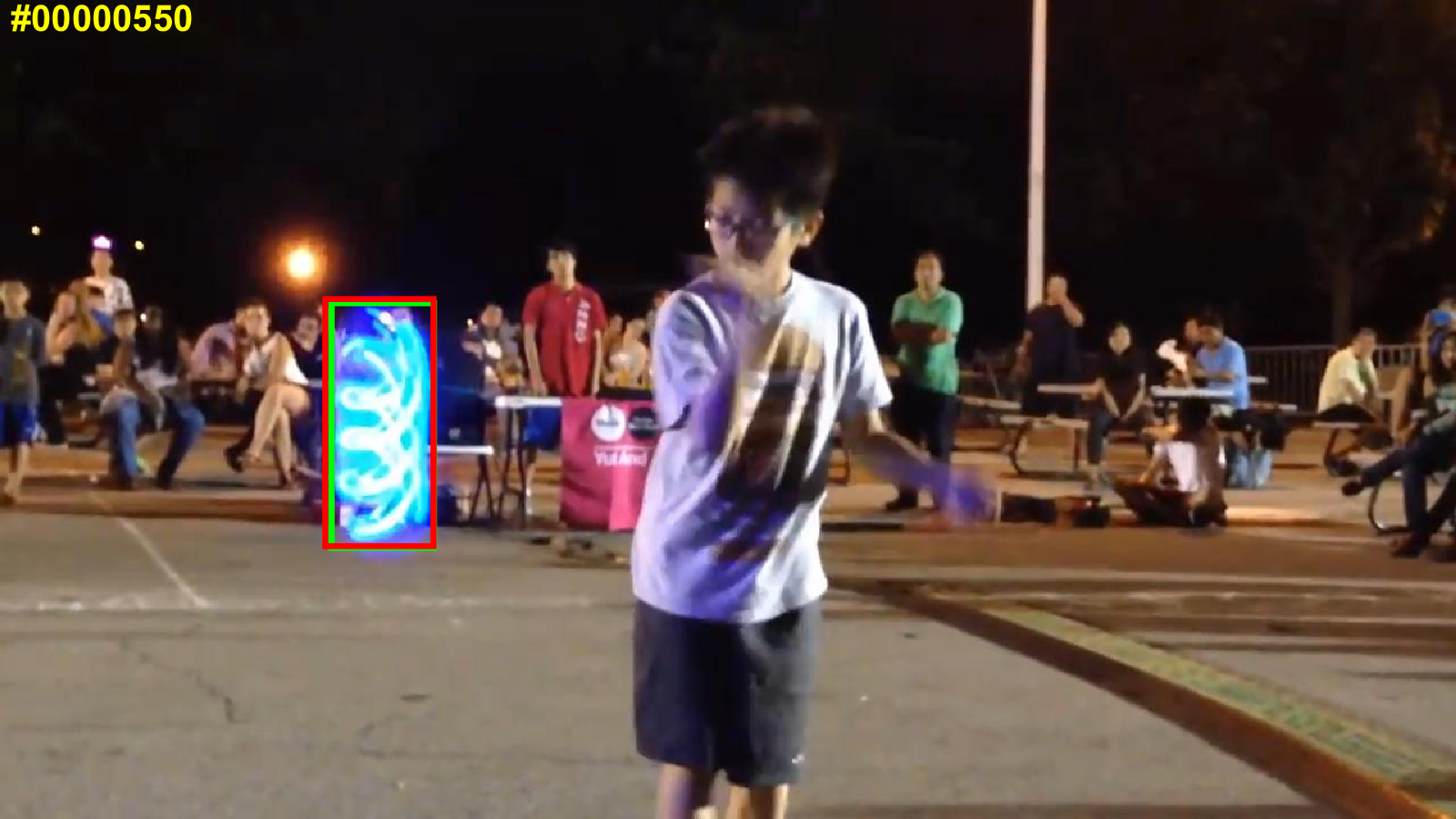}
		\includegraphics[width=0.195\linewidth]{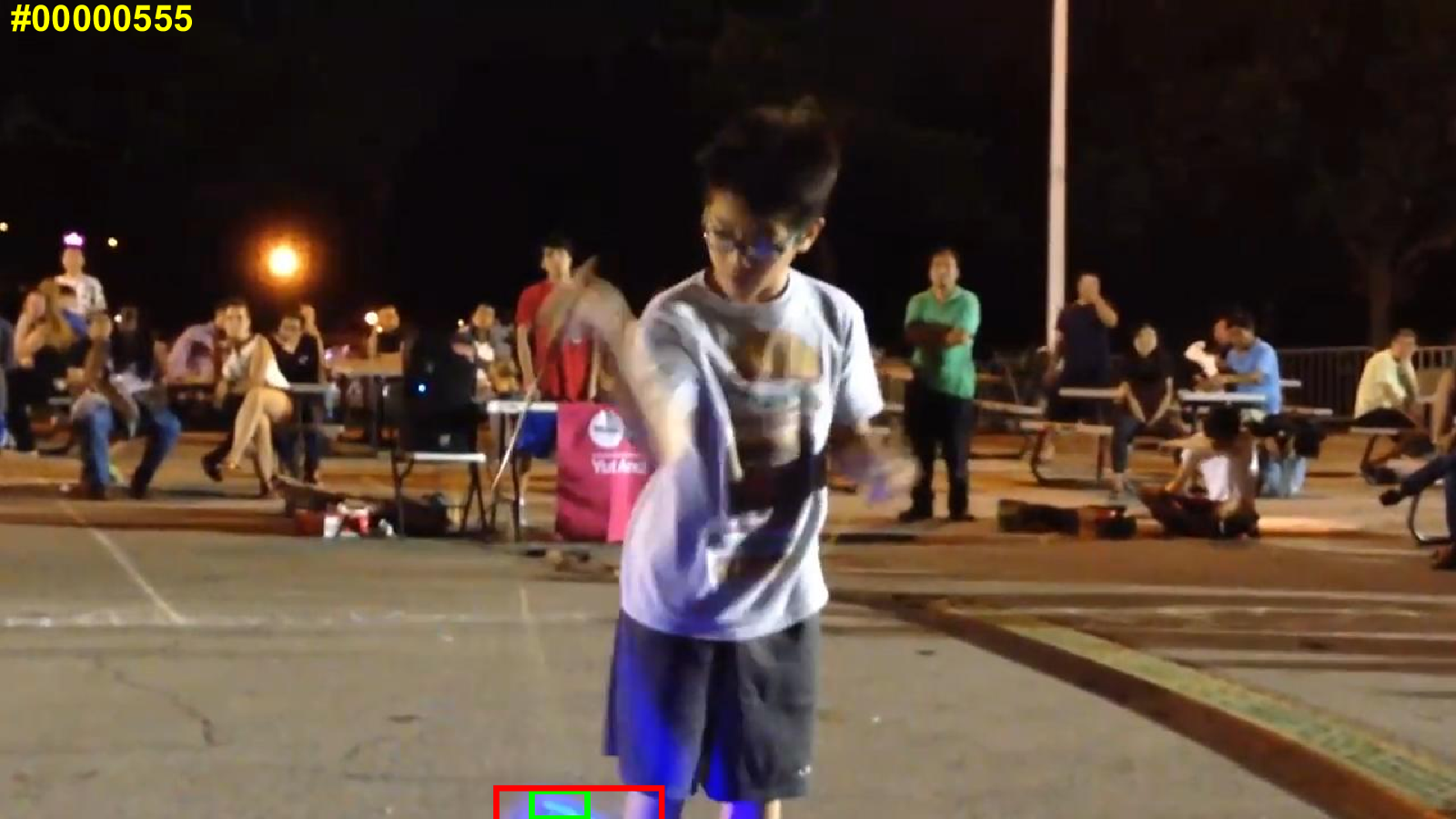}
		\includegraphics[width=0.195\linewidth]{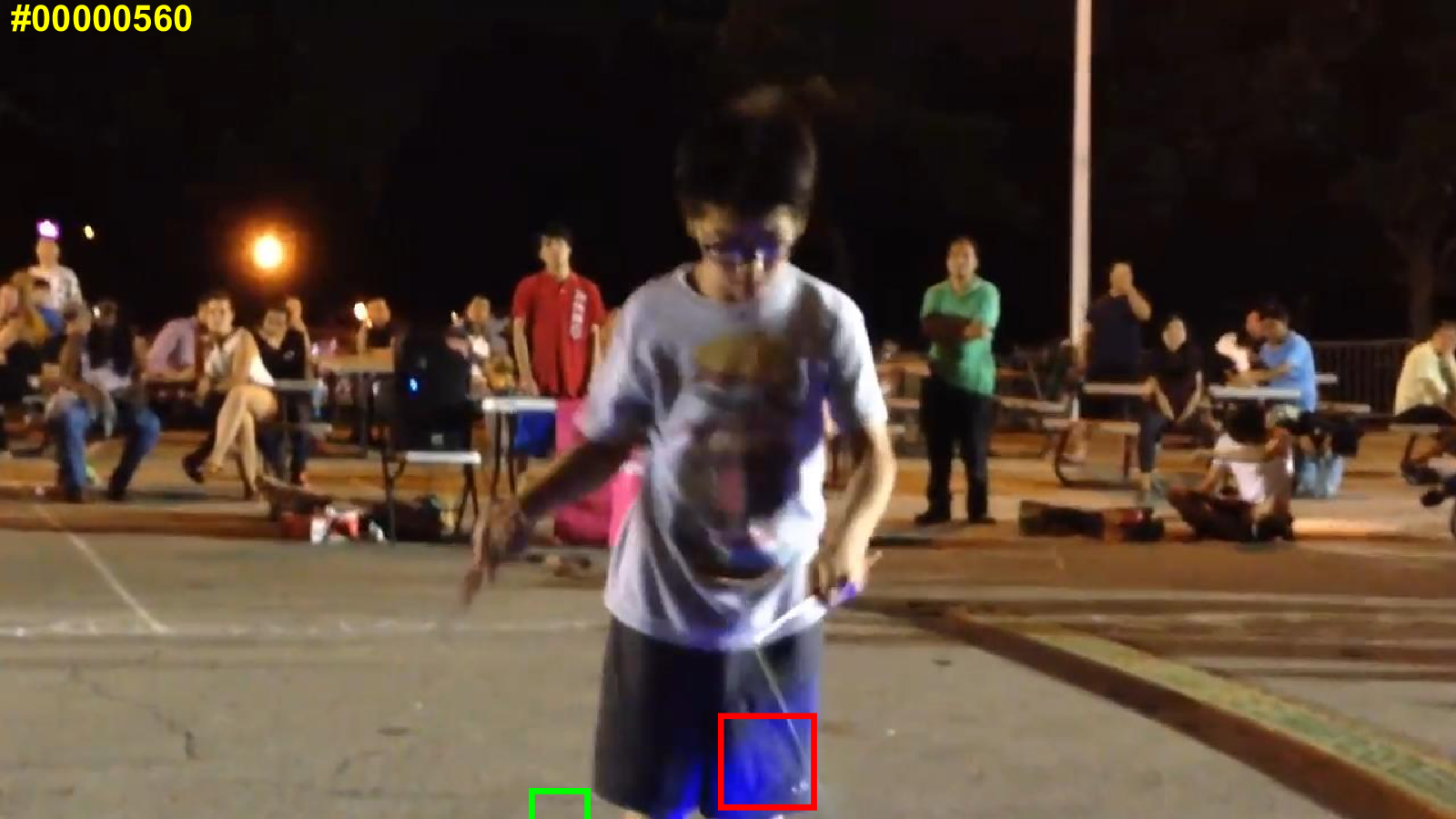}
		\includegraphics[width=0.195\linewidth]{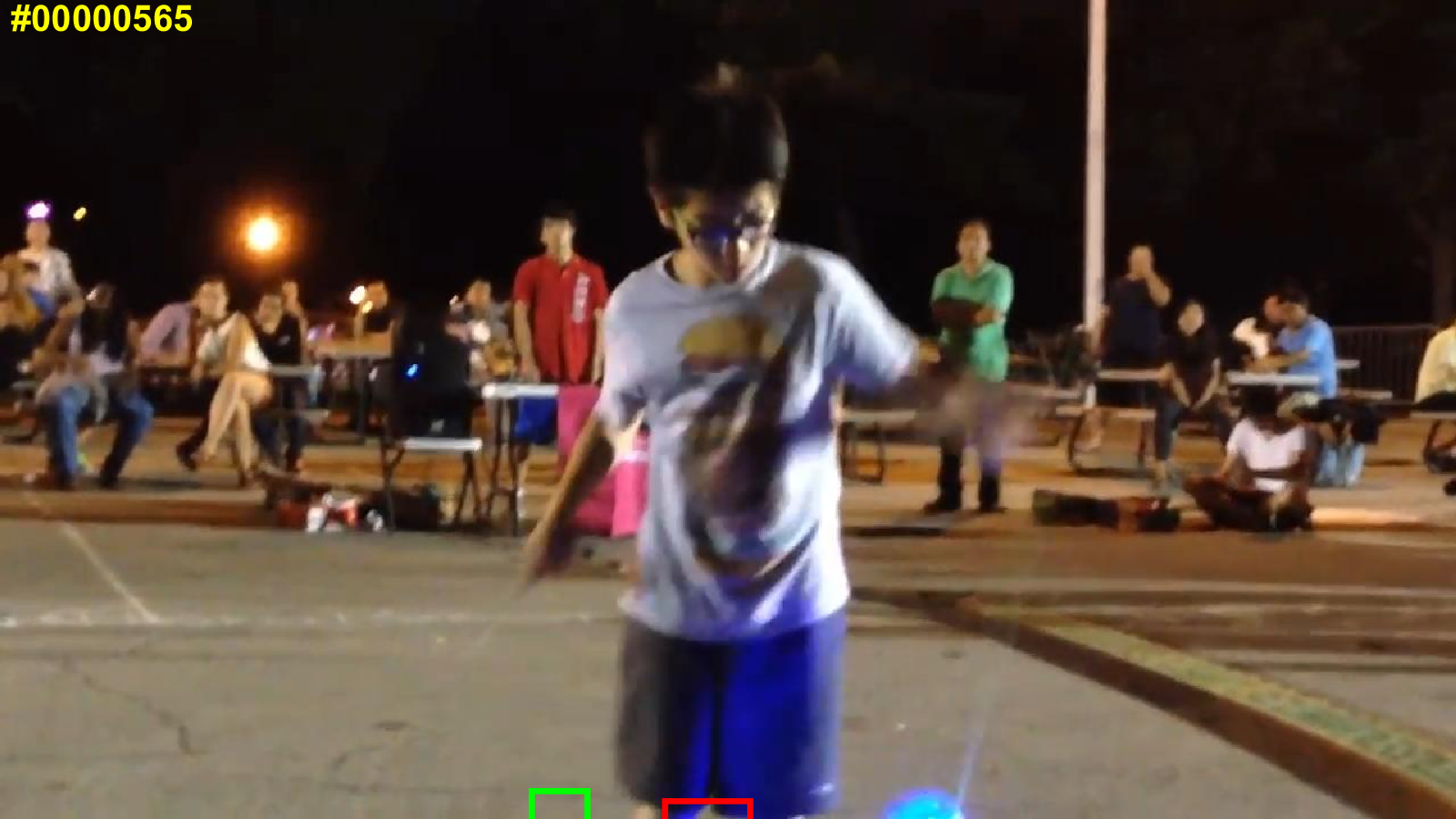}
		\includegraphics[width=0.195\linewidth]{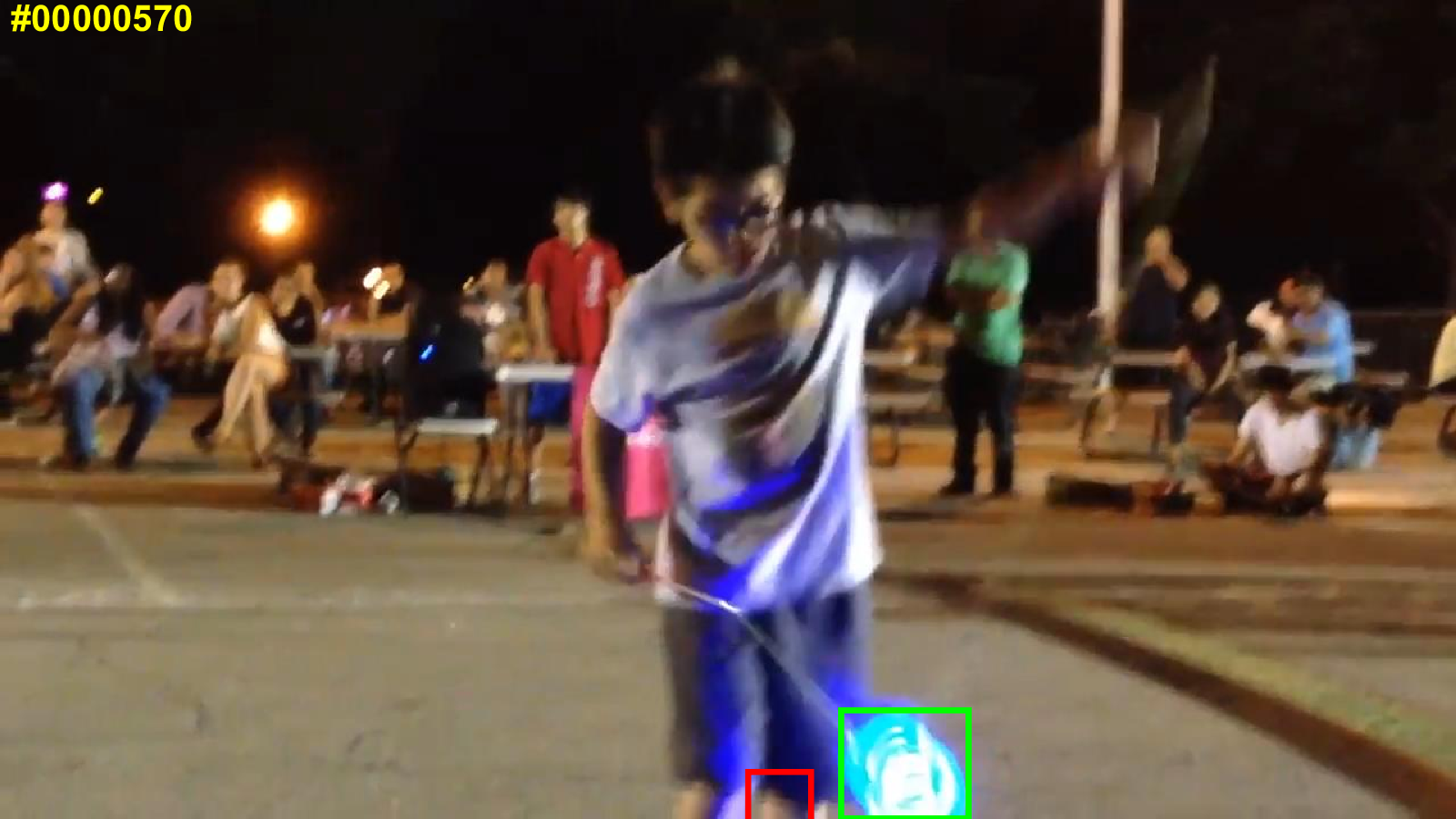}

		\includegraphics[width=0.3\linewidth]{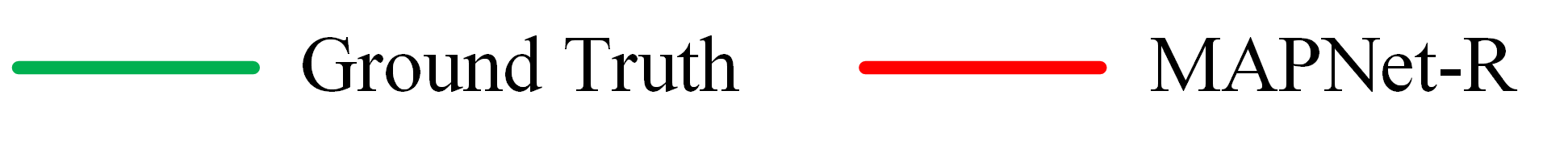}
	\end{center}
	\caption{Failing cases of our tracker on several challenging LaSOT sequences (truck-16, yoyo-15).}
	\label{tenth-fig}
\end{figure*}
  
\subsection{Qualitative Comparisons}
We compare our method with four state-of-the-art trackers qualitatively on a subset of challenging LaSOT \cite{lasot} sequences, and exhibit their tracking results in Fig. \ref{ninth-fig}. These results depict that the presented approach performs better than other recently released algorithms while addressing various distractors. The core reason is that the associate prediction network can fully learn the category-related semantic cues for classification and the spatial texture details for location, which is critical to achieve both accurate and robust tracking.

Concretely, in the sequence of bird-5, the proposed tracker adapts to the scale variations successfully, and tracks the object tightly. In book-10 sequence, our method can precisely locate the object, although its shape changes dramatically. For drone-13, MAPNet-R accurately distinguish the object, proving that our model is robust to background clutter, scale variation and motion blur. In the sequences of coin-18 and zebra-17, the interested objects are severely occluded by other instances. In this case, our approach still can sequentially identify the object, while other algorithms fall into tracking failure frequently.

\subsection{Failure Analysis}
In spite of achieving remarkable performance, the proposed method still falls into tracking failures in a few certain scenes, as shown in Fig. \ref{tenth-fig}. It is easy to observe that our tracker is difficult to correctly track the object after it is out-of-view or is full-occluded by background over a long time. In fact, the propose tracker searchs the object only in a local region, which lacks a global search scheme during tracking. It is why our approach usually fails to detect the object while it reappears in the observation scenarios.

\section{Conclusion}
In this work, we proposed a novel multi-attention associate prediction network for visual tracking, which can estimate the object state in a more effective manner. Firstly, we exploited multiple kinds of attentions to design two special matchers for feature interaction, i.e., category-aware matcher and spatial-aware matcher. Among them, the category-aware matcher can collect sufficient category-related attributes for distinguishing the object from background robustly, while the spatial-aware matcher pays more attention to capturing local spatial textures for accurate location. To the best of our knowledge, it is the first trial to introduce different matchers into an end-to-end decision architecture. Secondly, a dual alignment module was presented to enhance the correspondences between classification and regression branches, improving the overall prediction quality. Massive experimental results on five recent datasets depicted that the Siamese tracker based on associate prediction network outperformed most of state-of-the-art approaches.
 
Despite gaining pretty promising performance, the proposed model still encounters with several fatal drawbacks. The most critical problem is that we do not explore the object temporal contexts for state prediction, which is very important to adapt to the severe appearance variations of object. Therefore, future works may be devoted to studying how to combine multi-stage historic features to identify and locate the current object.

\appendices



\ifCLASSOPTIONcaptionsoff
  \newpage
\fi



%


\bibliographystyle{IEEEtran}
\bibliography{egbib}



%

\begin{IEEEbiography}[{\includegraphics[width=1in,height=1.25in,clip,keepaspectratio]{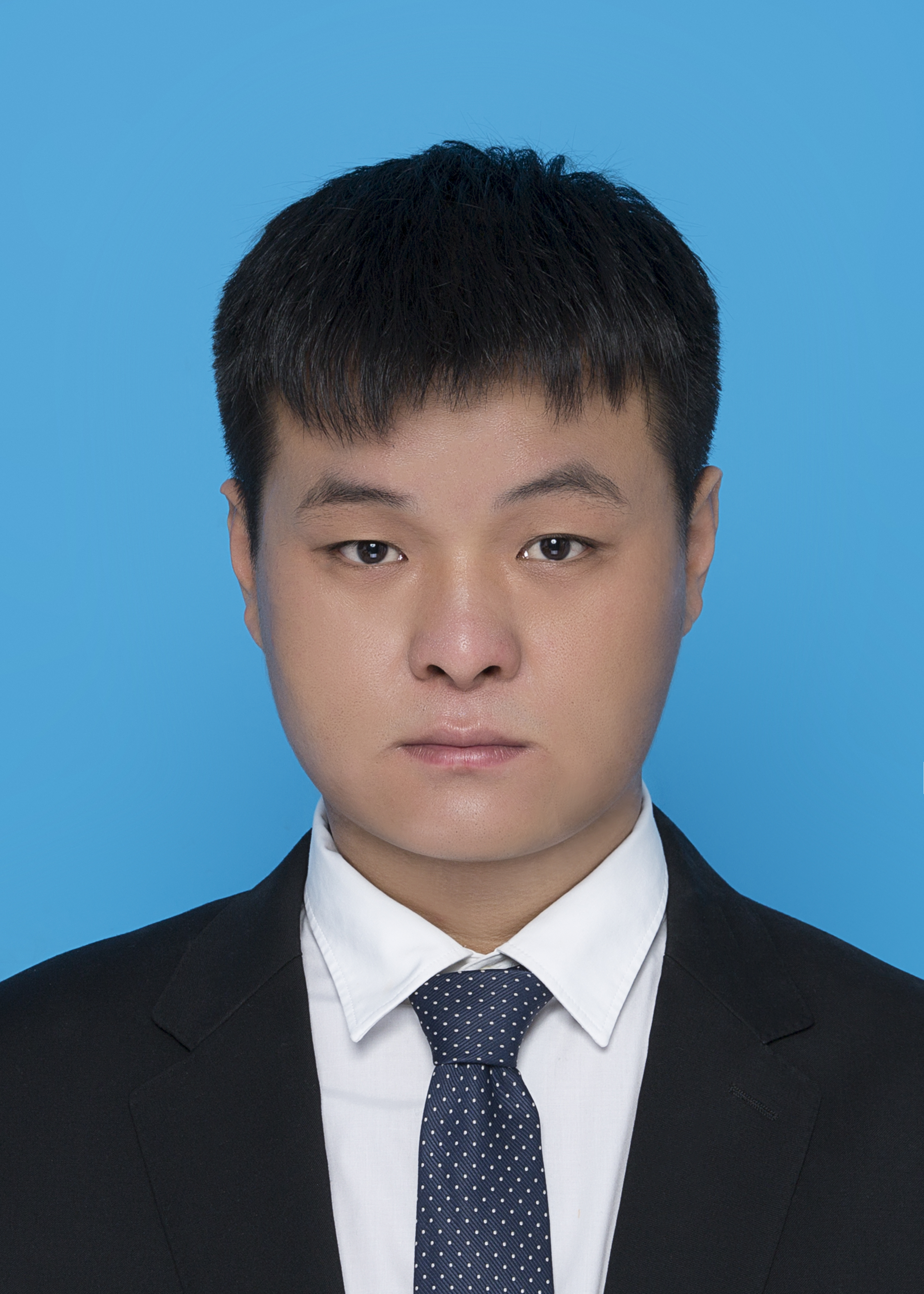}}]{Xinglong Sun}
Received the M.S. degree from Beijing Institute of Technology in 2018, and the Ph.D. degree from Changchun Institute of Optics, Fine Mechanics and Physics, Chinese Academy of Science, in 2022. His current research interests are mainly focused on deep learning, object tracking and image registration.
\end{IEEEbiography}

\begin{IEEEbiography}[{\includegraphics[width=1in,height=1.25in,clip,keepaspectratio]{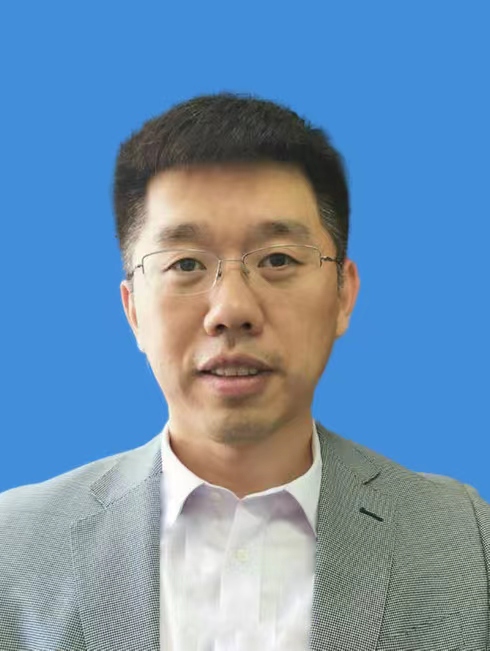}}]{Haijiang Sun}
Received the master’s and Ph.D. degrees from the Changchun Institute of Optics, Fine Mechanics and Physics, Chinese Academy of Sciences. His current research interests include high-speed image processing technology, target automatic recognition, tracking and measurement technology, and optical image enhancement display technology.
\end{IEEEbiography}

\begin{IEEEbiography}[{\includegraphics[width=1in,height=1.25in,clip,keepaspectratio]{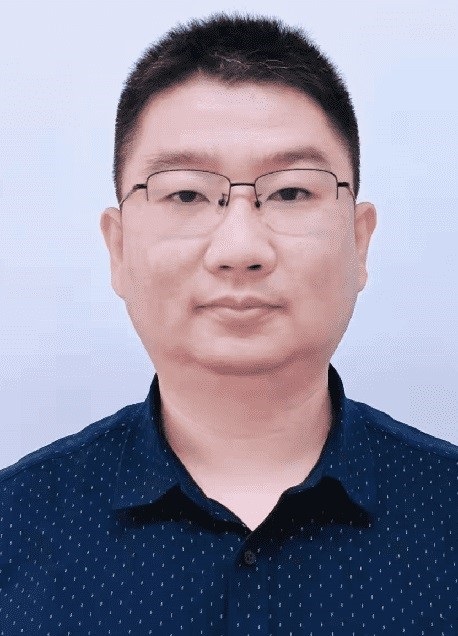}}]{Shan Jiang}
	Received the M.S degree from JiLin University in 2013, Associate researcher of Changchun Institute of Optical Precision Machinery and Physics, Chinese Academy of Sciences, Master tutor, He current research interests are mainly focused on  image processing and artificial intelligence. 
\end{IEEEbiography}

\begin{IEEEbiography}[{\includegraphics[width=1in,height=1.25in,clip,keepaspectratio]{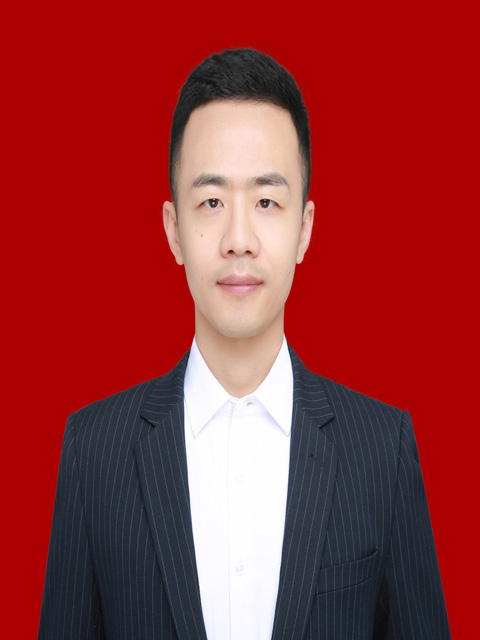}}]{Jiacheng Wang}
	Received the master’s degree in circuits and systems from Xidian University, Xi’an,China, in 2014. He is currently an Assistant  Professor  with Changchun Institute of Optics, Fine Mechanics and Physics, Chinese Academy of Sciences,
	Changchun, China. His research interests include target tracking, computer vision and embedded system design.
\end{IEEEbiography}

\begin{IEEEbiography}[{\includegraphics[width=1in,height=1.25in,clip,keepaspectratio]{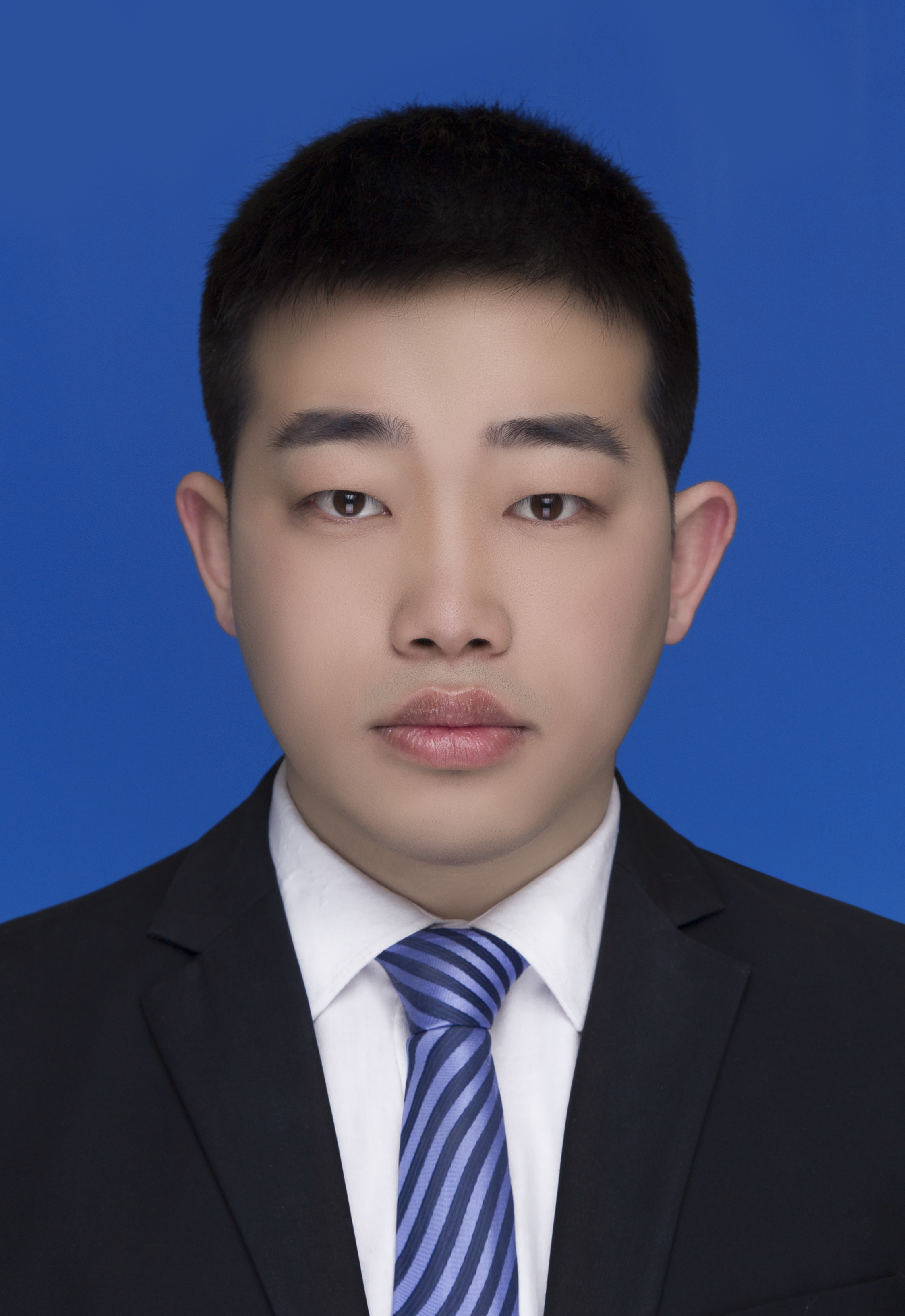}}]{Xilai Wei}
	Received the master's degree from Northeastern University in 2021. His current research interests mainly focus on traditional image processing algorithms, deep learning, and image registration..
\end{IEEEbiography}

\begin{IEEEbiography}[{\includegraphics[width=1in,height=1.25in,clip,keepaspectratio]{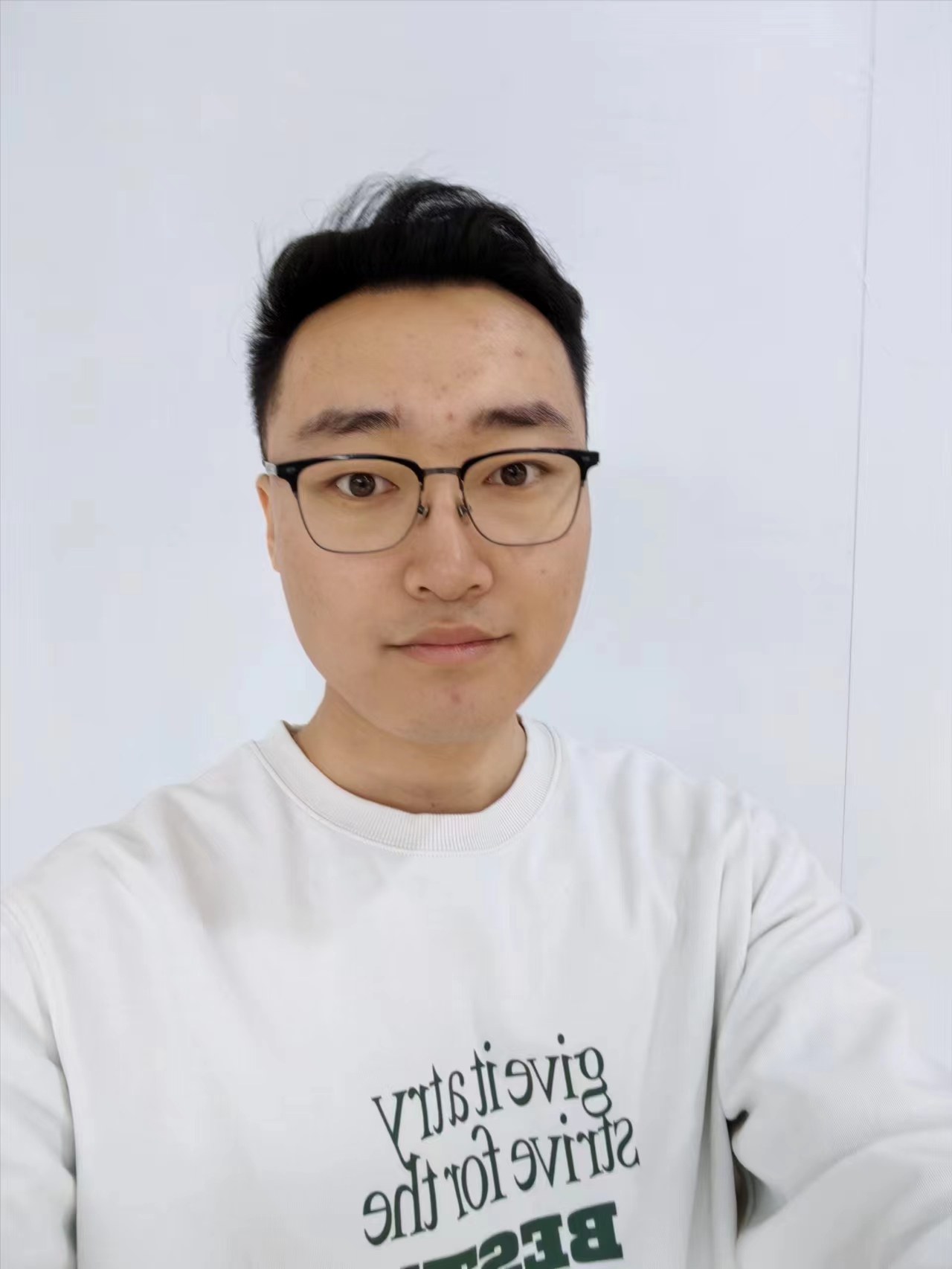}}]{Zhonghe Hu}
	Graduated from National University of Defense Technology with a bachelor's degree in 2018. Currently working as an assistant engineer, his research interests are mainly machine learning and automatic control.
\end{IEEEbiography}




\end{document}